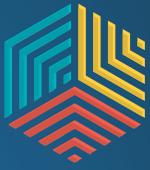

# Tri
# Rhena
# Tech

# ARTIFICIAL INTELLIGENCE
## RESEARCH IMPACT ON KEY INDUSTRIES

EDITED BY
ANDREAS CHRIST
FRANZ QUINT

**COLLECTION OF ACCEPTED PAPERS OF THE CANCELED SYMPOSIUM**
**KARLSRUHE, 13TH MAY 2020**

The Upper-Rhine Artificial Intelligence Symposium

UR-AI 2020

# Artificial Intelligence

## Research Impact on Key Industries

Andreas Christ, Franz Quint (eds.)

Collection of Accepted Papers of the Canceled Symposium

Karlsruhe, 13[th] May 2020

# The Upper-Rhine Artificial Intelligence Symposium

# UR-AI 2020

**Conference Chairs:**

Crispino Bergamaschi, University of Applied Sciences and Arts Northwestern Switzerland
Franz Quint, Karlsruhe University of Applied Sciences

**Program Committee:**

Andreas Christ, Offenburg University of Applied Sciences
Martin Christen, University of Applied Sciences and Arts Northwestern Switzerland
Ulrich Mescheder, Furtwangen University of Applied Sciences
Pierre Parrend, ECAM Strasbourg-Europe
Franz Quint, Karlsruhe University of Applied Sciences
Karl-Herbert Schäfer, Kaiserslautern University of Applied Sciences
Marie Wolkers, Asace Tech Strasbourg

**Organising Committee:**

Anna Dister, TriRhenaTech
Kerstin Heizmann, Karlsruhe University of Applied Sciences
Hendrik Hunsiger, Karlsruhe University of Applied Sciences
Jean Pacevicius, TriRhenaTech
Ira Pawlowski, Offenburg University of Applied Sciences
Marie Rüppell-Wee, Karlsruhe University of Applied Sciences
Elena Stamm, Karlsruhe University of Applied Sciences

The Upper-Rhine Artificial Intelligence Symposium

UR-AI 2020

We thank our sponsor!

# Table of Contents

# Foreword

Development in the field of artificial intelligence (AI) is making giant strides. Allow me to outline some of the areas addressed by the AI sector last year.

Supervised learning methods continued to dominate industrial AI applications, but they require large amounts of (labelled) data, which is time-consuming and costly. Self-supervised learning is about assigning data labelling to an AI. Here, experts label only a small amount of data manually, which is then used by an AI for the automated labelling of the remaining data. Self-supervised learning continues to face a number of technical challenges, including the fact that machine-labelled data must be of high quality if it is to produce results that are almost as good as those arising from manually labelled data.

In the area of privacy-preserving machine learning, many organisations could potentially benefit from sharing data with other, similar organisations to train good models. Health insurers could, for instance, work together on solving the automated processing of unstructured paperwork such as insurers' claim receipts. The issue here is that organisations cannot share their data with each other for confidentiality and privacy reasons, which is why secure collaborative machine learning – where a common model is trained on distributed data to prevent information from the participants from being reconstructed – is gaining traction. This shows that the biggest problem in the area of privacy-preserving machine learning is not technical implementation, but how much the entities involved (decision makers, legal departments, etc.) trust the technologies. As a result, the degree to which AI can be explained, and the amount of trust people have in it, will be an issue requiring attention in the years to come.

The representation of language has undergone enormous development of late: new models and variants, which can be used for a range of natural language processing (NLP) tasks, seem to pop up almost monthly. Such tasks include machine translation, extracting information from documents, text summarisation and generation, document classification, bots, and so forth. The new generation of language models, for instance, is advanced enough to be used to generate completely realistic texts.

These examples reveal the rapid development currently taking place in the AI landscape, so much so that the coming year may well witness major advances or even a breakthrough in the following areas:

- Healthcare sector (reinforced by the COVID-19 pandemic): AI facilitates the analysis of huge amounts of personal information, diagnoses, treatments and medical data, as well as the identification of patterns and the early identification and/or cure of disorders.

- Privacy concerns: how civil society should respond to the fast increasing use of AI remains a major challenge in terms of safeguarding privacy. The sector will need to explain AI to civil society in ways that can be understood, so that people can have confidence in these technologies.

- AI in retail: increasing reliance on online shopping (especially in the current situation) will change the way traditional (food) shops function. We are already seeing signs of new approaches with self-scanning checkouts, but this is only the beginning. Going forward, food retailers will (have to) increasingly rely on a combination of staff and automated technologies to ensure cost-effective, frictionless shopping.

- Process automation: an ever greater proportion of production is being automated or performed by robotic methods.

- Bots: progress in the field of language (especially in natural language processing, outlined above) is expected to lead to major advances in the take-up of bots, such as in customer service, marketing, help desk services, healthcare/diagnosis, consultancy and many other areas.

The rapid pace of development means it is almost impossible to predict either the challenges we will face in the future or the solutions destined to simplify our lives. One thing we can say is that there is enormous potential here. The universities in the TriRhenaTech Alliance are actively contributing interdisciplinary solutions to the development of AI and its associated technical, societal and psychological research questions.

Prof. Dr. Crispino Bergamaschi,
Chair of the TriRhenaTech Alliance,
President of FHNW University of Applied
Sciences and Arts Northwestern Switzerland

# Learning to Walk With Toes

Jens Fischer and Klaus Dorer

Offenburg University of Applied Sciences
{jens.fischer, klaus.dorer}@hs-offenburg.de

**Abstract.** This paper explains how a model-free (with respect to the robot model and the behavior to learn) approach can facilitate learning to walk from scratch. It is applied to a simulated Nao robot with toes. Results show an improvement of 30% in speed compared to a model without toes and also compared to our model-based approach, but with less stability.

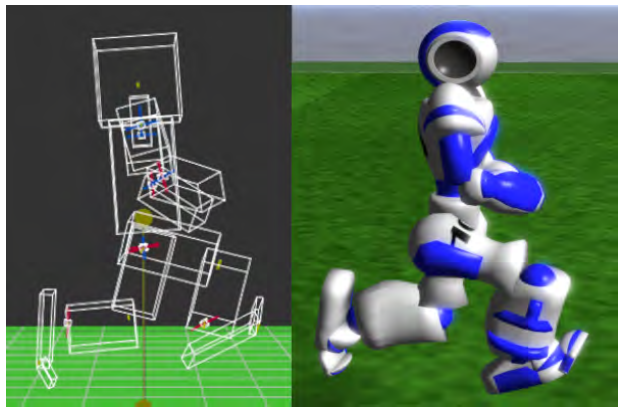

## 1   Introduction

Utilizing toes of a humanoid robot is difficult for various reasons, one of which is that inverse kinematics is overdetermined with the introduction of toe joints. Nevertheless, a number of robots with either passive toe joints like the Monroe or HRP-2 robots [1,2] or active toe joints like Lola, the Toyota robot or Toni [3,4,5] have been developed. Recent work shows considerable progress on learning model-free behaviors using genetic learning [6] for kicking with toes and deep reinforcement learning [7,8,9] for walking without toe joints. In this work, we show that toe joints can significantly improve the walking behavior of a simulated Nao robot and can be learned model-free.

The remainder of this paper is organized as follows: Section 2 gives an overview of the domain in which learning took place. Section 3 explains the approach for model free learning with toes. Section 4 contains empirical results for various behaviors trained before we conclude in Section 5.

## 2    Domain

The robots used in this work are robots of the RoboCup 3D soccer simulation which is based on SimSpark[1] and initially initiated by [10]. It uses the ODE physics engine[2] and runs at an update speed of 50Hz. The simulator provides variations of Aldebaran Nao robots with 22 DoF for the robot types without toes and 24 DoF for the type with toes, NaoToe henceforth. More specifically, the robot has 6 (7) DoF in each leg, 4 in each arm and 2 in its neck. There are several simplifications in the simulation compared to the real Nao:

- all motors of the simulated Nao are of equal strength whereas the real Nao has weaker motors in the arms and different gears in the leg pitch motors.
- joints do not experience extensive backlash
- rotation axes of the hip yaw part of the hip are identical in both robots, but the simulated robot can move hip yaw for each leg independently, whereas for the real Nao, left and right hip yaw are coupled
- the simulated Naos do not have hands
- the touch model of the ground is softer and therefore more forgiving to stronger ground touches in the simulation
- energy consumption and heat is not simulated
- masses are assumed to be point masses in the center of each body part

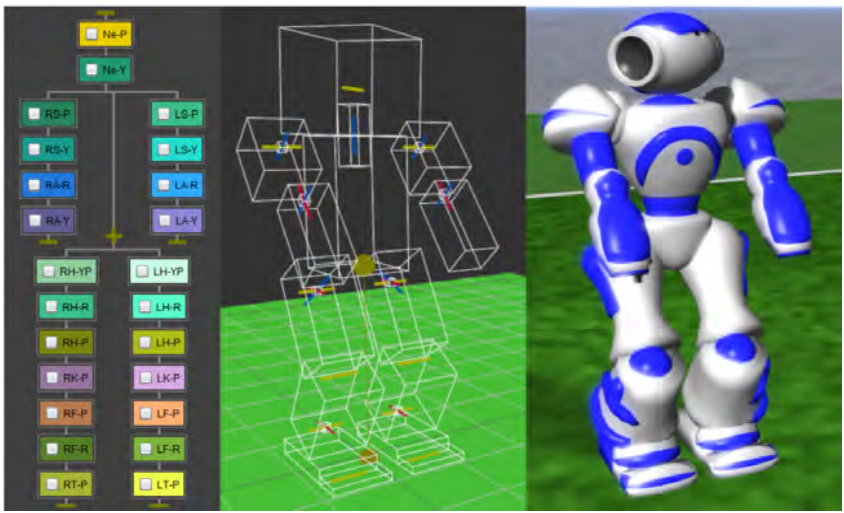

Fig. 1: Joint model (left), wire model (center) and rendered model (right) of the NaoToe robot.

The feet of NaoToe are modeled as rectangular body parts of size 8cm x 12cm x 2cm for the foot and 8cm x 4cm x 1cm for the toes (see Figure 1). The two body parts are connected with a hinge joint that can move from -1 degrees (downward) to 70 degrees.

---

[1] http://simspark.sourceforge.net/
[2] http://www.ode.org/

All joints can move at an angular speed of at most 7.02 degrees per 20ms. The simulation server expects to get the desired speed at 50 Hz for each joint. If no speeds are sent to the server it will continue movement of the joint with the last speed received. Joint angles are noiselessly perceived at 50Hz, but with a delay of 40ms compared to sent actions. So only after two cycles, the robot knows the result of a triggered action. A controller provided for each joint inside the server tries to achieve the requested speed, but is subject to maximum torque, maximum angular speed and maximum joint angles.

The simulator is able to run 22 simulated Naos in real-time on reasonable CPUs. It is used as competition platform for the RoboCup 3D soccer simulation league[3]. In this context, only a single agent was running in the simulator.

# 3  Approach

The following subsections describe how we approached the learning problem. This includes a description of the design of the behavior parameters used, what the fitness functions for the genetic algorithm look like, which hyperparameters were used and how the fitness calculation in the SimSpark simulation environment works exactly.

## 3.1  Behavior Parameters

The guiding goal behind our approach is to learn a model-free walk behavior. With model-free we depict an approach that does not make any assumptions about a robot's architecture nor the task to be performed. Thus, from the viewpoint of learning, our model consists of a set of flat parameters. These parameters are later grounded inside the domain.

The server requires 50 values per second for each joint. To reduce the search space, we make use of the fact that output values of a joint over time are not independent. Therefore, we learn *keyframes*, i.e. all joint angles for discrete phases of movement together with the duration of the phase from keyframe to keyframe. The experiments described in this paper used four to eight of such phases. The number of phases is variable between learning runs, but not subject to learning for now, except for skipping phases by learning a zero duration for it.

The RoboCup server requires robots to send the actual angular speed of each joint as a command. When only leg joints are included, this would require to learn 15 parameters per phase (14 joints + 1 for the duration of the phase), resulting in 60, 90 and 120 parameters for the 4, 6, 8 phases worked with. The disadvantage of this approach is that the speed during a particular phase is constant, thus making it unable to adapt to discrepancies between the desired and the actual motor movement.

Therefore, a combination of angular value and the maximum amount of angular speed each joint should have is used. The direction and final value of movement is entirely encoded in the angular values, but the speed can be controlled separately. It follows that:

- If the amount of angular speed does not allow reaching the angular value, the joint behaves like in version 1.
- If the amount of angular speed is bigger, the joint stops to move even if the phase is not over.

---
[3] http://www.robocup.org/robocup-soccer/simulation/

This almost doubles the amount of parameters to learn, but the co-domain of values for the speed values is half the size, since here we only require an absolute amount of angular speed. With these parameters, the robot learns a single step and mirrors the movement to get a double step.

## 3.2 Fitness Function

Feedback from the domain is provided by a fitness function that defines the utility of a robot. The fitness function subtracts a penalty for falling from the walked distance in X-direction in meters. There is also a penalty for the maximum deviation in Y-direction reached during an episode, weighted by a constant factor. In practice, the values chosen for $fallenPenalty$ and a factor $f$ were usually 3 and 2 respectively.

$$fitness_{walk} = distanceX - fallenPenalty - (f * maxY) \qquad (1)$$

This same fitness function can be used without modification for forward, backward and sideward walk learning, simply by adjusting the initial orientation of the agent. The also trained turn behavior requires a different fitness function.

$$fitness_{turn} = (g * totalTurn) - distance \qquad (2)$$

Where $totalTurn$ refers to the cumulative rotation performed in degrees, weighted by a constant factor $g$ (typically 1/100). We penalize any deviation from the initial starting X / Y position ($distance$) as incentive to turn in-place. It is noteworthy that other than swapping out the fitness function and a few more minor adjustments mentioned in 3.3, everything else about the learning setup remained the same thanks to the model-free approach.

## 3.3 Fitness Calculation

Naturally, the fitness calculation for an individual requires connecting an agent to the SimSpark simulation server and having it execute the behavior defined by the learned parameters. In detail, this works as follows:

At the start of each "episode", the agent starts walking with the old model-based walk engine at full speed. Once 80 simulation cycles (roughly 1.5 seconds) have elapsed, the robot starts checking the foot force sensors. As soon as the left foot touches the ground, it switches to the learned behavior. This ensures that the learned walk has comparable starting conditions each time. If this does not occur within 70 cycles (which sometimes happens due to non-determinism in the domain and noise in the foot force perception), the robot switches anyway.

From that point on, the robot keeps performing the learned behavior that represents a single step, alternating between the original learned parameters and a mirrored version (right step and left step). An episode ends once the agents has fallen or 8 seconds have elapsed.

To train different walk directions (forward, backward, sideward), the initial orientation of the player is simply changed accordingly. In addition, the robot uses a different walk direction of the model-based walk engine for the initial steps that are not subject to learning.

In case of training a morphing behavior (see 4.5), the episode duration is extended to 12 seconds. When a morphing behavior should be trained, the step behavior from another learning run is used. This also means that a morphing behavior is always trained for a specific set of walk parameters. After 6 seconds, the morphing behavior is triggered once the foot force sensors detect that the left foot has just touched the ground. Unlike the step / walk behavior, this behavior is just executed once and not mirrored or repeated. Then the robot switches back to walking at full speed with the model-based walk engine. To maximize the reward, the agent has to learn a morphing behavior that enables the transition between learned model-free and old model-based walk to work as reliably as possible.

Finally, for the turn behavior, the robot keeps repeating the learned behavior without alternating with a mirrored version. In any case, if the robot falls, a training run is over.

Experiments were run over 200 generations with 10 oversampling runs per robot to average out non-determinism. Combined with the population size of 200, this means that 400000 fitness calculations are done per learning run:

$$200 \text{ (Population Size)} * 200 \text{ (Generations)} * 10 \text{ (Oversampling)} = 400000 \qquad (3)$$

The overall runtime of each such learning run is 2.5 days on our hardware.

### 3.4 Hyperparameters

Learning is done using plain genetic algorithms. The following hyperparameters were used:

- Population Size: **200 Individuals**
- Genders: **2**
- Parents per Individual: **2**
- Selection Strategy: **MonteCarloSelection**
- Reproduction Strategy: **MultiCrossoverRecombination**
- Mutation Strategy: **RandomMutation**
- Mutation Probability per Individual: **10%**
- Mutation Probability per Gene: **20%**

More details on the approach can be found in [11].

## 4 Results

This section presents the results for each kind of behavior trained. This includes three different walk directions, a turn behavior and a behavior for morphing.

### 4.1 Forward Walk

The main focus of this work has been on training a forward walk movement. Figure 2 shows a sequence of images for a learned step. The best result reaches a speed of 1.3 m/s compared to the 1.0 m/s of our model-based walk and 0.96 m/s for a walk behavior learned on the Nao robot without toes. The learned walk with toes is less stable, however, and shows a fall rate of 30% compared to 2% of the model-based walk.

Regarding the characteristics of this walk, it utilizes remarkably long steps[4]. Table 1 shows an in-depth comparison of various properties, including step duration, length and height, which are all considerably bigger compared to our previous model-based walk. The forward leaning of the agent has increased by 80.4%, while 28.1% more time is spent with both legs off the ground. However, the maximum deviation from the intended path ($maxY$) has also increased by 137.8%.

---

[4] https://youtu.be/ytM61yTcJ-Q

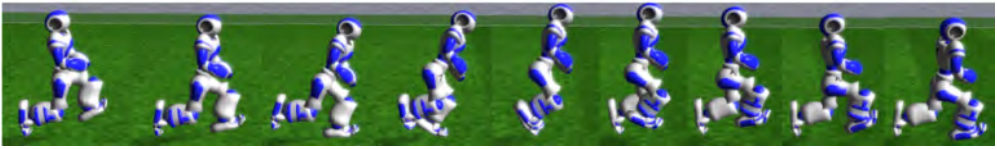

Fig. 2: Movement sequence of a learned walk behavior with toes.

| Value | Old | New | Diff | Diff % |
|---|---|---|---|---|
| utility | 5.102 | 5.118 | +0.016 | +0.3% |
| distanceX | 6.095 | 7.482 | +1.387 | +22.8% |
| speedX | 0.993 | 1.291 | +0.298 | +30% |
| maxY | 0.497 | 1.182 | +0.685 | +137.8% |
| bothLegsOffGround | 0.495 | 0.634 | +0.139 | +28.1% |
| oneLegOffGround | 0.507 | 0.366 | -0.141 | -27.8% |
| noLegsOffGround | 0 | 0.003 | +0.003 | |
| stepDuration | 0.08 | 0.301 | +0.221 | +276.3% |
| stepLength | 0.13 | 0.429 | +0.299 | +230% |
| stepHeight | 0.02 | 0.123 | +0.103 | +515% |
| leaningX | 0 | 0.001 | +0.001 | |
| leaningY | -0.107 | -0.193 | -0.086 | -80.4% |

Table 1: Comparison of the previously fastest and the fastest learned forward walk

## 4.2 Backward Walk

Once a working forward walk was achieved, it was natural to try to train a backward walk behavior as well, since this only requires a minor modification in the learning environment (changing the initial rotation of the agent and model-based walk direction to start with). The best backward walk learned reaches a speed of 1.03 m/s, which is significantly faster than the 0.67 m/s of its model-based counterpart. Unfortunately, the agent also falls 15% more frequently.

It is interesting just how backward-leaning the agent is during this walk behavior. It could almost be described as "controlled falling"[5] (see Figure 3).

## 4.3 Sideward Walk

Sideward walk learning was the least successful out of the three walk directions. Like with all directions, the agent starts out using the old walk engine and then switches to the learned behavior after a short time. In this case however, instead of continuing to walk sideward, the agent has learned to turn around and walk forward instead, see Figure 4.

The resulting forward walk is not very fast and usually causes the agent to fall within a few meters[6], but it is still remarkable that the learned behavior manages to both turn the agent around *and* make it walk forward with the same repeating step movement. It is also remarkable that the robot learned that it is quicker with the given legs at least for long distances to turn and run forward than to keep making sidesteps.

---

[5] https://youtu.be/4mqO6V_Sk9Y

[6] https://youtu.be/4WnPzM1pPfU

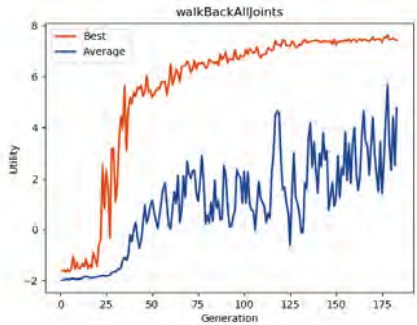
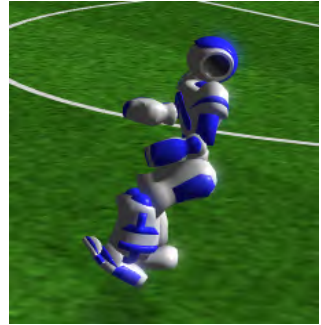

(a) Learning curve       (b) Heavy backward lean

Fig. 3: Learning curve and backward lean of the backward walk

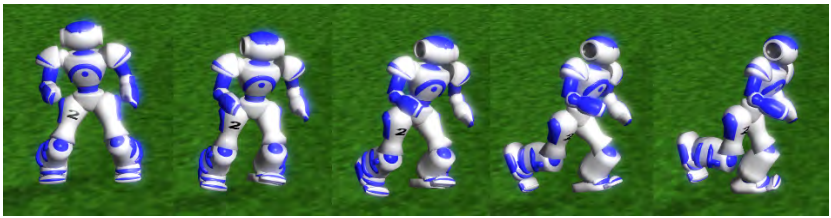

Fig. 4: Change of the walk direction during sideward walk learning

## 4.4 Turn

With the alternate fitness function presented in section 3, the agent managed to learn a turn behavior that is comparable in speed to that of the existing walk engine. Despite this, the approach is actually different: while the old walk engine uses small, angled steps[7], the learned behavior uses the left leg as a "pivot", creating angular momentum with the right leg[8]. Figure 5 shows the movement sequence in detail.

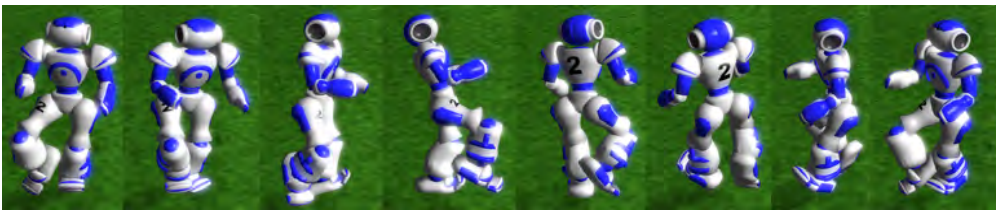

Fig. 5: Movement sequence of a learned turn behavior.

Unfortunately, despite the comparable speed, the learned turn behavior suffers from much worse stability. With the old turn behavior, the agent only falls in roughly 3% of cases, with the learned behavior it falls in roughly 55% of the attempts.

---

[7] https://youtu.be/PryjLMXIta0
[8] https://youtu.be/xgyZtCS68Wo

## 4.5 Morphing

One of the major hurdles for using the learned walk behaviors in a RoboCup competition is the smooth transition between them and other existing behaviors such as kicks. The initial transition to the learned walk is already built into the learning setup described in 3 by switching mid-walk, so it does not have to be given special consideration. More problematic is switching to another behavior afterwards without falling.

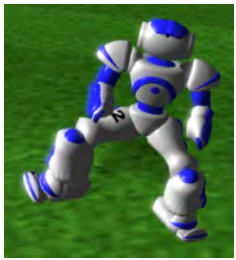

Fig. 6: Lunge performed by the trained morphing behavior.

To handle this, the robot simply attempted to train a "morphing" behavior using the same model-free learning setup. The result is something that could be described as a "lunge" (see Figure 6) that reduces the forward momentum sufficiently to allow it to transition to the slower model-based walk when successful.[9] However, the morphing is not successful in about 50% of cases, resulting in a fall.

## 5 Conclusion and Future Work

We were able to successfully train forward and backward walk behaviors, as well as a morphing and turn behavior using plain genetic algorithms and a very flexible model-free approach. The usage of the toe joint in particular makes the walks look more natural and human-like than that of the model-based walk engine.

However, while the learned behaviors outperform or at least match our old model-based walk engine in terms of speed, they are not stable enough to be used during actual RoboCup 3D Simulation League competitions. We think this is an inherent limitation of the approach: We train a static behavior that is unable to adapt to changing circumstances in the environment, which is common in SimSpark's non-deterministic simulation with perception noise.

*Deep Reinforcement Learning* seems more promising in this regard, as the neural network can dynamically react to the environment since sensor data serves as input. It is also arguably even less restrictive than the keyframe-based behavior parameterization we presented in this paper, as a neural network can output raw joint actions each simulation cycle. At least two other RoboCup 3D Simulation League teams, FC Portugal [8] and ITAndroids [9], have had great success with this approach, Everything points towards this becoming the *state-of-the-art* approach in RoboCup 3D Soccer Simulation in the near future, so we want to concentrate our future efforts here as well.

---

[9] `https://youtu.be/NB_4i2_b-Pg`

# Machine Learning assisted Gross Weight Prediction for Alcoholic Beverages in Retail

Christian Schorr

Trier University of Applied Sciences, Environmental Campus Birkenfeld
**c.schorr@umwelt-campus.de**

**Abstract.**

In this paper we present a machine learning pipeline developed specifically for the product group of alcoholic beverages with focus on the two segments wine and beer which constitute the major part of a retailer's alcoholic beverages inventory. We focus on exploiting expert knowledge about the data domain to engineer features tailored to prediction of the important attribute gross weight. Experiments with data from a major retail company show that our proposed machine learning approach with feature enriched data achieves superior results which are more robust than those obtained by traditional heuristic approaches on the original data. In practical terms this is a step towards fully automated product data generation and maintenance reducing manual effort and thus costs for a retail company.

**Keywords:** Product Classification, Feature Engineering, Machine Learning

## 1    Introduction

Retail companies dealing in alcoholic beverages are faced with a constant flux of products. Apart from general product changes like modified bottle designs and sizes or new packaging units two factors are responsible for this development. The first is the natural wine cycle with new vintages arriving at the market and old ones cycling out each year. The second is the impact of the rapidly growing craft beer trend which has also motivated established breweries to add to their range. The management of the corresponding product data is a challenge for most retail companies. The reason lies in the large amount of data and its complexity. Data entry and maintenance processes are linked with considerable manual effort resulting in high data management costs. Product data attributes like dimensions, weights and supplier information are often entered manually into the data base and are often afflicted with errors. Another widely used source of product data is the import from commercial data pools. A means of checking the data thus acquired for plausibility is necessary. Sometimes product data is incomplete due to different reasons and a method to fill the missing values is required. All these possible product data errors lead to complications in the downstream automated purchase and logistics processes.

We propose a machine learning model which involves domain specific knowledge and compare it a heuristic approach by applying both to real world data of a retail company. In this paper we address the problem of predicting the gross weight of product items in the merchandise category alcoholic beverages. To this end we introduce two levels of additional features. The first level consists of engineered features which can be determined by the basic features alone or by domain specific expert knowledge like which type of bottle is usually used for which grape variety. In the next step an advanced second level feature is computed from these first level features. Adding these two levels of engineered features increases the prediction quality of the suggestion values we are looking for. The results emphasize the importance of careful feature engineering using expert knowledge about the data domain.

## 2    Related Work

Feature Engineering is the process of extracting features from the data in order to train a prediction model. It is a crucial step in the machine learning pipeline, because the quality of the prediction is based on the choice of features used to training. The majority of time and effort in building a machine learning pipeline is spent on data cleaning and feature engineering [Domingos 2012]. A first overview of basic feature engineering principles can be found in [Zheng 2018]. The main problem is the dependency of the feature choice on the data set and the prediction algorithm. What works best for one combination does not necessarily work for another. A systematic approach to feature engineering without expert knowledge about the data is given in [Heaton 2016]. The authors present a study whether different machine learning algorithms are able to synthesize engineered features on their own. As engineered features logarithms, ratios, powers and other simple mathematical functions of the original features are used. In [Anderson 2017] a framework for automated feature engineering is described.

## 3    Data Set

The data set is provided by a major German retail company and consists of 3659 beers and 10212 wines. Each product is characterized by the seven features shown in table 1. The product name obeys only a generalized format. Depending on the user generating the product entry in the company data base, abbreviating style and other editing may vary. The product group is a company specific number which encodes the product category - dairy products, vegetables or soft drinks for example. In our case it allows a differentiation of the product into beer and wine. Additionally wines are grouped by country of origin and for Germany also into wine-growing regions. Note that the product group is no inherent feature like length, width, height and volume, but depends on the product classification system a company uses. The dimensions length, width, height and the volume derived by multiplicating them are given as float values. The feature (gross) weight, also given as a float value, is what we want to predict.

### 3.1    Feature description

| Feature | Type | Unit | Example |
|---|---|---|---|
| Product name | string | - | MW  PFUNGSTAED  EDEL  PILS  EXCLUSIV |
| Product group | int | - | 47114 |
| Length | float | mm | 69.0 |
| Width | float | mm | 69.0 |
| Height | float | mm | 270.0 |
| Volume | float | ml | 1285470.0 |
| Gross Weight | float | g | 915.0 |

**Table. 1.** Data set features, corresponding types, units and example

## 3.2 General Pre-Processing

As is often the case with real world data, a pre-processing step has to be performed prior to the actual machine learning in order to reduce data errors and inconsistencies. For our data we first removed all articles missing one or more of the required attributes of table 1. Then all articles with dummy values were identified and discarded. Dummy values are often introduced due to internal process requirements but do not add any relevant information to the data. If for example the attribute weight has to be filled for an article during article generation in order to proceed to the next step but the actual value is not know, often a dummy value of 1 or 999 is entered. These values distort the prediction model when used as training data in the machine learning step. The product name is subjected to lower casing and substitution of special German characters like umlauts. Special symbolic characters like #,! or separators are also deleted. With this pre-processing done the data is ready to be used for feature engineering.

Following this formal data cleaning we perform an additional content-focused pre-processing. The feature weight is discretized by binning it with bin width 10g. Volume is likewise treated with bin size 10ml. This simplifies the value distribution without rendering it too coarse. All articles where length is not equal to width are removed, because in these cases there are no single items but packages of items.

## 4 Feature Engineering

Often the data at hand is not sufficient to train a meaningful prediction model. In these cases feature engineering is a promising option. Identifying and engineering new features depends heavily on expert knowledge of the application domain. The first level consists of engineered features which can be determined by the original features alone. In the next step advanced second level features are computed from these first level and the original features.

For our data set the original features are product name and group as well as the dimensions length, width, height and volume. We see that the volume is computed in the most general way by multiplication of the dimensions. Geometrically this corresponds to all products being modelled as cuboids. Since angular beer or wine bottles are very much the exception in the real world, a sensible new feature would be a more appropriate modelling of the bottle shape. Since weight is closely correlated to volume, the better the volume estimate the better the weight estimate. To this end we propose four first level engineered features: capacity, wine bottle type, beer packaging type and beer bottle type which are in turn used to compute a second level engineered feature namely the packaging specific volume. Figure 1 shows all discussed features and their interdependencies.

## 4.1 Capacity

Let us have a closer look at the first level engineered features. The capacity of a beverage states the amount of liquid contained and is usually limited to a few discrete values. 0.33l and 0.5l are typical values for beer cans and bottles while wines are almost exclusively sold in 0.75l bottles and sometimes in 0.375l bottles. The capacity can be estimated from the given volume with sufficient certainty using appropriate threshold values. Outliers were removed from the data set.

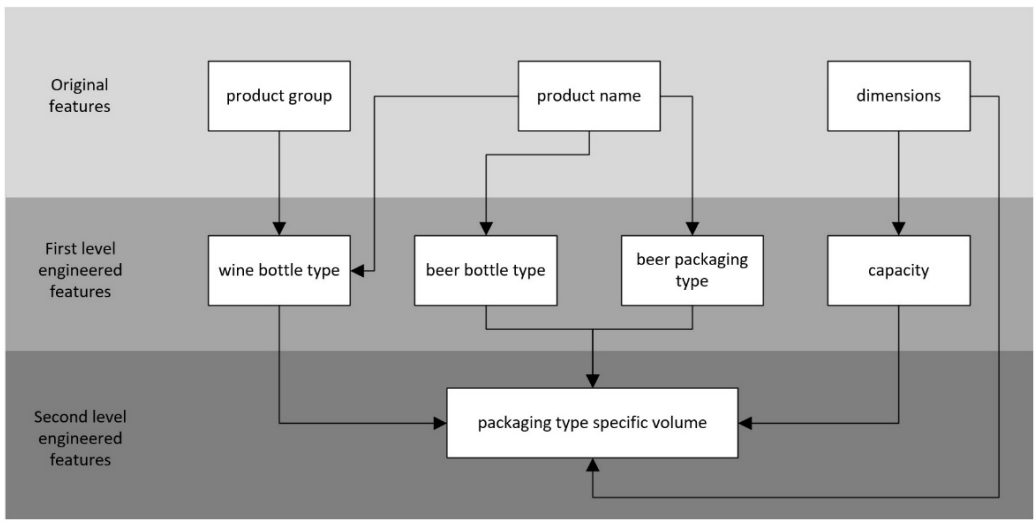

**Figure. 1.** First and second level engineered features

## 4.2 Beer packaging and bottle types

There are three main beer packaging types in retail: cans, bottles and kegs. While kegs are mainly of interest to pubs and restaurants and are not considered in this paper, cans and bottles target the typical super market shopper and come in a greater variety. In our data set, the product name in case of beers is preceded by a prefix denoting whether the product is packaged in a can  or a bottle. Extracting the relevant information is done using regular expressions. Not, though, that the prefix is not always correct and needs to be checked against the dimensions.

The shapes of cans are the same for all practical purposes, no matter the capacity. The only difference is in their wall thickness, which depends on the material, aluminium and tin foil  being the two common ones. The difference is weight is small and the actual material used is impossible to extract from the data. A further distinction for cans in different types like for beer and wine is therefore unnecessary.

Regarding the German beer market, the five bottle types shown in figure 2 are pre-dominant: longneck, NRW, Euro, Steini and NRW-Vichy. While not conforming to a DIN (the DIN 6199 Packmittel – Flaschen, Steinieform was withdrawn in 1997) standard anymore, the variations in length, width and height and weight are small. Therefore, bottles of the same type and brand may have different weights, even if they are sold in the same crate. Measurements for a crate of 18 filled Steini bottles resulted in 20 different weight values. A tolerance of +/- 2% was observed. As a pre-processing step, the weights were therefore rounded to the nearest 10g.

This holds true for all of the five bottle types in question. Since for beers there are no comparably meaningful reference points like color, grape or region of origin, the bottle type is determined by height and diameter using the minimum of the Euclidean distance to a set of idealized bottles with dimensions given in table 2. Bottles which were differing by more than 3% from these dimensions were removed from the data set.

| Bottle type | Capacity | Height | Diameter |
|---|---|---|---|
| NRW | 500 ml | 260.0 | 67.0 |
| Euro | 500 ml | 230.0 | 70.6 |
| Longneck | 500 ml | 270.0 | 68.3 |
| Longneck | 330 ml | 238.0 | 60.0 |
| NRW-Vichy | 330 ml | 233.0 | 61.0 |
| Steini | 330 ml | 174.0 | 70.0 |

**Table. 2.** Beer bottle types and corresponding typical dimensions

The engineered feature *beer packaging type* assigns each article identified as beer by its product group to one of the classes bottle or can. The feature beer bottle type contains the most probably member of the five main beer bottle types. Packages containing more than one bottle or can like crates or six packs are not considered in this paper and were removed from the data set.

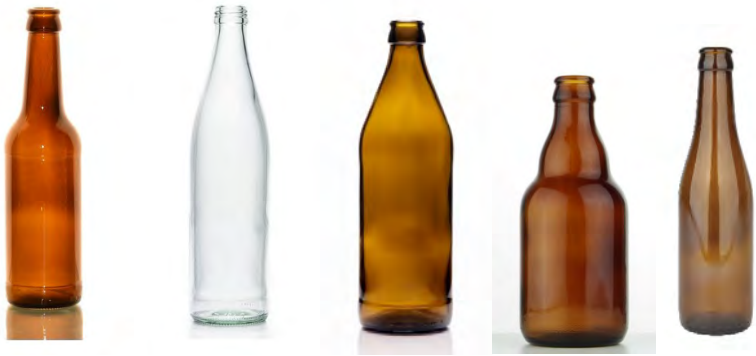

**Figure 2.** Beer bottle types, from left to right: Longneck, NRW, Euro, Steini, NRW-Vichy

## 4.3 Wine packaging types

Compared to beer the variety of commercially sold wine packagings is limited to bottles only. A corresponding packaging type attribute to distinguish between cans and bottles is not necessary. Again there are a few bottle types which are used for the majority of wines, namely Schlegel, Bordeaux and Burgunder (Figure 3). Deciding what product is filled in which bottle type is a question of domain knowledge. The original data set does not contain a corresponding feature. From the product group the country of origin and in the case of German wines the region can be determined via a mapping table. This depends on the type of product classification system the respective company uses and has not to be valid for all companies. Our data set uses a customer specific classification with focus on Germany. A more general one would be the Global Product Classification (GPC) standard for example. To determine wine growing regions in non-German countries like France the product name has to be analyzed using regular expressions. The type of grape is likewise to be deduced from the product name if possible. Using the country and specifically the region of origin and type of grape of the wine in question is the only way to assign a bottle type with acceptable certainty. There are countries and region in which a certain bottle type is used predominantly, sometimes also depending on the color of the wine. The Schlegel bottle, for example, is almost exclusively used for German and Alsatian white wines and almost nowhere else. Bordeaux and Burgunder bottles on the other hand are used throughout the world. Some countries like California or Chile use a mix of bottle types for their wines, which poses an additional challenge. With expert knowledge one can assign regions and grape types to the different bottle types. As with beer bottles this categorization is by no means comprehensive or free of exceptions but serves as a first step.

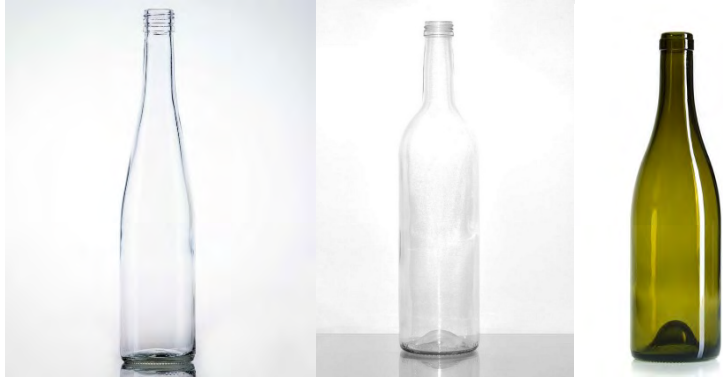

**Figure 2.** Wine bottle types, from left to right: Schlegel, Bordeaux, Burgunder

## 4.4 Packaging type specific volume computation

The standard volume computation by multiplying the product dimensions length, width and height is a rather coarse cuboid approximation to the real shape of alcoholic beverage packagings. Since the volume is intrinsically linked to the weight which we want to predict a packaging type specific volume computation is required for cans and especially bottles.

The modelling of a can is straightforward using a cylinder with the given height $h$ and a diameter of the given width $w$ and length $l$. Thus the packaging type specific volume is:

$$V_{Can}^{C} = 2\pi \cdot l \cdot w \cdot h \tag{1}$$

A bottle on the other hand needs to be modelled piecewise. Its height can be divided into three parts: base, shoulders and neck as shown in figure 4. Base and neck can be modeled by a cylinder. The shoulders are approximated by a truncated cone. With the help of the corresponding partial heights $h_{Base}$, $h_{Shoulders}$ and $h_{Neck}$ we can compute coefficients $k_{Base}$, $k_{Shoulders}$ and $k_{Neck}$ as fractions of the overall height $h$ of the bottle. The diameters of the bottle base and the neck opening are given by $d_{Base}$ and $d_{Opening}$ and are likewise used to compute the ratio $k_{opening}$. Since bottles have circular bases, the values for width $w$ and length $l$ in the original data have to be the same and either one may be used for $d_{Base}$. These four coefficients are characteristic for each bottle type, be it beer or wine (table 3). With their help, a bottle type specific volume from the original data length, width and height can be computed which is a much better approximation to the true volume than the former cuboid model.

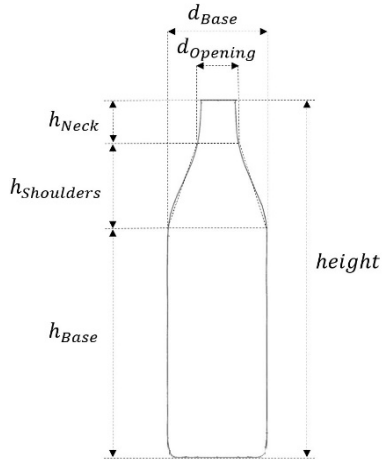

**Figure 4.** General parametric bottle description

| Coefficient | Burgunder | Bordeaux | Schlegel | NRW | Longneck | Steini | Euro |
|---|---|---|---|---|---|---|---|
| $k_{Base}^{Btype}$ | 0.48 | 0.60 | 0.34 | 0.59 | 0.52 | 0.52 | 0.62 |
| $k_{Shoulders}^{Btype}$ | 0.32 | 0.13 | 0.44 | 0.28 | 0.13 | 0.29 | 0.34 |
| $k_{Neck}^{Btype}$ | 0.20 | 0.27 | 0.22 | 0.13 | 0.35 | 0.19 | 0.24 |
| $k_{Opening}^{Btype}$ | 0.40 | 0.40 | 0.40 | 0.37 | 0.37 | 0.43 | 0.35 |

**Table 3.** Empirically measured coefficients for different bottle types

The bottle base can be modelled as a cylinder as follows:

$$V_{Base}^{C} = 2\pi \cdot l \cdot w \cdot h \cdot k_{Base}^{Btype} \tag{2}$$

The bottle shoulders have the form of a truncated cone and are described by formula 3:

$$V_{Shoulders}^{C} = \frac{1}{12}\pi \cdot h \cdot k_{Shoulders}^{Btype} \cdot l \cdot \left( l + l \cdot k_{Opening}^{Btype} + l \cdot k_{Opening}^{Btype\,2} \right) \tag{3}$$

The bottle neck again is a simple cylinder:

$$V_{Neck}^{C} = 2\pi \cdot l \cdot w \cdot k_{Opening}^{Btype} \cdot h \cdot k_{Neck}^{Btype} \tag{4}$$

Summing up all three sections yields the *packaging type specific volume* for bottles:

$$V_{Bottle}^{C} = V_{Base}^{C} + V_{Shoulders}^{C} + V_{Neck}^{C} \tag{5}$$

# 5    Experiments

The experiments follow the multi-level feature engineering scheme as shown in figure 1. First, we use only the original features product group and dimensions. Then we add the first level engineered features *capacity* and *bottle type* to the basic features. Next the second level engineered feature *packaging type specific volume* is used along with the basic features. Finally all features from every level are used for the prediction. After pre-processing and feature engineering the data set size is reduced from 3659 to 3380 beers and from 10212 to 8946 wines.

## 5.1    Algorithms and metrics

For prediction of the continuous valued attribute gross weight, we use and compare several regression algorithms. Both the decision-tree based Random Forests algorithm (Breimann, 2001) and support vector machines (SVM) (Cortes, 1995) are available in regression mode (Smola, 1997). Linear regression (Lai, 1979) and stochastic gradient descent (SGD) (Taddy, 2019) are also employed as examples of more traditional statics-based methods. Our baseline is a heuristic approach taking the median of the attribute gross weight for each product group and use this value as a prediction for all products of the same product group. Practical experience has shown this to be a surprisingly good strategy.

The implementation was done in Python 3.6 using the standard libraries sk-learn and pandas. All numeric features were logarithmized prior to training the models. The non-numeric feature *bottle type* was converted to numbers. The final results were obtained using tenfold cross validation (Kohavi, 1995). For model training 80% of the data was used while the remaining 20% constituted the test data.

We used the root mean square error (RSME) (6) as well as the mean and variance of the absolute percentage error $e_{ap}$ (7) as metrics for the evaluation of the performance of the algorithms.

$$RSME = \sqrt{\frac{\sum_{n=1}^{N}\left(v_n^{predicted} - v_n^{true}\right)^2}{N}} \tag{6}$$

$$e_{ap} = \frac{1}{N}\sum_{n=1}^{N}\frac{\left|v_n^{predicted} - v_n^{true}\right|}{V_{true}} \tag{7}$$

## 5.2    Results

| Features | Algorithm | RSME | Mean $e_{ap}$ | Var $e_{ap}$ |
|---|---|---|---|---|
| Product group | Baseline | 216.40 | 14.28 | 319.00 |
| Dimensions, product group | Linear Regression | 135.12 | 8.89 | 114.78 |
| Dimensions + capacity | Linear Regression | 142.97 | 9.19 | 137.32 |
| Dimensions + bottle_type | Linear Regression | 152.39 | 9.02 | 127.72 |
| Dimensions + bottle_type + capacity | Linear Regression | 136.45 | 8.19 | 115.92 |
| Dimensions + packaging_type_specific_volume | Linear Regression | 181.30 | 8.33 | 114.12 |
| Dimensions + all engineered features | Linear Regression | 133.91 | 8.08 | 122.09 |
| Dimensions, product group | SGD | 142.81 | 10.10 | 124.41 |
| Dimensions + capacity | SGD | 124.68 | 8.09 | 129.19 |
| Dimensions + bottle_type | SGD | 143.38 | 10.39 | 105.47 |
| Dimensions + bottle_type + capacity | SGD | 176.30 | 8.96 | 142.60 |
| Dimensions + packaging_type_specific_volume | SGD | 135.59 | 10.34 | 142.80 |
| Dimensions + all engineered features | SGD | 154.62 | 9.37 | 138.82 |
| Dimensions, product group | Random Forest | **91.73** | 6.72 | 121.45 |
| Dimensions + capacity | Random Forest | **98.75** | **4.98** | **106.12** |
| Dimensions + bottle_type | Random Forest | **91.75** | **4.59** | **80.94** |
| Dimensions + bottle_type + capacity | Random Forest | **83.25** | **3.76** | **71.17** |
| Dimensions + packaging_type_specific_volume | Random Forest | **111.72** | 5.68 | 113.28 |
| Dimensions + all engineered features | Random Forest | **99.93** | **3.62** | 81.77 |
| Dimensions, product group | SVM | 114.48 | **5.36** | **95.73** |
| Dimensions + capacity | SVM | 118.00 | 5.08 | 160.38 |
| Dimensions + bottle_type | SVM | 140.16 | 5.13 | 88.88 |
| Dimensions + bottle_type + capacity | SVM | 143.62 | 5.63 | 94.00 |
| Dimensions + packaging_type_specific_volume | SVM | 118.87 | **3.88** | **94.16** |
| Dimensions + all engineered features | SVM | 125.67 | 3.99 | **80.47** |

**Table. 4.** Weight prediction result metrics for beer with different feature combinations

All machine learning algorithms deliver significant improvements regarding the observed metrics compared to the heuristic median approach. The best results for each feature combination are highlighted in bold script. The results for the beer data set in table 4 show that the RSME can be more than halved, the mean $e_{ap}$ almost be reduced to a third and the variance of $e_{ap}$ quartered compared to the baseline approach. The Random Forest regressor achieves the best results in terms of RSME and $e_{ap}$ for almost all feature combinations except basic features and basic features combined with the packaging type specific volume, in which cases Support Vector Machines prove superior. Linear regression and SGD are are still better than the baseline approach but not on par with the other algorithms. Linear regression shows the tendency to improved results when successively adding features. SGD on the other hand exhibits no clear relation between number and level of features and corresponding prediction quality. A possible cause could be the choice of hyper parameters. SGD is very sensitive in this regard and depends more heavily upon a higher number of correctly adjusted hyper parameters than the other algorithms we used. Random Forests is a method which is very well suited to problems, where there is no easily discernable relation between the features. It is prone to overfitting, though, which we tried to avoid by using 20% of all data as test data. Adding more engineered features leads to increasingly better results using Random Forest with an outlier for the *packaging type specific volume* feature. SVM are not affected by only first level engineered features but profit from using the bottle type specific volume.

| Features | Algorithm | RSME | Mean $e_{ap}$ | Var $e_{ap}$ |
|---|---|---|---|---|
| Product group | Baseline | 38931.0 | 13.10 | 1125.00 |
| Dimensions, product group | Linear Regression | **30653.4** | 7.55 | 144.10 |
| Dimensions + capacity | Linear Regression | 36168.2 | 7.41 | 121.60 |
| Dimensions + bottle_type | Linear Regression | 37132.2 | 7.56 | 143.80 |
| Dimensions + bottle_type + capacity | Linear Regression | 40533.5 | 7.40 | 120.50 |
| Dimensions + packaging_type_specific_volume | Linear Regression | 42894.9 | 7.48 | **125.47** |
| Dimensions + all engineered features | Linear Regression | 34322.8 | 7.54 | 134.60 |
| Dimensions, product group | SGD | 34079.4 | 8.12 | 131.04 |
| Dimensions + capacity | SGD | **32039.8** | 7.83 | **95.69** |
| Dimensions + bottle_type | SGD | **32448.1** | 10.31 | 137.76 |
| Dimensions + bottle_type + capacity | SGD | 43545.4 | 11.69 | 178.17 |
| Dimensions + packaging_type_specific_volume | SGD | 37905.3 | 8.43 | 144.81 |
| Dimensions + all engineered features | SGD | **31762.2** | 11.57 | 146.07 |
| Dimensions, product group | Random Forest | 30999.9 | **6.98** | 121.08 |
| Dimensions + capacity | Random Forest | 38385.0 | **6.84** | 124.00 |
| Dimensions + bottle_type | Random Forest | 34032.1 | **6.65** | **118.17** |
| Dimensions + bottle_type + capacity | Random Forest | 39852.3 | **6.45** | **107.87** |
| Dimensions + packaging_type_specific_volume | Random Forest | 35509.5 | **7.18** | 143.60 |
| Dimensions + all engineered features | Random Forest | 34986.3 | **6.67** | **118.83** |
| Dimensions, product group | SVM | 39030.4 | 7.07 | **102.15** |
| Dimensions + capacity | SVM | 38132.8 | 7.18 | 118.53 |
| Dimensions + bottle_type | SVM | 38843.5 | 7.18 | 128.90 |
| Dimensions + bottle_type + capacity | SVM | **30109.9** | 7.23 | 140.37 |
| Dimensions + packaging_type_specific_volume | SVM | **35044.4** | 7.47 | 159.21 |
| Dimensions + all engineered features | SVM | 38993.4 | 7.35 | 135.40 |

**Table. 5.** Weight prediction result metrics for wine with different feature combinations

Regarding the wine data set the results depicted in table 5 are not as good as for the beer data set though still much better than the baseline approach. A reduction of the RSME by over 29% and of the mean $e_{ap}$ by almost 50% compared to the baseline were achieved. The variance of $e_{ap}$ could even be limited to under 10% of the baseline value. Again Random Forests is the algorithm with the best $e_{ap}$ metrics. Linear regression and SVM are comparable in terms of $e_{ap}$ while SGD is worse but shows good RSME values. In conclusion the general results of the wine data set show not much improvement when applying additional engineered features.

# 6 Discussion and Conclusion

The experiments show a much better predicting quality for beer than for wine. A possible cause could be the higher weight variance in bottle types compared to beer bottles and cans. It is also more difficult to correctly determine the bottle type for wine, since the higher overlap in dimensions does not allow to compute the bottle type with the help of idealized bottle dimensions. Using expert knowledge to assign the bottle type by region and grape variety seems not to be as reliable, though. Especially with regard to the lack of a predominant bottle type in the region with the most bottles (red wine from Baden for example), this approach should be improved. Especially Bordeaux bottles often sport an indentation in the bottom, called a 'culot de bouteille'. The size and thickness of this indentation cannot be inferred from the bottle's dimensions. This means that the relation between bottle volume and weight is skewed compared to other bottles without these indentations, which in turn decreases prediction quality.

Predicting gross weights with machine learning and domain-specifically engineered features leads to smaller discrepancies than using simple heuristic approaches. This is important for retail companies since big deviations are much worse for logistical reasons than small ones which may well be within natural production tolerances for bottle weights. Our method allows to check manually generated as well as data pool imported product data for implausible gross weight entries and proposes suggestion values in case of missing entries.

The method we presented can easily be adapted to non-alcoholic beverages using the same engineered features. In this segment, plastics bottles are much more common than glass ones and hence the impact of the bottle weight compared to the liquid weight is significantly smaller. We assume that this will cause a smaller importance of the bottle type feature in the prediction. A more problematic kind of beverage is liquor. Although there are only a few different standard capacities, the bottle types vary so greatly, that identifying a common type is almost impossible. One of the main challenges of our approach is determining the correct bottle types. Using expert knowledge is a solid approach but cannot capture all exemptions. Especially if a wine growing region has no predominant bottle type and is using mixed bottle types instead. Additionally many wine growers use bottle types which haven't been typical for their wine types because they want to differ from other suppliers in order to get the customer's attention. Assuming that all Rieslings are sold in Schlegel bottles, for example, is therefore not exactly true. One option could be to model hybrid bottles using a weighted average of the coefficients for each bottle type in use. If a region uses both Burgunder and Bordeaux bottles with about equal frequency, all products from this region could be assigned a hybrid bottle with coefficients computed by the mean value of each coefficient. If an initially bottle type labeled data set is available, preliminary simulations have shown that most bottle types can be predicted robustly using classification algorithms. The most promising strategy, in our opinion, is to learn the bottle types directly from product images using deep neural nets for example. With regard to the ever increasing online retail sector, web stores need to have pictures of their products on display, so the data is there to be used.

# AI-Powered Analysis of Industrial CT Data

Tim Schanz[1], Robin Tenscher-Philipp[2], Martin Simon[3]

[1] Tim Schanz (M.Sc)
**tim.schanz@hs-karlsruhe.de**
[2] Robin Tenscher-Phlipp (M.Sc)
**robin.tenscher-philipp@hs-karlsruhe.de**
[3] Martin Simon (Prof. Dr.-Ing.)
**martin.simon@hs-karlsruhe.de**

Hochschule Karlsruhe - Technik und Wirtschaft
University of Applied Sciences
Fakultät für Maschinenbau und Mechatronik
Moltkestr. 30
76133 Karlsruhe

**Abstract.**

Extensive research and development has been conducted in the field of AI-powered analysis of medical CT data during the past years – with significant progress. Although voxel data from industrial parts differ from medical data in the contrast level and the resolution, the medical DL approach is promising. Therefore, the aim of this work is exploring and developing neural network models for detecting defects in industrial CT data. Network architectures, successfully applied to medical CT data, were investigated and derivates were developed. Different neural network models were trained utilising a mixture of synthetic data and real data. The evaluation showed very good results for a modified U-Net neural network.

**Keywords:** AI; CT; INDUSTRIAL, CNN, DEEP LEARNING

## 1 Introduction

Quality assurance is one of the key issues for modern production technologies. Especially new production methods like additive manufacturing and composite materials require high resolution 3D quality assurance methods. Computed tomography (CT) is one of the most promising technologies to acquire material and geometry data non-destructively at the same time.

With CT it is possible to digitalize subjects in 3D, also allowing to visualize their inner structure. A 3D-CT scanner produces voxel data, comprising of volumetric pixels that correlate with material properties. The voxel value (grey value) is approximately proportional to the material density. Nowadays it is still common to analyse the data by manually inspecting the voxel data set, searching for and manually annotating defects. The drawback is that for high-resolution CT data, this process it very time consuming and the result is operator-dependent. Therefore, there is a high motivation to establish automatic defect detection methods.

There are established methods for automatic defect detection using algorithmic approaches. However, these methods show a low reliability in several practical applications. At this point artificial neural networks come into play that have been already implemented successfully in medical applications [1]. The most common networks, developed for medical data segmentation, are by Ronneberger et al., the U-Net [2] and by Milletari et al., the V-Net [3] and their derivates. These networks are widely used for segmentation tasks. Fuchs et al. describes three different ways of analysing industrial CT Data [4]. One of these contains a 3D-CNN. This CNN is based on the U-Net architecture and is shown in their previous paper [5]. The authors enhance and combine the U-Net and V-Net architecture to build a new network for examination of 3D volumes. In contrast, we investigate in our work how the networks introduced by Ronneberger et al. and Milletari et al. perform in industrial environments. Furthermore, we investigate if derivates of these architectures are able to identify small features in industrial CT data.

## 2    Industrial vs. medical CT data

In industrial CT systems, not only in the hardware design but also in the resulting 3D imaging data differs from medical CT systems. Voxel data from industrial parts differ from medical data in the contrast level and the resolution. State-of-the-art industrial CT scanner produce one to two order of magnitude larger data sets compared to medical CT systems. The corresponding resolution is necessary to resolve small defects. Medical CT scanners are optimised for a low x-ray dose for the patient, the energy of x-ray photons are typically up to 150 keV, industrial scanner typically use energies up to 450 KeV. In combination with the difference of the scan "object", the datasets differ significantly in size and image content.

To store volume data there are a lot of different file formats. Some of them are mainly used in medical applications like DICOM [6], NifTi[1] or RAW. In industrial applications VGL[3], RAW and TIFF[4] are commonly used. Also depending on the format, it is possible to store the data slice wise or as a complete volume stack.

## 3    CT data for training and evaluation

Industrial CT data, as mentioned in previous section, has some differences to medical CT data. One aspect is the size of the features to be detected or learned by the neural network. Our target is to find defects in industrial parts. As an example, we analyse pores in casting parts. These features may be very small, down to 1 to 7 voxels in each dimension. Compared to the size of the complete data volume (typically larger than 512 x 512 x 512 voxel), the feature size is very small. The density difference between material and pores may be as low as 2% of the maximum grey value. Thus, it is difficult to annotate the data even for human experts. The availability of real industrial data of good quality, annotated by experts, is very low. Most companies don't reveal their quality analysis data. Training a neural network with a small quantity of data is not possible. For medical applications, especially AI applications, there are several public datasets available. Yet these datasets are not always sufficient and researchers are creating synthetic medical data [7].

---

[1] Details available at: https://nifti.nimh.nih.gov/ - 20/03/12
[3] Details available at: https://www.volumegraphics.com/ - 20/03/12
[4] Details available at: https://kb.iu.edu/d/afjn - 20/03/12

Therefore, we decided to create synthetic industrial CT data. Another important reason for synthetic data is the quality of annotations done by human experts. The consistency of results is not given for different experts. Fuchs et al. have shown that training on synthetic data and predicting on real data lead to good results [4]. However, synthetic data may not reflect all properties of real data. Some of the properties are not obvious, which may lead to ignoring some varieties in the data. In order to achieve a high associability, we use a large numbers of synthetic data mixed with a small number of real data. To achieve this, we developed an algorithm which generates large amounts of data, containing a large variation of aspects, needed to generalize a neural network. The variation includes material density, pore density, pore size, pore amount, pore shape and size of the part.

There are some samples that could be learned easily, because the pores are clearly visible inside the material. However, some samples are more difficult to be learned, because the pores are nearly invisible. This allows us to generate data with a wide variety and hence the network can predict on different data. To train the neural networks, we can mix the real and synthetic data or use them separately. The real data was annotated manually by two operators.

To create a dataset of this volume we sliced it into 64x64x64 blocks. Only the blocks with a mean density greater than 50% of the grayscale range are used, to avoid too much empty volumes in the training data. Another advantage of synthetic data is the class balance. We have two classes, where 0 corresponds to material and surrounding air and 1 for the defects. Because of the size of the defects there is a high imbalance between the classes. By generating data with more features than in the real data, we could reduce the imbalance. Reducing the size of the volume to 64x64x64 also leads to better balance between the size of defects compared to full volume. In **Table 1** details of our dataset for training, evaluation and testing are shown. The synthetic data will not be recombined to a larger volume as they represent separate small components or full material units.

**Table 1:** Overview of used datasets.

| Name | Description | Resolution | No. of samples | No. of training samples | No. of evaluation samples | No. of test samples |
|------|-------------|------------|----------------|-------------------------|---------------------------|---------------------|
| Gdata | synthetic | 64x64x64 | 7249 | 6198 | 688 | 363 |
| Rdata | real | 64x64x64 | 156 | 135 | 15 | 6 |
| Mdata | mixed | 64x64x64 | 7405 | 6334 | 703 | 368 |

The following two slices of real data (**Figure 1**) and synthetic data (**Figure 2**) with annotated defects show the conformity between the data.

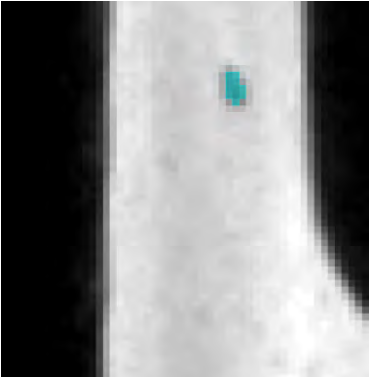 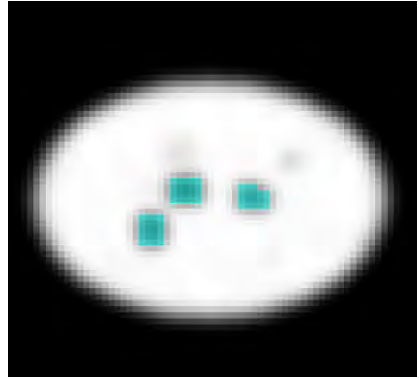

**Figure 1:** Sample slice of real data with size of 64x64x64 voxel.

**Figure 2:** Sample slice of synthetic data with size of 64x64x64 voxel.

## 4    Hardware and software setup

Deep learning (DL) consist of two phases: The training and its application. While DL models can be executed very fast, the training of the neural network can be very time-consuming, depending on several factors. One major factor is the hardware. The time consumed can be reduced by the factor of around ten when graphics cards (GPUs) are used. [8] To cache the training data, before it is given into the model, calculated on the GPU, a lot of random-access memory (RAM) is used [9] [10] [11]. Our system is built on a dual CPU hardware with 10 cores each running at 2.1 GHz and a Nvidia GPU Titan RTX[5] with 24GB of VRAM and 64GB of regular RAM. All measurements in this work concerning training and execution time are related to this hardware setup.

The operating system is Ubuntu 18.4LTS. Anaconda is used for python package management and deployment. The DL-Framework is Tensorflow[6] 2.1 and Keras as a submodule in Python[7].

## 5    Neural network architecture

Based on the 3DU-Net [12] and 3DV-Net [3] architecture compared from Paichao et al. [13] we created modified versions which differ in number of layers and their hyperparameters. Due to the small size of our data, no patch division is necessary. Instead the training is performed on the full volumes. We actually do not use the Z-Net enhancement proposed in their paper. The input size, depending on our data, is defined to 64x64x64x1 with 1 dimension for channel. The incoming data will be normalized. As we have a binary segmentation task, our output activation is the sigmoid [14] function. Based on Paichao et al. [13] the convolutional layer of our 3DU-Nets have a kernel size of (3, 3, 3) and the 3DV-Nets have a kernel size of (5, 5, 5). As convolution activation function we are using ELU [14] [15] and he_normal [16] as kernel initialization [17]. The ADAM optimisation method [18] [19] is used with a starting learning

---

rate of 0.0001, a decay factor of 0.1 and the loss function is the binary cross-entropy [20]. **Figure 3** shows a sample 3DU-Net architecture where downwards max pooling and upwards transposed convolution are used. Compared to **Figure 4**, the 3DV-Net, where we have a fully convolutional neural network, the descend is done with a (2, 2, 2) convolution and a stride of 2 and ascent with transposed convolution. It also has a layer level addition of the input of this level added to the last convolution output of the same level, as marked by the blue arrows. To adapt the shapes of the tensors for adding them, the down-convolution and the last convolution of the same level, have to have the same number of kernel filters.

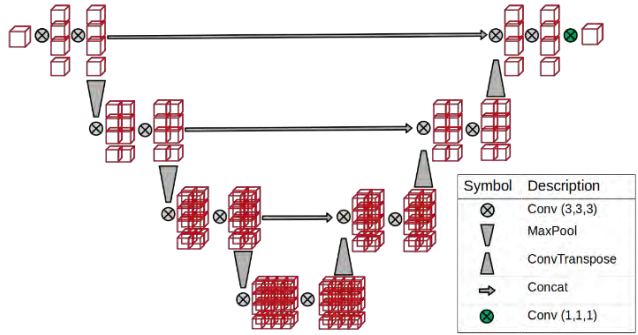

**Figure 3:** Sample U-Net architecture for building reference.

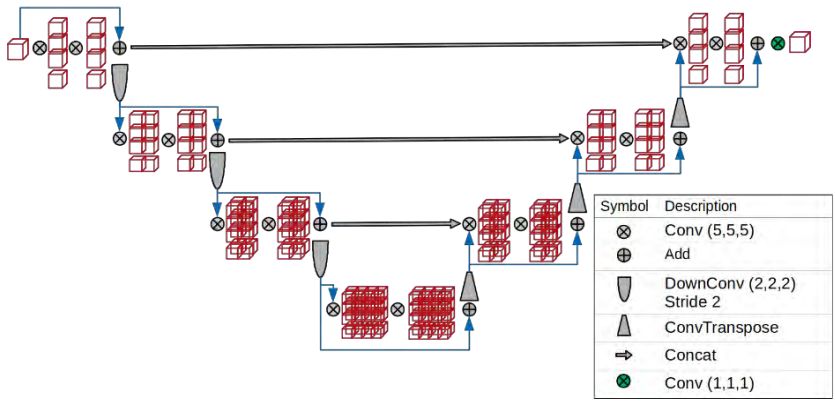

**Figure 4:** Sample V-Net architecture for building reference.

Our modified neural network differ in the levels of de-/ascending, the convolution filter kernel size and their hyperparameters, shown in **Table 2**. The convolutions on one level have the same number of filter kernel. After every down convolution the number of filters is multiplied by 2 and on the way up divided by 2.

**Table 2:** Tested models and their base specifications.

| Architecture | Starting Kernel Number | Kernel Size | Down/Up-Levels | Hyperparameter |
|---|---|---|---|---|
| U-Net_small | 16 | (3, 3, 3) | 3_BOTTOM_3 | $1,40 \times 10^6$ |
| U-Net_large | 16 | (3, 3, 3) | 5_BOTTOM_5 | $3,41 \times 10^7$ |
| V-Net_small | 16 | (5, 5, 5) | 3_BOTTOM_3 | $1,06 \times 10^7$ |
| V-Net_large | 16 | (5, 5, 5) | 5_BOTTOM_5 | $1,77 \times 10^8$ |

# 6    Training and evaluation of the neural networks

The conditions of a training and a careful parameters selection is important. In **Table 3** the training conditions fitted to our system and networks are shown. We are also taking into account that different network architectures and number of layers are better performing on different learning rates, batch size, etc.

**Table 3:** Training conditions.

| Parameter | Description | Value |
|---|---|---|
| Batchsize | Number of sample per iteration | 5 |
| Epochs | Number of iterations of all samples | 15-90 |
| Learning rate | Factor for weight adjustment | 0.001 |
| Shuffle data | Shuffle data before loading in batches | True |
| Learning rate decay | Reduction of learning rate when reaching plateau | 0.1 |

To evaluate our trained models, we are mainly focusing on the IoU metric, also called Jackard Index, which is the intersection over union. This metric is widely used for segmentation tasks and compares the intersection over union between the prediction and ground truth for each voxel. The value of IoU range between 0 and 1, whereas the loss values range between 0 and infinite. Therefore, the IoU is a much clearer indicator. An IoU close to 1 indicates a high intersection-precision between the prediction and the groundtruth. Our networks where trained between 30 and 90 epochs until no more improvement could be achieved. Both datasets consist of a similar number of samples, which means the epoch time is equivalent. One epoch took around 4 minutes.

**Figure 5** shows the loss determined based on the evaluation data. As described in Fehler! Verweisquelle konnte nicht gefunden werden.**,** all models are trained on and evaluated against the synthetic dataset Gdata and on the mixed dataset Mdata. In general, the loss achieved by all models is higher on Mdata because the real data is harder to learn. A direct comparison between the models is only possible between models with the same architecture. The IoU metric shown in **Figure 6**. Here the evaluation is sorted based on the IoU metric. If we compare the loss of UNET-Mdata with UNET-Gdata, which are nearly the same for Mdata, with their corresponding IoU (UNET-Mdata (~0.8) and UNET-Gdata (~0.93)), we can see that a lower loss does not necessarily lead to higher IoU score. If only the loss and IoU are considered, the UNets tend to be better than the VNets. As a conclusion, considering the IoU metric for model selection, the UNET-Gdata is the best performing model and VNET-Gdata the least performing.

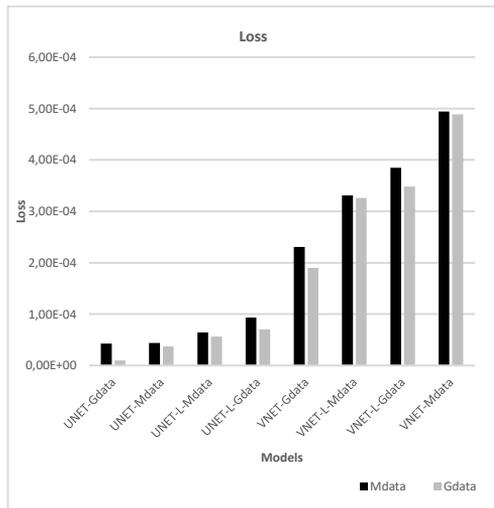

**Figure 5:** The evaluation loss determined based on the evaluation data sorted from lowest to highest.

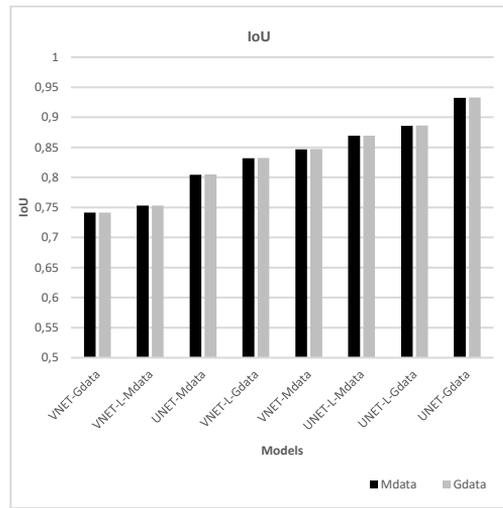

**Figure 6:** The evaluation IoU determined based on the evaluation data sorted from lowest to highest.

After comparing the automatic evaluation, we show prediction samples of different models on real and synthetic data (**Table 4**). Rows 1 and 2 show the comparison between UNET-Gdata and VNET-Gdata, predicting on a synthetic test sample. The result of UNET-Gdata exactly hits the groundtruth, whereas the VNET-Gdata prediction has a 100% overlap to the groundtruth but with surrounding false positive segmentations. In row 3 and 4 both models predict the groundtruth plus some false positive segmentations in the close neighbourhood. In row 5 and 6 the prediction results of the same two models on real data is shown, taking into account that both models are not trained on real data. UNET-Gdata delivers a good precision with some false positive segmentations in thegroundtruth area and one additional segmented defect. This shows that the model was able to find a defect which was missed by the expert. VNET-Gdata shows a very high number of false positive segmentations.

**Table 4:** Overview of predictions by different models on synthetic and real test samples.

| Model | Data type | Sample | Groundtruth | Prediction |
|---|---|---|---|---|
| UNET-Gdata | synthetic | 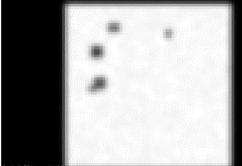 | 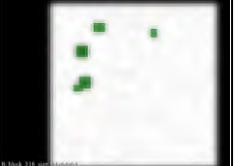 | 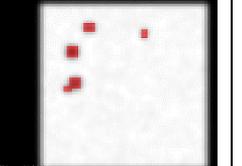 |
| VNET-Gdata | synthetic | 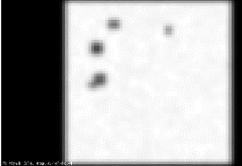 | 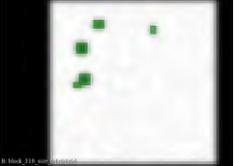 | 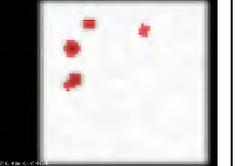 |
| UNET-Gdata | synthetic | 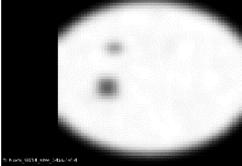 | 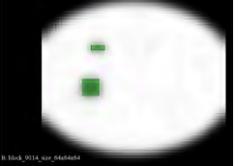 | 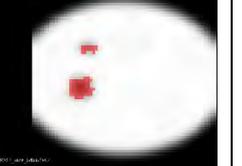 |
| VNET-Gdata | synthetic | 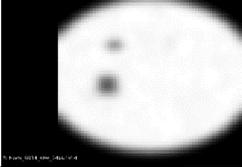 | 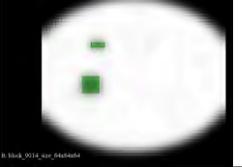 | 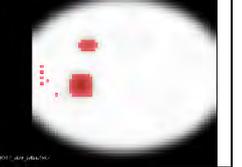 |
| UNET-Gdata | real | 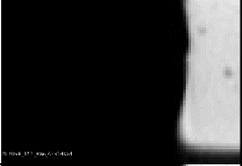 | 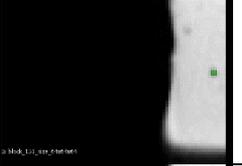 | 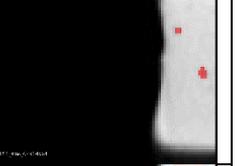 |
| VNET-Gdata | real | 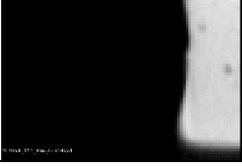 | 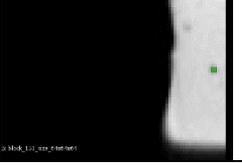 | 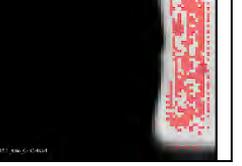 |

## 7 Conclusion

In this paper, we have proposed a neural network to find defects in real and synthetic industrial CT volumes. We have shown that neural networks, developed for medical applications can be adapted to industrial applications. To achieve high accuracy, we used a large variety of features in our data. Based on the evaluation and manually reviewing random samples we have chosen the UNET architecture for further research. This model achieved great performance on our real and synthetic dataset. In summery this paper shows that the artificial intelligence and their neural networks will take an import enrichment in industrial issues.

# Towards Classification and Prediction of Stress Patterns using Multiple Physiological Signals

Clarissa Almeida Rodrigues[1,] William da Rosa Fröhlich[1], Sandro José Rigo[1], Eliza Kern de Castro[1], Andreia Rodrigues[1], Rodrigo Marques Figueiredo[1], Ana Paula Mallmann[1]

[1] University of Vale do Rio dos Sinos
**clarissa.ar@gmail.com, williamrfrohlich@gmail.com, rigo@unisinos.br, elisa.kerndecastro@gmail.com, andreiakrodrigues@yahoo.com.br, marquesf@unisinos.br, apmallmann@unisinos.br**

**Abstract.** The stress is increasing in our society in the last years, due the large and tiring routines besides few time to rest. Keeping this in mind, this paper intends to determine patterns in stress' events using physiological signs, because these signals are a reliable source to identify stress states. The literature shows that the use of physiological signs as a source for stress patterns identification is a promising investigation subject and there are few studies evaluating the effect of combining several different signals. The objective of this article is to investigate the possible integration of data obtained from electrocardiographic (ECG), electrodermal activity (EDA) and electromyography (EMG) to detect stress patterns using wearable sensors to acquisition of biofeedback and propose algorithms to set some patterns. It was developed a dataset to made the pre-processing in all of data to evaluate the plausibility and develop an adequate database for the application of machine learning techniques establishing as a reference the obtained annotated data.

**Keywords:** Wearable sensors, Stress, Biofeedback.

## 1 Introduction

Stress can affect all aspects of our lives, including our emotions, behaviors, thinking ability, and physical health, making our society sick – both mentally and physically. Among the effects that the stress and anxiety can cause are heart diseases, such as coronary heart disease and heart failure [5]. Due this information, this research will present a proposal to help people handling stress using the benefit of technology development and to set patters of stress status as way to propose some intervention, once the first step to controlling stress is to know the symptoms of stress.

The stress symptoms are very board and can be confused with others diseases according The American Institute of Stress [15], for example the frequent headache, irritability, insomnia, nightmares, disturbing dreams, dry mouth, problems swallowing, increased or decreased appetite, or even cause other diseases such as frequent colds and infections. In view of the wide variety of symptoms caused by stress, this research intends to define, through physiological signals, the patterns generated by the body and obtained by wearable sensors and develop a standardized database to apply the machine learning.

## 2 Problem and Motivation

According to a research of The American Institute of Stress [15], 77% of people regularly experience physical symptoms caused by stress, 51% of them are related to fatigue. In other hand, advances in sensor technology, wearable devices and mobile growth would help to online stress identification based on physiological signals and delivery of psychological interventions. Currently with the advancement of technology and improvements in the wearable sensors area, made it possible to use these devices as a source of data to monitor the user's physiological state. The majority of the wearable devices consist of low-cost board that can be used to the acquisition of physiological signals [1, 10]. After the data are obtained it is necessary apply some filters to clear signal, without noise or distortions aiming to use some Machine Learning approaches to model and predict these stress states [2, 11].

The wide-spread use of mobile devices and microcomputers, as Raspberry Pi, and its capabilities presents a great possibility to collect, and process those signs with an elaborated application. These devices can collect the physiological signals and detect specific stress states to generate interventions following the predetermined diagnosis based on the standards already evaluated in the system [9, 6]. During the literature review it was evident the presence of few works dedicated to evaluating comprehensively the complete cycle of biofeedback, which comprises using the wearable devices, applying Machine Learning patterns detection algorithms, generate the psychologic intervention, besides monitoring its effects and recording the history of events [9, 3].

## 3   Background and Related Works

Stress is identified by professionals using human physiology, so wearables sensors could help on data acquisition and processing, through machine learning algorithms on biosignal data, suggesting psychological interventions. Some works [6, 14] are dedicated to define patterns as experiment for data acquisition simulation real situations. Jebelli, Khalili and Lee [6] showed a deep learning approach that was used to compare with a baseline feedforward artificial neural network.

Schmidt et al. [12] describes Wearable Stress and Affect Detection (WESAD), one public dataset used to set classifiers and identify stress patterns integrating several sensors signals with the emotion aspect with a precision of 93% in the experiments. The work of Gaglioli et al. [4] describe the main features and preliminary evaluation of a free mobile platform for the self-management of psychological stress.

In terms of the wearables, some studies [13, 14] evaluate the usability of devices to monitory the signals and the patient's well-being. Pavic et al. [13] showed a research performed to monitor cancer patients remotely and as the majority of the patients have a lot of symptoms but cannot stay at hospital during all treatment. The authors emphasize that was obtained good results and that this system is viable, as long as the patient is not a critical case, as it does not replace medical equipment or the emergency care present in the hospital.

Henriques et al. [5] focus was to evaluated the effects of biofeedback in a group of students to reduce anxiety, in this paper was monitored the heart rate variability with two experiments with duration of four weeks each. The work of Wijman [8] describes the use of EMG signals to identify stress, this experiment was conducted with 22 participants, evaluating both the wearables signals and questionnaires.

## 4   Approach and Uniqueness

In this section will be described the uniqueness of this research and the devices that was used. This solution is being proposed by several literature study about stress patterns and physiological aspects but with few results, for this reason, our project will address topics like experimental study protocol on signals acquisition from patients/participants with wearables to data acquisition and processing, in sequence will be applied machine learning modeling and prediction on biosignal data regarding stress (Fig. 1).

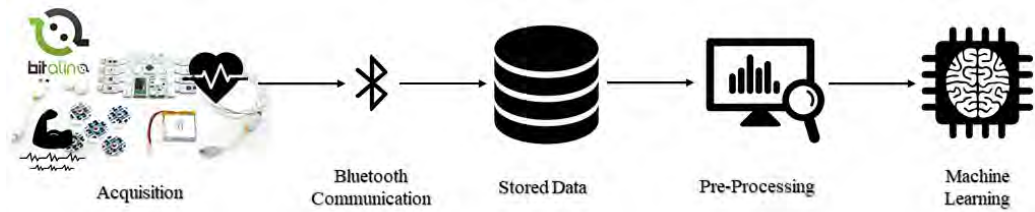

**Fig. 1.** The acquisition system

The protocol followed to the acquisition of signals during all different status is the Trier Social Stress Test (TSST) [7], recognized as the gold standard protocol for stress experiments. The estimated total protocol time, involving pre-tests and post-tests, is 116 minutes with a total of thirteen steps, but applied experiment was adapted and it was established with ten stages: Initial Evaluation: The participant arrives, with the scheduled time, and answer the questionnaires; Habituation: It will take a rest time of twenty minutes before the pre-test to avoid the influence of events and to establish a safe baseline of that organism; Pre-Test: The sensors will be allocated (Fig. 2), collected saliva sample and applied the psychological instruments.

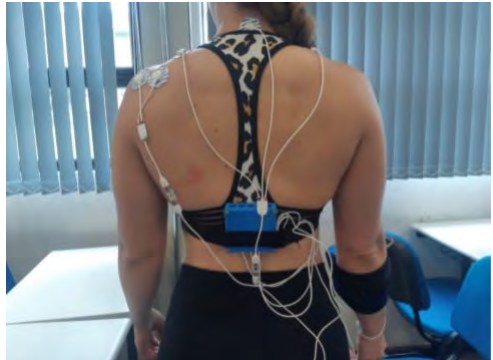

**Fig. 2.** Sensors Allocated

The next step is Explanation of procedure and preparation: The participant reads the instructions and the researcher ensures that he understands the job specifications, in sequence, he is sent to the room with the jurors (Fig. 3), composed of two collaborators of the research, were trained to remain neutral during the experiment, not giving positive verbal or non-verbal feedback; Free Speech: After three minutes of preparation, the participant is requested to start his speech, being informed that he cannot use the notes.

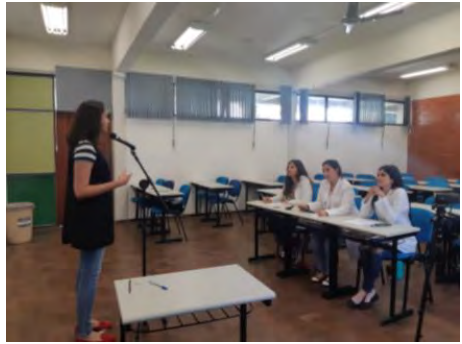

**Fig. 3.** Participant and the jurors

This will follow the Arithmetic Task: the jurors request an arithmetic task in which the participant must subtract mentally, sometimes, the jurors interrupt and warn that the participant has made a mistake; Post-Test Evaluation: The experimenter receives the subject outside the room for the post-test evaluations; Feedback and Clarification: The investigator and jurors talk to the subject and clarify what the task was about; Relaxation technique: A recording will be used with the guidelines on how to perform a relaxation technique, using only the breathing; Final Post-Test: Some of the psychological instruments will be reapplied, saliva samples will be collected, and the sensors will still be picking up the physiological signals.

Based on literature [14] and wearable devices available the signals that was selected to analysis is the ECG, EDA and EMG for an initial experiment. This experimental study protocol on data acquisition started with 71 participants, where data annotation each step was done manually, from protocol experiment, preprocessing data based on features selection. In the Machine Learning step, it was evaluated the metrics of different algorithms as Decision Tree, Random Forest, AdaBoost, KNN, K-Means, SVM.

The experiment was made using the BITalino Kit - PLUX Wireless Biosignals S.A. (Fig. 4) composed by ECG sensor, which will provide data on heart rate and heart rate variability; EDA sensor that will allow measure the electrical dermal activity of the sweat glands; EMG sensor that allows the data collect the activity of the muscle signals.

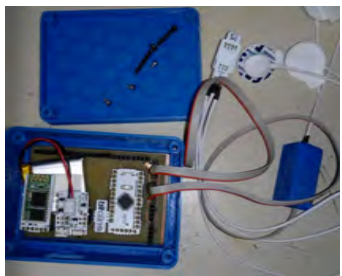

**Fig. 4.** BeWell prototype

## 5 Results

This section will describe the results in the pre-processing step and how it was made, listing all parts regarded to categorization and filtering data, evaluating the signal to know if it has plausibility and create a standardized database. The developed code is written in Python due to the wide variety of libraries available, in this step was used the libraries NumPy and Pandas, both used to data manipulation and analysis.

In the first step it is necessary read the files with the raw data and the timestamp, during this process the used channels are renamed to the name of the signal, because the BITalino store the data with the channel number as name of each signals. In sequence, the data timestamp is converted to a useful format, with goal to compare with the annotations, after time changed to the right format all channels unused are discarded to avoid unnecessary processing. The next step is to read the annotations taken manually in the experiment, as said before, to compare the time and classify each part of the experiment with its respective signal.

After all signals are classified with its respective process of the TSST, each part of the experiment is grouped in six categories, which will be analyzed later. The first category is the "*baseline*", with just two parts of the experiment, representing the beginning of the experiment, when the participants had just arrived. The second is called of "*tsst*" comprises the period in which the participant spoke, the third category is the "*arithmetic*" with the data in acquired in the arithmetic test.

The others two relevant categories are the "*post_test_sensors_1*" and "*post_test_sensors_2*", with its respective signals in the parts called with the same name. Every other part of the experiment was categorized as "*no_category*", in sequence, this category is discarded in function of it will not be necessary in the machine learning stage. After the dataframe is right with all signals properly classified, the columns with the participants number and the timestamp are removed of the dataframe. The next step is evaluated the signal, to verify if the signal is really useful in the process of machine learning. For this, it is analyzed the signals using the BioSPPy library, which performs the data filtering process and makes it possible to view the data.

Finally, the script checks the volume of data present in each classification and returns the value of the smallest category. This is done because it was found that the categories have different volumes of data, which would become a problem in the machine learning stage, by offering more data from a determinate category than from the others. Due this fact, the code analyzes the others categories and reduce its size until all categories stay with the same number of rows in each category (); after this the dataframe is exported in a CSV file, to be read in the machine learning stage.

```
In [46]:  minimum_size = comparator(new_data)

          Baseline: 139230
          TSST: 73931
          Arithmetic: 346907
          Post Test Sensors I: 584527
          Post Test Sensors II: 741600
          73931

          Call the "data_generator" Method to Generate the File with Standardized Data

In [47]:  data_save = data_generator(new_data, minimum_size)

          Call the "comparator" Verify the DataFrame

In [48]:  comparator(data_save)

          Baseline: 73931
          TSST: 73931
          Arithmetic: 73931
          Post Test Sensors I: 73931
          Post Test Sensors II: 73931
          73931
```

**Fig. 5.** Standardization of data

# 6 Conclusion

The purpose of this article is to describe some stages of the development of a system for the acquisition and analysis of physiological signals to determine patterns in these signals that would detect stress states. During the development of the project was verified that there are data gaps in the dataframe in the middle of the experiment in some participants; A hypothesis about the motivation of this had happened is the sampling of the acquisition of BITalino regarding communication issues in some specifics sampling rates.

It evaluate the results obtained when reducing this acquisition rate, however, it is necessary to carefully evaluate the extent to which the reduction in the sampling rate will interfere with the results. During the evaluation of the plausibility of the signals, it was verified that there are evident differences between the signals patterns in the different stages of the process, thus validating the protocol followed in the acquisition of the standards. The next step in this project is implement the machine learning stage, applying different algorithms as SVM, Decision Tree, Random Forest, AdaBoost, KNN and K-Means; besides to evaluate the results using metrics like Accuracy, Precision, Recall and F1.

The next steps of this research will support the confirmation of the hypothesis raised about being able to define patterns of physiological signals to detect stress states. From the definition of the patterns, a system can be applied that identifies the acquisition of the signals and, in real time, performs the analysis of these data based on the machine learning results. Therefore we can detect the state of the person and that the psychologist can indicate a proposal intervention and monitor whether the decrease is occurring.

# Artificial Intelligence Assisted Creation
## *Fostering Inspiration & Raising Moral Issues*

Matthias Wölfel

Faculty of Computer Science and Business Information Systems
Karlsruhe University of Applied Sciences
`matthias.woelfel@hs-karlsruhe.de`

**Abstract.** The promise of *artificial intelligence* (AI), in particular its latest developments in deep learning, has been influencing all kinds of disciplines such as engineering, business, agriculture, and humanities. More recently it also includes disciplines that were exclusively reserved for humans such as art and design. While there is a strong debate going on if creativity is profoundly human, we investigate if creativity can be fostered by AI. To get a better understanding of the creative potential offered by AI we open the black box and investigate where and how the magic is happening. Besides the potentials of AI, we also point out and discuss ethical and social implications caused by the latest developments in AI with respect to the creative sector.

**Keywords:** inspirational AI; human-machine co-design; moral issues

## 1    Introduction

Technological developments have been influencing all kinds of disciplines by transferring more competences from human beings to technical devices. The steps inculde [1]:

1. *tools*: transfer of mechanics (material) from the human being to the device
2. *machines*: transfer of energy from the human being to the device
3. *automatic machines*[1]: transfer of information from the human being to the device
4. *assistants*: transfer of decisions from the human being to the device

With the introduction of *artificial intelligence* (AI), in particular its latest developments in deep learning, we let the system (in step 4) take over our decisions and creation processes. Thus, tasks and disciplines that were exclusively reserved for humans in the past can now co-exist or even take the human out of the loop. It is no wonder that this transformation is not stopped at disciplines such as engineering, business, agriculture but also affects humanities, art and design. Each new technology has been adopted for artistic expression—just see the many wonderful examples in media art. Therefore, it is not surprising, that AI is going to be established as a novel tool to produce creative content of any form. However, in contrast to other disruptive technologies, AI seems particular challenging to be accepted in the area of art because it offers capabilities we thought once only humans are able to perform—the art is no longer done by artists using new technology to perform their art, but the art is done by the machine itself without the need for a human to intervene. The question of "what is art" has always been an emotionally debated topic in which everyone has a slightly different definition depending on his or her own experiences, knowledge base and personal aesthetics. However, there seems to be a broad consensus that art requires *human creativity and imagination* as, for instance, stated by the Oxford dictionary "The expression or application of human creative skill and imagination, typically in a visual form such as painting or sculpture, producing works to be appreciated primarily for their beauty or emotional power."

---

[1] Automatic machine is called *Automat* or *automate* in other languages such as German or French respectively.

Every art movement challenges old ways and uses artistic creative abilities to spark new ideas and styles. With each art movement diverse intentions and reasons for creating the artwork came along with critics who did not want to accept the new style as an artform. With the introduction of AI into the creation process another art movement is trying to be established which is fundamentally changing the way we see art. For the first time, AI has the potential to take the artist out of the loop, to leave humans only in the positions of curators, observers and judges to decide if the artwork is beautiful and emotionally powerful.

## 2 Fostering Inspiration

While there is a strong debate going on in the arts if creativity is profoundly human, we investigate how AI can foster inspiration, creativity and produce unexpected results. It has been shown by many publications that AI can generate images, music and the like which can resemble different styles and produce artistic content. For instance, Elgammal et al. [2] have used *generative adversarial networks* (GAN) to generate images by learning about styles and deviating from style norms. The promise of AI-assisted creation is "a world where creativity is highly accessible, through systems that empower us to create from new perspectives and raise the collective human potential" as Roelof Pieters and Samim Winiger pointed out [3]. To get a better understanding of the process on how AI is capable to propose images, music, etc. we have to open the black box to investigate where and how the magic is happening.

### 2.1 Random & Constrained Variations

Random variations in the *image space* (sometimes also referred to as *pixel space*) are usually not leading to any interesting result. This is because semantic knowledge cannot be applied. Therefore, methods need to be applied which constrain the possible variations of the given dataset in a meaningful way. This can be realized by *generative design* or *procedural generation*. It is applied to generate geometric patterns, textures, shapes, meshes, terrain or plants. The generation processes may include, but are not limited, to self-organization, swarm systems, ant colonies, evolutionary systems, fractal geometry, and generative grammars. McCormack et al. [4] review some generative design approaches and discuss how art and design can benefit from those applications. These generative algorithms which are usually realized by writing program code are very limited. AI can change this process into data-driven procedures. AI, or more specifically artificial neural networks, can learn patterns from (labeled) examples or by reinforcement.

Before an artificial neural network can be applied to a task (classification, regression, image reconstruction), the general architecture is to extract features through many hidden layers. These layers represent different levels of abstractions. Data that have a similar structure or meaning should be represented as data points that are close together while divergent structures or meanings should be further apart from each other. To convert the image back (with some conversion/compression loss) from the low dimensional vector, which is the result of the first component, to the original input an additional component is needed. Together they form the *autoencoder* which consists of the *encoder* and the *decoder* . The *encoder* compresses the data from a high dimensional input space to a low dimensional space, often called the *bottleneck* layer. Then, the *decoder* takes this encoded input and converts it back to the original input as closely as possible. The *latent space* is the space in which the data lies in the bottleneck layer. If you look at Figure 1 you might be wondering why a model is needed that converts the input data into a "close as possible" output data. It seems rather useless if all it outputs is itself. As discussed, the latent space contains a highly compressed representation of the input data, which is the only information the decoder can use to reconstruct the input as faithfully as possible. The magic happens by interpolating between points and performing vector arithmetic between points in latent space. These transformations result in meaningful effects on the generated images. As dimensionality is reduced, information which is distinct to each image is discarded from the latent space representation, since only the most important information of each image can be stored in this low-dimensional space. The latent space captures the structure in your data and usually offers some semantic meaningful interpretation. This semantic meaning is, however, not given a priori but has to be discovered.

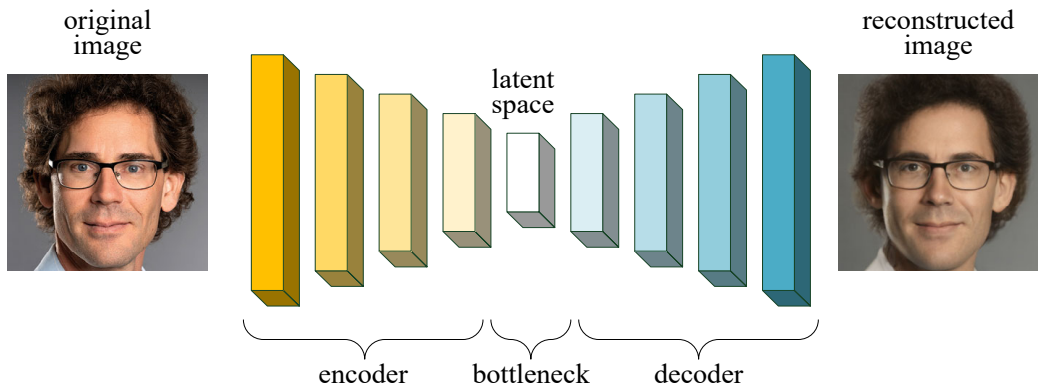

**Fig. 1.** The general architecture of an autoencoder.

## 2.2 Exploring the Latent Space

As already discussed autoencoders, after learning a particular non-linear mapping, are capable of producing photo-realistic images from randomly sampled points in the latent space. The latent space concept is definitely intriguing but at the same time non-trivial to comprehend. Although latent space means hidden, understanding what is happening in latent space is not only helpful but necessary for various applications. Exploring the structure of the latent space is both interesting for the problem domain and helps to develop an intuition for what has been learned and can be regenerated. It is obvious that the latent space has to contain some structure that can be queried and navigated. However, it is non-obvious how semantics are represented within this space and how different semantic attributes are entangled with each other.

To investigate the latent space one should favor a dataset that offers a limited and distinctive feature set. Therefore, faces are a good example in this regard because they share features common to most faces but offer enough variance. If aligned correctly also other meaningful representations of faces are possible, see for instance the widely used approach of *eigenfaces* [5] to describe the specific characteristic of faces in a low dimensional space.

In the latent space we can do vector arithmetic. This can correspond to particular features. For example, the vector $A_{smiling\ woman}$ representing the face of a smiling woman minus the vector $A_{neutral\ woman}$ representing a neutral looking woman plus the vector $A_{neutral\ man}$ representing a neutral looking man resulted in the vector $A_{smiling\ man}$ representing a smiling man.

$$A_{smiling\ woman} - A_{neutral\ woman} + A_{neutral\ man} = A_{smiling\ man}$$

This can also be done with all kinds of images; see e.g. the publication by Radford et al. [6] who first observed the vector arithmetic property in latent space. A visual example is given in Figure 2. Please note that all images shown in this publication are produced using BigGAN [7]. The photo of the author on which most of the variations are based on is taken by Tobias Schwerdt.

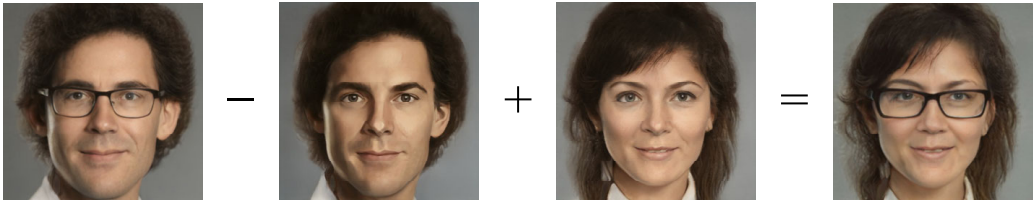

**Fig. 2.** In latent space, vector algebra can be carried out.

Semantic editing requires to move within the latent space along a certain 'direction'. Identifying the 'direction' of only one particular characteristic is non-trivial since editing one attribute may affect others because they are correlated. This correlation can be attributed to some extent to pre-existing correlations in 'the real world' (e.g. old persons are more likely to wear eyeglasses) or bias in the training dataset (e.g. more women are smiling on photos than men). To identify the semantics encoded in the latent space Shen et al. proposed a framework for interpreting faces in latent space [8]. Beyond the vector arithmetic property, their framework allows decoupling some entangled attributes (remember the aforementioned correlation between old people and eyeglasses) through linear subspace projection. Shen et al. found that in their dataset pose and smile are almost orthogonal to other attributes while gender, age, and eyeglasses are highly correlated with each other. Disentangled semantics enable precise control of facial attributes without retraining of any given model. In our examples, in Figures 3 and 4, faces are varied according to gender or age.

It has been widely observed that when linearly interpolate between two points in latent space the appearance of the corresponding synthesized images 'morphs' continuously from one face to another; see Figure 5. This implies that also the semantic meaning contained in the two images changes gradually. This is in stark contrast to having a simple fading between two images in image space. It can be observed that the shape and style slowly transform from one image into the other. This demonstrates how well the latent space understands the structure and semantics of the images. Other examples are given in Section 3.

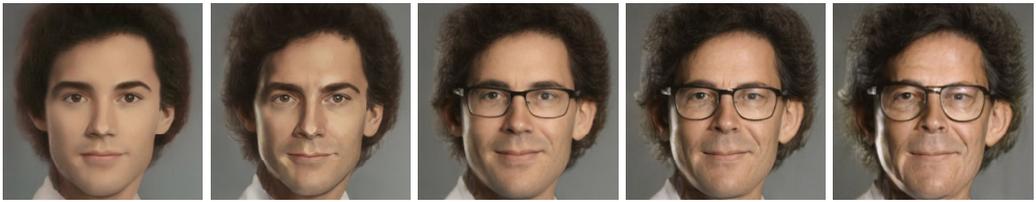

**Fig. 3.** The *age* of a person can be changed by moving from the location of young to old in the latent space.

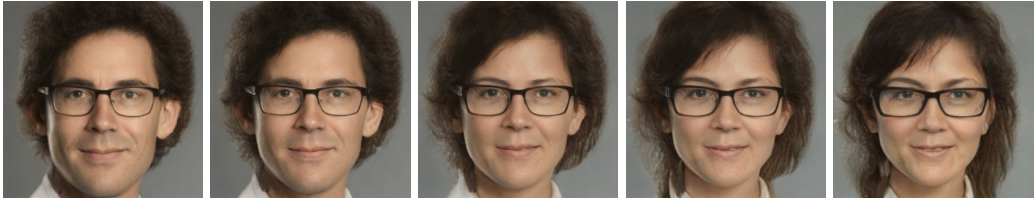

**Fig. 4.** The *gender* of a person can be changed by moving from the location of male to female in the latent space.

Even though our analysis has focused on face editing for the reasons discussed earlier it holds true also for other domains. For instance, Bau et al. [9] generated living rooms using similar approaches. They showed that some units from intermediate layers of the generator are specialized to synthesize certain visual concepts such as sofas or TVs.

So far we have discussed how autoencoders can connect the latent space and the image semantic space, as well as how the latent code can be used for image editing without influencing the image style. Next, we want to discuss how this can be used for artistic expression.

## 2.3 Sweet Spots in Latent Space

While in the former section we have seen how to use manipulation in the latent space to generate mathematical sound operations not much artistic content has been generated—just variations of photography like faces. Imprecision in AI systems can lead to unacceptable errors in the system and even result in deadly decisions; e.g. at autonomous driving or at cancer treatment. In the case of artistic applications, errors or glitches might lead to interesting, non-intended, artifacts. If those errors or glitches are treated as a bug or a feature lies in the eye of the artist. To create higher variations in the generated output some artists randomly introduce glitches within the autoencoder. Due to the complex structure of the autoencoder these glitches (assuming that they are introduced at an early layer in the network) occur on a semantic level as already discussed and might cause the models to misinterpret the input data in interesting ways. Some could even be interpreted as glimpses of autonomous creativity; see for instance the artistic work 'Mistaken Identity' by Mario Klingemann [10].

So far the latent space is explored by humans either by *random walk* or *intuitive steering* into a particular direction. It is up to human decisions if the synthesized image of a particular location in latent space is producing a visually appealing or otherwise interesting result. The question arises where to find those places and if those places can be spotted by an automatized process. The latent space is usually defined by a space of $d$ dimensions for which it is assumed the data to be represented as multivariate Gaussian distributions $\mathcal{N}(\mathbf{0}, \mathbf{I}_d)$ [11]. Therefore, the mean representation of all images lies in the center of the latent space. But what does that mean for the generated results? It is said that "beauty lies in the eyes of the beholder", however, research shows that there is a common understanding of beauty. For instance, averaged faces are perceived as more beautiful [12]. Adopting these findings to latent space let us assume that the most beautiful images (in our case faces) can be found in the center of the space. Particular deviations from the center stand for local sweet spots (e.g. female and male, ethnic groups). These types of sweet spots can be found by common means of data analysis (e.g. clustering). But where are interesting local sweet spots if it comes to artistic expression? Figure 6 demonstrates some variation in style within the latent space.

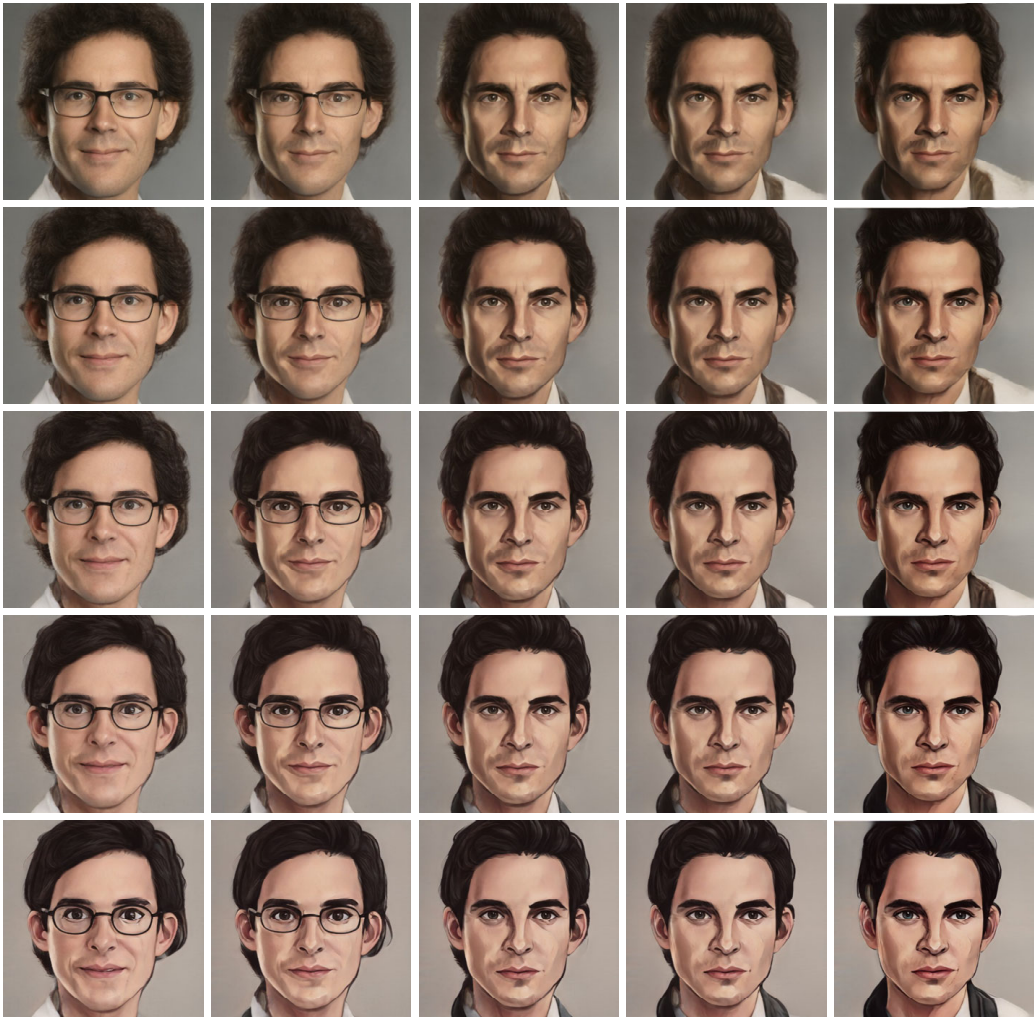

**Fig. 5.** The *facial features* of a person or the *style* of representation can be explored by changing the respective parameters in the latent space.

Of course, one can search for locations in the latent space where particular artworks from a given artist or art styles are located; see e.g. Figure 7 where the styles of different artists, as well as *white noise*[2], have been used for adoption. But isn't lingering around these sweet spots not only producing "more of the same"? How to find the local sweet spots which can define a new art style and can be deemed truly creative? Or do those discoveries of new art style lie outside of the latent space, because the latent space is trained within a particular set of defined art styles and can, therefore, produce only interpolations of those styles but nothing conceptually new?

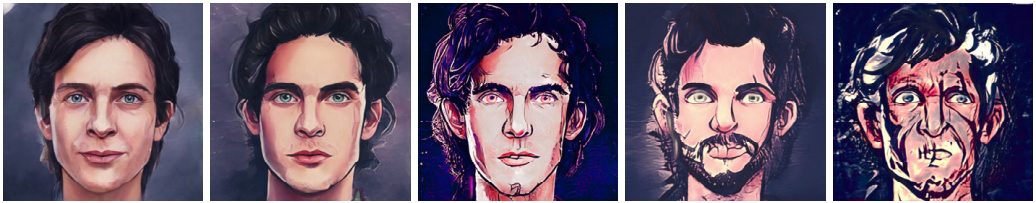

**Fig. 6.** The *style* of the image can be changed (from left to right with increasing variation) by varying parameters in the latent space which represent style instead of facial features.

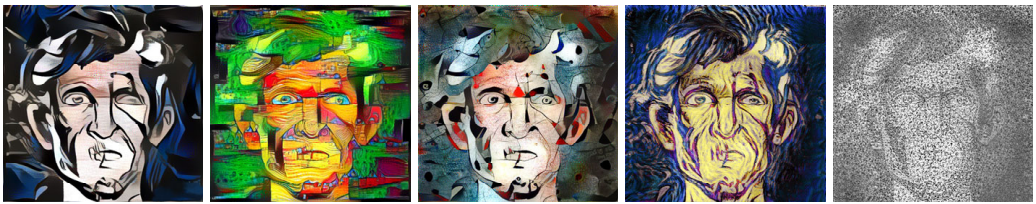

**Fig. 7.** The style of different source images is transferred to the target image. The styles of the images (from left to right) are Roy Lichtenstein, Friedensreich Hundertwasser, Joan Miro, Vincent van Gogh, and white noise.

---

[2] White noise is a signal with an equal spread frequency spectrum.

# 3 Example Artworks

So far we have discussed how AI can help to generate different variations of faces and where to find visually interesting sweet spots. In this section, we want to show how AI is supporting the creation process by applying the discussed techniques to other areas of image and object processing.[3]

## 3.1 Images

Probably, different variations of *image-to-image translation* are the most popular approach at least if looking at the mass media. The most prominent example is *style transfer*—the capability to transfer the style of one image to draw the content of another (examples are shown in Figure 7). But mapping an input image to an output image is also possible for a variety of other applications such as *object transfiguration* (e.g. horse-to-zebra, apple-to-orange, *season transfer* (e.g. summer-to-winter) or *photo enhancement* [13]. While some of the just mentioned systems are not yet in a state to be widely applicable, AI tools are taking over and gradually automating design processes which used to be time-consuming manual processes. Indeed, the most potential for AI in art and design is seen in its application to tedious, uncreative tasks such as coloring black-and-white images [14].

Marco Kempf and Simon Zimmerman used AI in their work dubbed 'DeepWorld' to generate a compilation of 'artificial countries' using data of all existing countries (around 195) to generate new anthems, flags and other descriptors [15]. Roman Lipski uses an *AI muse* (developed by Florian Dohmann et al.) to foster his/her inspiration [16]. Because the AI muse is trained only on the artist's previous drawings and fed with the current work in progress it suggests image variations in line with Roman's taste.

## 3.2 Objects

Cluzel et al. have proposed an interactive genetic algorithm to progressively sketch the desired side-view of a car profile [17]. For this, the user has taken on the role of a fitness function[4] through interaction with the system. The *chAIr Project* [18] is a series of four chairs co-designed by AI and human designers. The project explores a collaborative creative process between humans and computers. It used a GAN to propose new chairs which then have been 'interpreted' by trained designers to resemble a chair. *Deep-Wear* [19] is a method using deep convolutional GANs for clothes design. The GAN is trained on features of brand clothes and can generate images that are similar to actual clothes. A human interprets the generated images and tries to manually draw the corresponding pattern which is needed to make the finished product. Li et al. [20] introduced an artificial neural network for encoding and synthesizing the structure of 3D shapes which—according to their findings—are effectively characterized by their hierarchical organization. German et al. [21] have applied different AI techniques trained by a small sample set of shapes of bottles, to propose novel bottle-like shapes. The evaluation of their proposed methods revealed that it can be used by trained designers as well as non-designers to support the design process in different phases and that it could lead to novel designs not intended/foreseen by the designers.

---

[3] Of course these techniques have been also successfully applied to other areas such as audio and video, but should not be presented here.

[4] also referred to as objective function

# 4 Ethical and Social Implications

For decades, AI has fostered (often false) future visions ranging from transhumanist utopia to "world run by machines" dystopia. Artists and designers explore solutions concerning the semiotic, the aesthetic and the dynamic realm, as well as confronting corporate, industrial, cultural and political aspects. The relationship between the artist and the artwork is directly connected through their intentions, although currently mediated by third-parties and media tools. Understanding the ethical and social implications of AI-assisted creation is becoming a pressing need. The implications, where each has to be investigated in more detail in the future, include:

- *Bias*: AI systems are sensitive to bias. As a consequence, the AI is not being a neutral tool, but has pre-decoded preferences. Bias relevant in creative AI systems are:
  - *Algorithmic Bias* occurs when a computer system reflects the implicit values of the humans who created it; e.g. the system is optimized on dataset A and later retrained on dataset B without reconfiguring the neural network (this is not uncommon, as many people do not fully understand what is going on in the network, but are able to use the given code to run training on other data).
  - *Data Bias* occurs when your samples are not representative of your population of interest.
  - *Prejudice Bias* results from cultural influences or stereotypes which are reflected in the data.
- *Art Crisis*: Until 200 years ago painting served as the primary method for visual communication and was a widely and highly respected art form. With the invention of photography, painting began to suffer an identity crisis because painting—in its current form then—was not able to reproduce the world as accurate and with as low effort as photography. As a consequence visual artists had to change to different forms of representations not possible by photography inventing different art styles such as impressionism, expressionism, cubism, pointillism, constructivism, surrealism, up to abstract expressionism. At the time AI can perfectly simulate those styles what will happen with the artists? Will artists still be needed, be replaced by AI, or will they have to turn to other artistic work which yet cannot be simulated by AI?
- *Inflation*: Similar to the image flood which has reached us the same can happen with AI art. Because of the glut, nobody is valuing and watching the images anymore.
- *Wrong Expectations*: Only esthetic appealing or otherwise interesting or surprising results are published which can be contributed to similar effects as the well-known *publication bias* [22] in other areas. Eventually, this is leading to wrong expectations of what is already possible with AI. In addition, this misunderstanding is fueled by content claimed to be created by AI but has indeed been produced—or at least reworked—either by human labor or by methods not containing AI.
- *Unequal Judgment*: Even though the raised emotions in viewing artworks emerge from its underlying structure in the works, people also include the creation process in their judgment (in the cases where they know about it). Frequently, becoming to know that a computer or an AI has created the artwork, in the opinion of the people it turns boring, has no guts, no emotion, no soul while before it was inspiring, creative and beautiful.
- *Authorship*: The authorship of AI-generated content has not been clarified. For instance, is the authorship of a novel song composed by an AI trained exclusively on songs by Johann Sebastian Bach belonging to the AI, the developer/artist, or Bach? See e.g. [23] for a more detailed discussion.

- *Trustworthiness*: New AI-driven tools make it easy for non-experts to manipulate audio and/or visual media. Thus, image, audio as well as video evidence is not trustworthy anymore. Manipulated image, audio, and video are leading to fake information, truth skepticism, and claims that real audio/video footage is fake (known as the *liar's dividend*) [24].

## 5    Conclusion

The potential of AI in creativity has just been started to be explored. We have investigated on the creative power of AI which is represented—not exclusively—in the semantic meaningful representation of data in a dimensionally reduced space, dubbed latent space, from which images, but also audio, video, and 3D models can be synthesized. AI is able to imagine visualizations that lie between everything the AI has learned from us and far beyond and might even develop its own art styles (see e.g. deep dream [25]). However, AI still lacks intention and is just processing data.

Those novel AI tools are shifting the creativity process from crafting to generating and selecting—a process which yet can not be transferred to machine judgment only. However, AI can already be employed to find possible sweet spots or make suggestions based on the learned taste of the artist [21]. AI is without any doubt changing the way we experience art and the way we do art. Doing art is shifting from handcrafting to exploring and discovering. This leaves humans more in the role of a curator instead of an artist, but it can also foster creativity (as discussed before in the case of Roman Lipski) or reduce the time between intention and realization. It has the potential, just as many other technical developments, to democratize creativity because the handcrafting skills are not so much in need to express his/her own ideas anymore. Widespread misuse (e.g. image manipulation to produce fake pornography) can limit the social acceptance and require AI literacy. As human beings, we have to ask ourselves if feelings are wrong just because the AI never felt alike in its creation process as we do? Or should we not worry too much and simply enjoy the new artworks created no matter if they are done by humans, by AI or as a co-creation between the two ones?

# Supporting Quality Assessment in Manufacturing
# by Machine Learning:
# First Results of PREFERML Project

Alexander Gerling[1], Alaa Saleh[1], Ulf Schreier[1], Holger Ziekow[1]

[1]Furtwangen University
{ alexander.gerling, alaa.saleh, ulf.schreier, holger.ziekow }
@hs-furtwangen.de

**Keywords:** Reference Model; Machine Learning; Assembly Line; Manufacturing.

Machine learning (ML) algorithms have shown tremendous potentials in numerous fields. Our on-going research project PREFERML (Proactive Error Prevention in Manufacturing Based on Machine Learning) [1] aims to design and implement a machine learning system for the sake of generating prediction models with respect to quality checks and reducing faulty products in manufacturing processes. It is based on an industrial case study in cooperation with SICK AG. We will present first results of the project concerning a new process model for cooperating data scientists and quality engineers, a product testing model as knowledge base for machine learning computing and visual support of quality engineers in order to explain prediction results.

A typical production line consists of various test stations that conduct several measurements. Those measurements are processed by the system on the fly, to point out problematic products. Among the many challenges, one focus of the project is on support for quality engineers. Preparation of prediction models is usually done by data scientists. But the demand for data scientists is increasing too fast, when a big number of products, production lines and changing circumstances have to be considered. Hence, a software is needed which quality engineers can operate directly and leverage the results from prediction models.

Based on quality management and data science standard processes [2] [3] we created a reference process model for production error detection and correction which includes needed actors and associated tasks. With ML system and data scientist assistance we bolster the quality engineer in his work.

To support the ML system, we developed a product testing model which includes crucial information about a specific product. In this model we describe the relation to product specific features, test systems, production lines sequences etc. The idea behind this, is to provide metadata information which in turn is used by the ML system instead of individual script solutions for each product.

A ML model with good predictions has often a lack of information about the internal decisions. Therefore, it is beneficial to support the quality engineer with useful feature visualizations. By default, we support the quality engineer with 2D - 3D feature plots and histograms, in which the error distribution is visualized. On top, we developed further feature importance measures based on SHAP values [4]. These can be used to get deeper insight for particular ML decisions to significant features which get lower ranked by standard feature importance measures.

# Validation of Continuously Learning AI/ML Systems in Medical Devices – A Scenario-based Analysis

Haimerl Martin[1]

[1] Hochschule Furtwangen University

**Martin.Haimerl@hs-furtwangen.de**

**Abstract.** This paper discusses the use of continuously learning AI/ML based medical devices, i.e. devices which optimize their performance during the product's lifetime. For such devices, a regulatory strategy was recently proposed by the US Food & Drug Administration (FDA). The paper analyzes the options this approach provides as well as potential shortcomings it may pose. In particular, it studies the proposed concept of automated validation for these devices. In this analysis, the assessment of the relationship between technical parameters and clinical effects is a main focus. This includes the association to potential risks as well the dependencies between the algorithmic outcomes and the clinical environment. Additionally, potential issues w.r.t. bias, explainability, and fairness of the algorithms are addressed. The paper uses application scenarios, where ML based devices are utilized in the intensive care unit (ICU). In summary, ML based medical devices and especially continuously learning devices still possess considerable challenges which should be addressed thoroughly. Regarding appropriate regulatory strategies, a deliberate approach is recommended which prioritizes the collection of sufficient experience with ML based devices over amplifying their use in a rather uncontrolled fashion.

**Keywords:** Machine Learning, Continuously learning systems, SaMD – Software as a Medical Device, Regulatory requirements, Automated validation

## 1    Introduction

Medicine is a highly empirical discipline, where important aspects have to be demonstrated using adequate data and sound evaluations. This is one of the core requirements, which were emphasized during the development of the Medical Device Regulation (MDR) of the European Union (EU) [1]. This applies to all medical devices, including mechanical and electrical devices as well as software systems. Also, the US Food & Drug Administration (FDA) recently set a focus on the discussions about using data for demonstrating the safety and efficacy of medical devices [2]. Beside pure approval steps, they foster the use of data for optimization of the products, as nowadays data can be acquired more and more, using modern IT technology. In particular, they pursue the use of real world evidence, i.e. data that is collected through the lifetime of a device, for demonstrating improved outcomes. [2]

Such approaches require the use of sophisticated data analysis techniques. Beside classical statistics, artificial intelligence (AI) and machine learning (ML) are considered to be powerful techniques for this purpose. Currently, they gain more and more attention. These techniques allow to detect dependencies in complex situations, where inputs and/or outputs of a problem have high-dimensional parameter spaces. This can e.g. be the case when extensive data is collected from diverse clinical studies or also treatment protocols from local sites. Furthermore, AI/ML based techniques may be used in the devices themselves. For example, devices may be developed which are considered to improve complex diagnostic tasks or find individualized treatment options for specific medical conditions (see e.g. [3, 4] for an overview). For some applications, it already has been demonstrated that ML algorithms are able to outperform human experts with respect to specific success rates (e.g. [5, 6]). In this paper, it will be discussed how ML based techniques can be brought onto the market including an analysis of appropriate regulatory requirements. For this purpose, the main focus lies on ML based devices applied in the intensive care unit (ICU) as e.g. proposed in [7, 8].

The need for specific regulatory requirements comes from the observation, that AI/ML based techniques pose specific risks which need to be considered and handled appropriately. For example, AI/ML based methods are more challenging w.r.t. bias effects, reduced transparency, vulnerability to cybersecurity attacks, or general ethical issues (see e.g. [9, 10]). In particular cases, ML based techniques may lead to noticeably critical results, as it has been shown for the IBM Watson for Oncology device. In [11], it was reported that the direct use of the system in particular clinical environments resulted in critical treatment suggestions. The characteristics of ML based systems led to various discussions about their reliability in the clinical context. It requires to find appropriate ways to guarantee their safety and performance. (cf. [12]) This applies to the field of medicine / medical devices as well as AI/ML based techniques in general. The latter was e.g. approached by the EU in their *Ethics guidelines for trustworthy AI* [9].

Driven by this overall development, the FDA started a discussion regarding an extended use of ML algorithms in SaMD (software as a medical device) with a focus in quicker release cycles. In [13], it pursued the development of a specific process which makes it easier to bring ML based devices onto the market and also to update them during their lifecycle. Current regulations for medical devices, e.g. in US or EU, do not provide specific guidelines for ML based devices. In particular, this applies to systems which continuously collect data in order to improve the performance of the device. Current regulations focus on a fixed status of the device, which may only be adapted in a minor extent after the release. Usually, a new release or clearance by the authority is required, when the clinical performance of a device is modified. But continuously learning systems exactly want to do such improvement steps using additional real-world data from daily applications without extra approvals (see fig. 1).

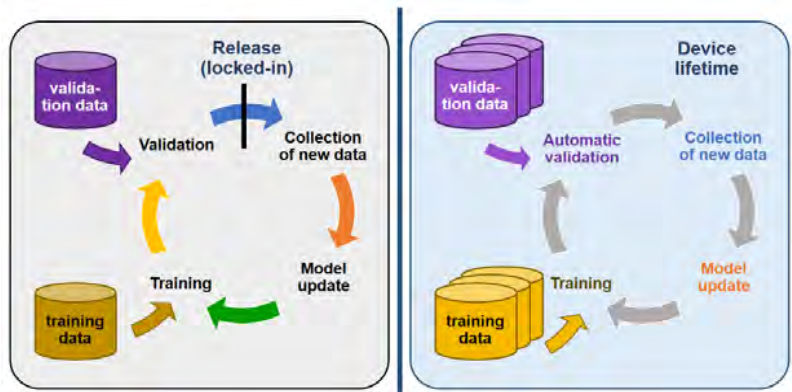

**Fig. 1.** Basic approaches for AI/ML based medical devices. Left side: classical approach, where the status of the software has to be fixed after the release / approval stage. Right side: continuously learning system where data is collected during the lifetime of the device without a separated release / approval step. In this case, an automatic validation step has to guarantee proper safety and efficacy.

In [13], the FDA made suggestions how this could be addressed. It proposed the definition of so called SaMD Pre-Specifications (SPS) and an Algorithm Change Protocol (ACP), which are considered to represent major tools for dealing with modifications of the ML based system during its lifetime. Within the SPS, the manufacturer has to define the anticipated changes which are considered to be allowed during the automatic update process. In addition, the ACP defines the particular steps which have to be implemented to realize the SPS specifications. See [13] for more information about SPS and ACP. But the details are not yet well elaborated by the FDA at the moment. The FDA requested for suggestions with respect to this.

In particular, these tools serve as a basis for performing an automated validation of the updates. The applicability of this approach depends on the risk of the SaMD. In [13], the FDA uses the risk categories from the International Medical Device Regulators Forum (IMDRF) [14]. This includes the categories *State of healthcare situation or condition* (*critical* vs. *serious* vs. *non-critical*) and *Significance of information provided by SaMD to healthcare decision* (*treat or diagnose* vs. *drive clinical management* vs. *inform clinical management*) as the basic attributes. According to [13], the regulatory requirements for the management of ML based systems are considered to depend on this classification as well as the particular changes which may take place during the lifetime of the device. The FDA categorizes them as changes in *performance*, *inputs*, and *intended use*. Such anticipated changes have to be defined in the SPS in advance.

The main purpose of the present paper is to discuss the validity of the described FDA approach for enabling continuously learning systems. Therefore, it uses a scenario based technique to analyze whether validation in terms of SPS and ACP can be considered adequate tools. The scenarios represent applications of ML based devices in the ICU. It checks its consistency with other important regulatory requirements and analyzes pitfalls which may jeopardize the safety of the devices. Additionally, it discusses whether more general requirements can be sufficiently addressed in the scenarios, as e.g. proposed in ethical guidelines for AI based systems like [9, 10]. This is not considered as a comprehensive analysis of the topics, but as an addition to current discussions about risks and ethical issues, as they are e.g. discussed in [10, 12].

Finally, the paper proposes own suggestions to address the regulation of continuously learning ML based systems. Again, this is not considered to be a full regulatory strategy, but a proposal of particular requirements, which may overcome some of the current limitations of the approach discussed in [13]. The overall aim of this paper is to contribute to a better understanding of the options and challenges of AI/ML based devices on the one hand and to enable the development of best practices and appropriate regulatory strategies, in the future.

## 2   Methods

Within this paper, the analysis of the FDA approach proposed in [13] is performed using specific reference scenarios from ICU applications, which are particularly taken from [13] itself. The focus lies on ML based devices which allow continuous updates of the model according to data collected during the lifetime of the device. In this context, SPS and ACP are considered as crucial steps which allow an automated validation of the device based on specified measures. In particular, the requirements and limitations of such an automated validation are analyzed and discussed, including the following topics / questions.

- Is automated validation reasonable for these cases? What are limitations / potential pitfalls of such an approach when applied in the particular clinical context?

- Which additional risks could apply to AI/ML based SaMD, in general, which go beyond the existing discussions in the literature as e.g. presented in [9, 10, 12]?

- How should such issues be taken into account in the future?
  What could be appropriate measures / best practices to achieve reliability?

The following exemplary scenarios are used for this purpose.

- Base Scenario ICU: ML based Intensive Care Unit (ICU) monitoring system where the detection of critical situations (e.g. regarding physiological instability, potential myocardial infarcts or sepsis) is addressed by using ML. Using auditory alarms, the ICU staff is informed to initiate appropriate measures to treat the patients in these situations. This scenario addresses a 'critical healthcare situation or condition' and is considered to 'drive clinical management' (according to the risk classification used in [13]).

- Modification "locked": ICU scenario as presented above, where the release of the monitoring system is done according to a locked state of the algorithm.

- Modification "cont-learn": ICU scenario as presented above, where the detection of alarm situations is continuously improved according to data acquired during daily routine, including adaptation of performance to sub-populations and/or characteristics of the local environment. In this case, SCS and ACP have to define standard measures like success rates of alarms/detection and requirements for the management of data, update of the algorithm, and labeling. More details of such requirements are discussed later. This scenario was presented as scenario 1A in [13] with minor modifications.

## 3    Analysis

This section provides the basic analysis of the scenarios according to the particular aspects addressed in this paper. It addresses the topics automated validation, man-machine interaction, explainability, bias effects, and confounding, fairness and non-discrimination as well as corrective actions to systematic deficiencies.

**Basic Considerations about Automated Validation**

According to standard regulatory requirements [1, 15, 16], validation is a core step in the development and for the release of medical devices. According to [17], a change in performance of a device (including an algorithm in a SaMD) as well as a change in particular risks (e.g. new risks, but also new risk assessment or new measures) usually triggers a new premarket notification (510(k)) for most of the devices which get onto the market in the US. Thus, such situations require an FDA review for clearance of the device. For SaMD, this requires to include an analytical evaluation, i.e. correct processing of input data to generate accurate, reliable, and precise output data. Additionally, a clinical validation as well as the demonstration of a valid clinical association need to be provided. [18] This is intended to show that the outputs of the device appropriately work in the clinical environment, i.e. have a valid association regarding the targeted clinical condition and achieve the intended purpose in the context of clinical care. [18]

Thus, based on the current standards, a device with continuously changing performance usually requires a thorough analysis regarding its validity. This is one of the main points, where [13] proposes to establish a new approach for the "cont-learn" cases. As already mentioned, SPS and ACP basically have to be considered as tools for automated validation in this context. Within this new approach, the manual validation step is replaced by an automated process with only reduced or even no additional control by a human observer. Thus, it may work as an automated of fully automatic, closed loop validation approach. The question is whether this change can be considered as an appropriate alternative. In the following, this question is addressed using the ICU scenario with a main focus on the "cont-learn" case. Some of the aspects also apply to the "locked" cases. But the impact is considered to be higher in the "cont-learn" situation, since the validation step has to be performed in an automated fashion. Human oversight, which is usually considered important, is not included here during the particular updates.

## Analysis of Validation Steps

Within the ICU scenario, the validation step has to ensure that the alarm rates stay on a sufficiently high level, regarding standard factors like specificity, sensitivity, area under curve (AUC), etc. Basically, these are technical parameters which can be analyzed according to an analytical evaluation as discussed above. (see also [18]) This could also be applied to situations, where continuous updates are made during the lifecycle of the device, i.e. in the "cont-learn". However, there are some limitations of the approach. On the one hand, it has to be ensured, that this analysis is sound and reliable, i.e. it is not compromised according to statistical effects like bias or other deficiencies in the data. On the other hand, it has to be ensured that the success rates really have a valid clinical association and can be used as a sole criterion for measuring the clinical impact. Thus, the relationship between pure success rates and clinical effects has to be evaluated thoroughly and there may be some major limitations.

One major question in the ICU scenario is, whether better success rates really guarantee a higher or at least sufficient level of clinical benefit. This is not innately given. For example, a higher success rate of the alarms may still have a negative effect when the ICU staff relies more and more on the alarms and subsequently reduces attention. Thus, it may be the case that the initiation of appropriate treatment steps may be compromised even though the actually occurring alarms seem to be more reliable.

In particular, this may apply in situations where the algorithms are adapted to local settings, like in the "cont-learn" scenario. Here, the ML based system is intended to be optimized to sub-populations in the local environment or to specific treatment preferences at the local site. According to habituation effects, the staff's expectations get aligned to the algorithm's behavior to a certain degree after a period of time. But when the algorithm changes or an employee from another hospital or department takes over duties in the local unit, the reliability of the alarms may be affected. In these cases, it is not clear whether the expectations are well aligned with the current status of the algorithm – either in the positive or negative direction. Since the data updates of the device are intended to improve its performance w.r.t. detection rates, it is clear that significant effects on user interaction may happen. Under some circumstances, the overall outcome in terms of the clinical effect may be impaired.

Evaluation of such risks have to be addressed during validation. It is questionable whether this can be performed by using an automatic validation approach which focuses on alarm rates but does not include an assessment of the associated risks. At least a clear relationship between these two aspects has to be demonstrated in advance. It is also unclear, whether this could be achieved by assessment of pure technical parameters which are defined in advance as required by the SPS and ACP. Usually, ML based systems are trained to a specific scenario. They provide a specific solution for this particular problem. But they do not have a more general intelligence and reasoning about potential risks, which were not under consideration at that point of time. Such a more general intelligence can only be provided when using human oversight.

## Consideration of Risks and Man-Machine Interaction

In general, it is not clear whether technical aspects like alarms lead to valid reactions by the users. In technical terms, alarm rates are basically related to the probability of occurrence of specific hazardous situations. But they do not address a full assessment of occurrence of harm. However, this is pivotal for risk assessment in medical devices, in particular for risks related to potential use errors. This is considered to be one of the main reasons why a change in risk parameters triggers a new premarket approval in the US according to [17]. Also, the MDR [1] sets high requirements to address the final clinical impact and not only technical parameters.

Basically, the example emphasizes the importance to consider the interaction between man and machine, or in this case, the algorithm and its clinical environment. This is addressed in the usability standards for medical devices, e.g. ISO 62366 [19]. For this reason, the ISO 62366 requires that the final (summative) usability evaluation is performed using the final version of the device (in this case, the algorithm) or an equivalent version. This is in conflict with the FDA proposal which allows to perform this assessment based on previous versions. At most, a predetermined relationship between technical parameters (alarm rates) and clinical effects (in particular, use related risks) can be obtained. For usage of ML based devices, it remains crucial to consider the interaction between the device and the clinical environment as there usually are important interrelationships.

**Comprehensiveness of Included Inputs and Outcomes**

The outcome of an ML based algorithm always depends on the data it gets provided. Whenever an input parameter is omitted, which is clinically relevant, the resulting outcome of the ML based system is limited. In the presented scenarios, the pure alarm rates may not be the only clinically relevant outcomes. Even though, such parameters are usually the main focus regarding the quality of algorithms, e.g. in publications about ML based techniques. This is due to the fact, that such quality measures are commonly considered the best available objective parameters, which allow a comparison of different techniques.

This even more applies to other ML based techniques which are also very popular in the scientific community, like segmentation tasks in medical image analysis. Here the standard quality measures are general distance metrics, i.e. differences between segmented areas. [20] They usually do not include specific clinical aspects like the accuracy in specific risk areas, e.g. important blood vessels or nerves. But such aspects are key factors to ensure the safety of a clinical procedure in many applications. Again, only technical parameters are typically in focus. The association to the clinical effects is not assessed accordingly. This situation is depicted in fig. 2 for the ICU as well as image segmentation cases.

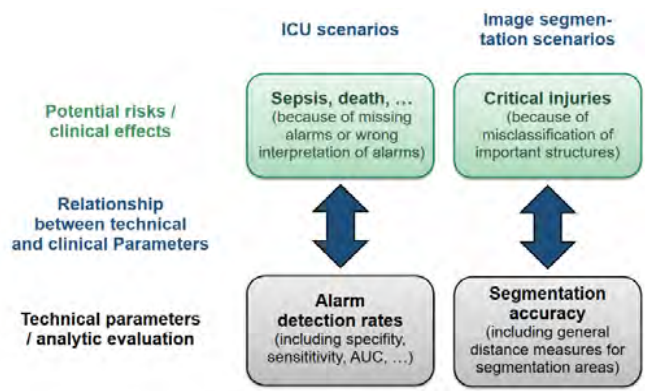

**Fig. 2.** Relationship between technical parameters / analytical evaluation (as usually considered in many publications about ML based techniques) on the one hand and potential risks / clinical effects (finally relevant for release of medical devices) on the other hand (using ICU and image segmentation scenarios).

Additionally, the validity of an outcome in medical treatments depends on many factors. Regarding input data, multiple parameters from a patient's individual history may be important for deciding about a particular diagnosis or treatment. A surgeon usually has access to a multitude of data and also side conditions (like socio-economic aspects) which should be included in an individual diagnosis or treatment decision. His general intelligence and background knowledge allow him to include a variety of individual aspects, which have to be considered for a specific case-based decision. In contrary, ML based algorithms rely on a more standardized structure of input data and are only trained for a specific purpose. They lack a more general intelligence, which allows them to react in very specific situations. Even more, ML based algorithms need to generalize and thus to mask out very specific conditions, which could by fatal in some cases.

In [13], the FDA presents some examples where changes of the inputs in an ML based SaMD are included. It is surprising, that the FDA considers some of them as candidates for a continuous learning system, which does not need an additional review, when a tailored SPS/ACP is available.

**Lack of Explainability and its Impact on Clinical Practice**

Such discrepancies between technical outcomes and clinical effects also apply to situations like the ICU scenario, which only informs or drives clinical management. Often users rely on automatically provided decisions, even when they are informed that this only is a proposal. Again, this is a matter of man-machine interaction. This gets even worse due to the lack of explainability which ML based algorithms typically have. [9, 21] When surgeons or more general users (e.g, ICU staff) detect situations which require a diverging treatment because of very specific individual conditions, they should overrule the algorithm. But users will often be confused by the outcome of the algorithm and do not have a clear idea how they should treat conflicting results between the algorithm's suggestions and their own belief. As long as the ML based decision is not transparent to the user, they will not be able to merge these two directions. The IBM Watson example, referenced in the introduction shows, that this actually is an issue [11].

This may be even more serious, when the users (i.e. healthcare professionals) fear litigation because they did not trust the algorithm. In a situation, where the algorithm's outcome finally turns out to be true, they may be sued because of this documented deviation. Because of such issues, the EU General Data Protection Regulation (GFPR) [22] requires that the users get autonomy regarding their decisions and transparency about the mechanisms underlying the algorithm's outcome. [23] This may be less relevant for the patients, who usually have only limited medical knowledge. They will probably also not understand the medical decisions in conventional cases. But it is highly relevant for responsible healthcare professionals. They require to get basic insights how the decision emerged, as they finally are in charge of the treatment. This demonstrates that methods regarding the explainability of ML based techniques are important. Fortunately, this currently gets a very active field. [21, 24] This need for explainability applies to locked algorithms as well as situations where continuous learning is applied.

**Bias Effects due to Imbalance of Data**

Based on their own data-driven nature, ML based techniques highly depend on a very high quality of data which are provided for learning and validation. In particular, this is important for the analytical evaluation of the ML algorithms. One of the major aspects are bias effects due to unbalanced input data. For example, in [25] a substantially different detection rate between white and colored people was recognized due to unbalanced data. Beside ethical considerations, this demonstrates dependencies of the outcome quality on sub-populations, which may be critical in some cases. Even though, the FDA proposal [13] currently does not consequently include specific requirements for assessing bias factors or imbalance of data.

However, high quality requirements for data management are crucial for ML based devices. In particular, this applies to the ICU "cont-learn" cases. There have to be very specific protocols that guarantee that new data and updates of the algorithms are highly reliable w.r.t. bias effects. Most of the currently used ML based algorithms fall under the category of supervised learning. Thus, they require accurate and clinically sound labeling of the data. During the data collection, it has to be ensured how this labeling is performed and how the data can be fed back into the system in a "cont-learn" scenario. Additionally, the data needs to stay balanced – whatever this means in a situation where adaptions to sub-populations and/or local environments are intended for optimization. It is unclear, whether and how this could be achieved by staff who is only operating with the system but possibly does not know potential algorithmic pitfalls.

In the ICU scenario, many data points probably need to be recorded by the system itself. Thus, a precise and reliable recording scheme has to be established which automatically avoids imbalance of data on the one hand and fusion with manual labelings on the other hand. Basically, the SPS and ACP (proposed in [13]) are tools to achieve this. The question is whether this is possible in a reliable fashion using automated processes. A complete closed loop validation approach seems to be questionable, especially when the assessment of clinical impact has to be included. Thus, the integration of humans in adequate healthcare professionals as well as ML/AI experts with sufficient statistical knowledge seems reasonable. At least, bias assessment steps should be included. As already mentioned, this is not addressed in [13] in a dedicated way.

**Confounding Factors and Dependence on Clinical Environment**

Further on, the outcomes may be compromised by side effects in the data. It may be the case, that the main reason for a specific outcome of the algorithm is not a relevant clinical parameter but a specific data artifact, i.e. some confounding factor. In the ICU case, it could be the case, that the ICU staff reacts early to a potentially critical situation and e.g. gives specific medication in advance to prevent upcoming problems. The physiological reaction of the patient can then be visible in the data as some kind of artifact. During its learning phase, the algorithm may recognize the critical situation not based on a deeper clinical reason, but on detecting the physiological reaction pattern. This may cause serious problems as shown subsequently. In the presented scenario, the definition of clinical situation and the pattern can be deeply coupled by design, since the labeling of the data by the ICU staff and the administration of the medication will probably be done in combination at the particular site. This may increase the probability of such effects.

Usually, confounding factors are hard to determine. Even when they can be detected, they are hard to be communicated and managed in an appropriate way. How should healthcare professionals react, when they get such potentially misleading information (see discussion about liability). This further limits the explanatory power of ML based systems. When confounders are not detected, they may have unpredictable outcomes w.r.t. the clinical effects. For example, consider the following case. In the ICU scenario, an ML based algorithm gets trained in a way that it basically detects the medication artifact as described above during the learning phase. In the next step, this algorithm is used in clinical practice and the ICU staff relies on the outcome of the algorithm. Then, on the one hand, the medication artifact is not visible unless the ICU staff administers the medication. On the other hand, the algorithm does not recognize the pattern and thus does not provide an alarm. Subsequently, the ICU staff does no act appropriately to manage the critical situation.

In particular, such confounders may be more likely in situations where a strong dependence between the outcome of the algorithm and the clinical treatment exists. Further examples of such effects were discussed in [7] for ICU scenarios. The occurrence of confounders may be a bit less probable in pure diagnostic cases without influence of the diagnostic task onto the generation of data. But even here, such confounding factors may occur. The discussion in [10] provided examples where confounders may occur in diagnostic cases e.g. because of rulers placed for measurements on radiographs. In most of the publications about ML based techniques, such side effects are not discussed (or only in a limited fashion). In many papers, the main focus is the technical evaluation and not the clinical environment and the interrelation between technical parameters and clinical effects.

## Fairness and Non-Discrimination

Additional important aspects which are amply discussed in the context of AI/ML based systems are discrimination and fairness (see e.g. [10]). In particular, the EU puts a high priority of their future AI/ML strategy on fairness requirements [9]. Fairness is often closely related to bias effects. But it goes beyond to more general ethical questions, e.g. regarding the natural tendency of ML based systems to favor specific subgroups. For example, the ICU scenario "cont-learn" is intended to optimize w.r.t. to specifics of sub-populations and local characteristics, i.e. it tries to make the outcome better for specific groups. Based on such optimization, other groups (e.g. minorities, underrepresented groups) which are not well represented may be discriminated in some sense. This is not a statistical but a systematic effect.

Superiority of a medical device for a specific subgroup (e.g. gender, social environment, etc.) is not uncommon. For example, some diagnosis steps, implants, or treatments achieve deviating success rates when applied to women in comparison to men. This also applies to differences between adults and children. When assessing bias in clinical outcome in ML based devices, it will probably often be unclear whether this is due to imbalance of data or a true clinical difference between the groups. Does an ML based algorithm has to adjust the treatment of a subgroup to a higher level, e.g. a better medication, to achieve comparable results, when the analysis recognized worse results for this subgroup? Another example could be a situation where the particular group does not have the financial capabilities to afford the high-level treatment. This could e.g. be the case in a developing country or in subgroups with a lower insurance level. In these cases, the inclusion of socio-economical parameters into the analysis seems to be unavoidable. Subsequently, this compromises the notion of fairness as basic principle in some way.

This is nothing genuine to ML based devices. But in the case of ML based systems with a high degree of automation, the responsibility for the individual treatment decision more and more shifts from the health care professional to the device. It is implicitly defined in the ML algorithm. In comparison to human reasoning, which allows some weaknesses in terms of individual adjustments of general rules, ML based algorithms are rather deterministic / unique in their outcome. For a fixed input, they have one dedicated outcome (when we neglect statistical algorithms which may allow minor deviations). Differences of opinions and room for individual decisions are main aspects of ethics. Thus, it remains unclear how fairness can be defined and implemented at all when considering ML based systems. This is even more challenging as socio-economical aspects (even more than clinical aspects) are usually not included in the data and analysis of ML based techniques in medicine. Additionally, they are hard to assess and implement in a fair way, especially when using automated validation processes.

## Corrective Actions regarding Systematic Deficiencies

Another disadvantage of ML based devices is the limited opportunities to fix systematic deficiencies in the outcome of the algorithm. Let us assume that during the lifetime of the ICU monitoring system a systematic deviation of the intended outcome was detected, e.g. in the context of post-market surveillance or due to an increased number of serious adverse events. According to standard rules, a proper preventive respectively corrective action has to be taken by the manufacturer. In conventional software devices, the error simple should be eliminated, i.e. some sort of bug fixing has to be performed. For ML based devices it is less clear, how bug fixing should work especially when the systematic deficiency is deeply hidden in the data and/or ML model. In these cases, there usually is no clear reason for the deficiency. Subsequently, the deficiency cannot be resolved in a straightforward way using standard bug fixing. There is no dedicated route to find the deeper reasons and to perform changes which could cure the deficiencies, e.g. by providing additional data or changing the ML model. Even more, other side effects may easily occur, when data and model are changed manually by intent to fix the issue.

## 4    Discussion and Outlook

In summary, there are many open questions, which are not yet clarified. There still is little experience how ML based systems work in clinical practice and which concrete risks may occur. Thus, the FDA's commitment to foster the discussion about ML based SaMD is necessary and appreciated by many stakeholders as the feedback docket [26] for [13] shows. However, it is a bit surprising that the FDA proposes to substantially reduce its very high standards in [13] at this point of time. In particular, it is questionable whether an adequate validation can be achieved by using a fully automatic approach as proposed in [13]. ML based devices are usually optimized according to very specific goals. They can only account for the specific conditions that are reflected in the data and the used optimization / quality criteria. They do not include side conditions and a more general reasoning about potential risks in a complex environment. But this is important for medical devices.

For this reason, a more deliberate path would be suited, from the author's perspective. In a first step, more experience should be gained w.r.t. to the use of ML based devices in clinical practice. Thus, continuous learning should not be a first hand option. First, it should be demonstrated that a device works in clinical practice before a continuous learning approach should be possible. This could also be justified from a regulatory point-of-view. The automated validation process itself should be considered as a feature of the device. It should be considered as part of the design transfer which enables safe use of the device during its lifecycle. As part of the design transfer, it should be validated itself. Thus, it has to be demonstrated that this automated validation process, e.g. in terms of the SPS and ACP, works in a real clinical environment. Ideally, this would have been demonstrated during the application of the device in clinical practice.

Thus, one reasonable approach for a regulatory strategy could be to reduce or prohibit the options for enabling automatic validation in a first release / clearance of the device. During the lifetime, direct clinical data could be acquired to demonstrate a better insight into the reliability and limitations of the automatic validation / continuous learning approach. In particular, the relation between technical parameters and clinical effects could be assessed on a broader and more stable basis. Based on this evidence in real clinical environments, the automated validation feature could then be cleared in a second round.

Otherwise, the validity of the automated validation approach would have to be demonstrated in a comprehensive setting during the development phase. In principle, this is possible when enough data is available which truly reflects a comprehensive set of situations. As discussed in this paper, there are many aspects which do not render this approach impossible but very challenging. In particular, this applies to the clinical effects and the interdependency between the users and clinical environment on the one hand and the device, including the ML algorithm, data management, etc., on the other hand. This also includes not only variation in the status and needs of the individual patient but also the local clinical environment and potentially also the socioeconomic setting.

Following a consequent process validation approach, it would have to be demonstrated that the algorithm reacts in a valid and predictable way no matter which training data have been provided, which environment have to be addressed, and which local adjustments have been applied. This also needs to include deficient data and inputs in some way. In [20], it has been shown that the variation of outcomes can be substantial, even w.r.t. rather simple technical parameters. In [20], this was analyzed for scientific contests ("challenges") where renowned scientific groups supervised the quality of the submitted ML algorithms. This demonstrates the challenges validation steps for ML based systems still include, even w.r.t. technical evaluation.

For these reasons, it seems adequate to pursue the regulatory strategy in a more deliberate way. This includes the restriction of the "cont-learn" cases as proposed. This also includes a better classification scheme, where automated or fully automatic validation is possible. Currently, the proposal in [13] does not provide clear rules when continuous learning is allowed. It does not really address a dedicated risk-based approach that defines which options and limitations are applicable. For some options, like the change of the inputs, it should be reviewed, whether automatic validation is a natural option. Additionally, the dependency between technical parameters and clinical effects as well as risks should get more attention. In particular, the grade of interrelationship between the clinical actions and the learning task should be considered.

In general, the discussions about ML based medical devices are very important. These techniques provide valuable opportunities for improvements in fields like medical technologies, where evidence based on high quality data is crucial. This applies to the overall development of medicine as well as to the development of sophisticated ML based medical devices. This also includes the assessment of treatment options and success of particular devices during their lifetime. Data-driven strategies will be important for ensuring high-level standards in the future. They may also strengthen regulatory oversight in the long term by amplifying the necessity of post-market activities. This seems to be one of the promises the FDA envisions according to their concepts of "total product lifecycle quality (TPLC)" and "organizational excellence" [13]. Also, the MDR strengthens the requirements for data-driven strategies in the pre- as well as post-market phase. But it should not shift the priorities for a basically proven-quality-in-advance (ex-ante) to a primarily ex-post regulation, which boils down to a trial-and-error oriented approach in the extreme. Thus, we should aim at a good compromise between pushing these valuable and innovative options on the one hand and potential challenges and deficiencies on the other hand.

# Augmented Reality in the Operating Room for Neurosurgical Interventions

Christian Kunz[1], Franziska Mathis-Ullrich[1], and Björn Hein[1,2]

[1] Karlsruhe Institute of Technology (KIT), Institute for Anthropomatics and Robotics, Karlsruhe, Germany
franziska.ullrich@kit.edu
[2] Karlsruhe University of Applied Sciences, Karlsruhe, Germany
bjoern.hein@hs-karlsruhe.de

**Abstract.** Neurosurgical procedures are associated with great challenges for the surgeon since a high degree of precision is required. Operations are performed within a limited space and often concealed structures are not visible to the surgeon. A system is proposed that integrates augmented reality into a digital operating room. The basis for this is an understanding of the scene and the integration into the surgical workflow. In a first step a two-stage process is implemented to detect the patient on the operating table with high precision. Further a solution is presented to semantically segment the surgical scene to detect and track medical instruments. For better understanding of the situation in the operation room the medical staff is tracked with OpenPose. These solutions build the base for a precise and robust integration of augmented reality into the digital operating room.

**Keywords:** Computer Assisted Surgery, Augmented Reality, Neurosurgical Interventions, Digital OR, Ventricular Puncture.

## 1    Introduction

Computer-assisted technologies in medical interventions are intended to support the surgeon during treatment and improve the outcome for the patient. One possibility is to augment reality with additional information that would otherwise not be perceptible to the surgeon. In medical applications, it is particularly important that demanding spatial and temporal conditions are adhered to. Challenges in augmenting the operating room are the correct placement of holograms in the real world, and thus, the precise registration of multiple coordinate frames to each other, the exact scaling of holograms, and the performance capacity of processing and rendering systems.

In general, two different scenarios can be distinguished. First, applications exist, in which a placement of holograms with an accuracy of 1 cm and above are sufficient. These are mainly applications where a person needs a three-dimensional view of data. An example in the medical field may be the visualization of patient data, e.g. to understand and analyse the anatomy of a patient, for diagnosis or surgical planning. The correct visualization of these data can be of great benefit to the surgeon. Often only 2D patient data is available, such as CT or MRI scans. The availability of 3D representations depend strongly on the field of application. In neurosurgery 3D views are available but often not extensively utilized due to their limited informative value. Additionally computer monitors are a big limitation, because the data can not be visualized in real world scale. Further application areas are the translation of known user interfaces into augmented reality (AR) space. The benefit here is that a surgeon refrains from touching anything, but can interact with the interface in space using hand or voice gestures. Applications visualizing patient data, such as CT scans, only require a rough positioning of the image or holograms in the operation room (OR). Thus, the surgeon can conveniently place the application freely in space. The main requirement is then to keep the holograms in a constant position. Therefore, the internal tracking of the AR device is sufficient to hold the holograms at a fixed position in space. The second scenario covers all applications, in which an exact registration of holograms to the real world is required, in particular with a precision below 1 cm. These scenarios are more demanding, especially when holograms must be placed precisely over real patient anatomy. To achieve this, patient tracking is essential to determine position and to follow patient movements. The system therefore needs to track the patient and adjust the visualization to the current situation. Furthermore, it is necessary to track and augment surgical instruments and other objects in the operating room. The augmentation needs to be visualized at the correct spatial position and time constraints need to be fulfilled. Therefore, the AR system needs to be embedded into the surgical workflow and react to it. To achieve these goals modern state of the art machine learning algorithms are required. However, the computing power on available AR devices is often not yet sufficient for sophisticated machine learning algorithms. One way to overcome this shortcoming is the integration of the AR system into a distributed system with higher capabilities, such as the digital operating theatre OP:Sense (see Fig. 2).

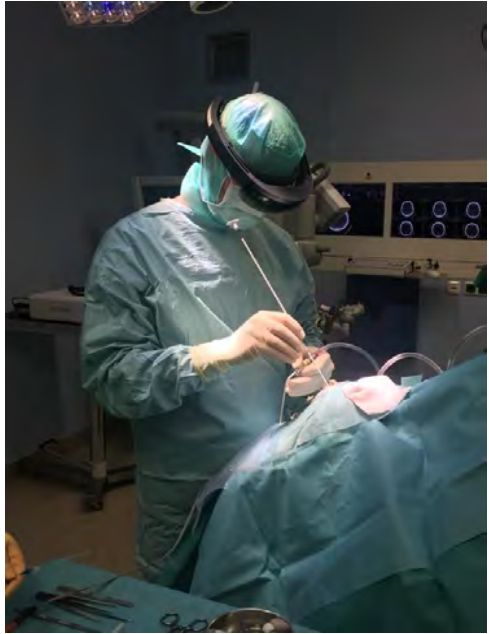

**Fig. 1.** A neurosurgeon performing a ventricular puncture wearing a HoloLens.

In this work an augmented reality system *HoloMed* [4] (see Fig. 1) is integrated into the surgical research platform for robot assisted surgery *OP:Sense* [5]. The objective is to enable high-quality and patient-safe neurosurgical procedures in order to increase the surgical outcome by providing surgeons with an assistance system that supports them in cognitively demanding operations. The physician's perception limits are extended by the AR system, which bases on supporting intelligent machine learning algorithms. AR glasses allow the neurosurgeon to perceive the internal structures of the patient's brain. The complete system is demonstrated by applying this methodology to the ventricular puncture of the human brain, one of the most frequently performed procedures in neurosurgery. The ventricle system has an elongated shape with a width of 1-2 cm and is located in a depth of 4 cm inside the human head. Patient models are generated fast (< 2s) from CT-data [3], which are superimposed over the patient during operation and serve as a navigation aid for the surgeon. In this work the expanded system architecture is presented to overcome some limitations of the original system where all information were processed on the Microsoft HoloLens, which lead to performance deficits. To overcome these shortcomings the HoloMed project was integrated into OP:Sense for additional sensing and computing power.

## 2    Material and Methods

To achieve integration of AR into the operation room and the surgical workflows, the patient, the instruments and the medical staff need to be tracked. To track the patient, a marker system is fixated on the patient head and registration from the marker system to the patient is determined. A two-stage process was implemented for this purpose. First the rough position of the patient's head is determined on the OR table by applying a YOLO v3 net to reduce the search space. Then a robot with a mounted RGB-D sensor is used to scan the acquired area and build a point cloud of the same. To determine the patient's head in space as precisely as possible a two-step surface matching approach is utilized. During recording, the markers are also tracked. With known position of the patient and the markers, the registration matrix can be calculated. For the ventricular puncture a solution is proposed to track the puncture catheter to determine the depth of insertion into the human brain. By tracking the medical staff the system is able to react to the current situation, e.g. if an instrument is passed. In the following the solutions are described in detail.

### 2.1   Experimental Setup

Our digital operating room OP:Sense (illustrated in Fig. 2a) consists of an OR table with two robots attached to it, a Kuka LWR4 and a Franka Panda lightweight robot. Several sensors are integrated into the setup on the ceiling rack: an ARTRRACK 2 system to track retroreflective markers consisting of six IR cameras and four Microsoft Kinect sensors (shown in Fig. 2b). Any objects can be tagged with markers to track them, provided that marker-to-instrument registration is available. Robot and operating table can be tracked in the room via ART markers. The Microsoft Kinect sensors provide a RGB-D stream of the operation area. Intel RealSense D415 and D435 cameras can additionally be placed inside the OR or can be mounted on the robots, to capture a defined near-field area. A patient phantom head from Synbone and a custom-build phantom skull were used during experiments. A Microsoft HoloLens is used to visualize AR to the surgeon. It employs the Unity 3D graphics engine to visualize the holographic scene. Patient tracking is provided through two different marker systems: 1) the Aruco library in combination with OpenCV and 2) the Vuforia library. OP:Sense is based on the Robot Operating System (ROS), which is a middleware for robotic platforms, consisting of a set of software libraries and tools. Core components are so-called nodes connecting all system components with each other.

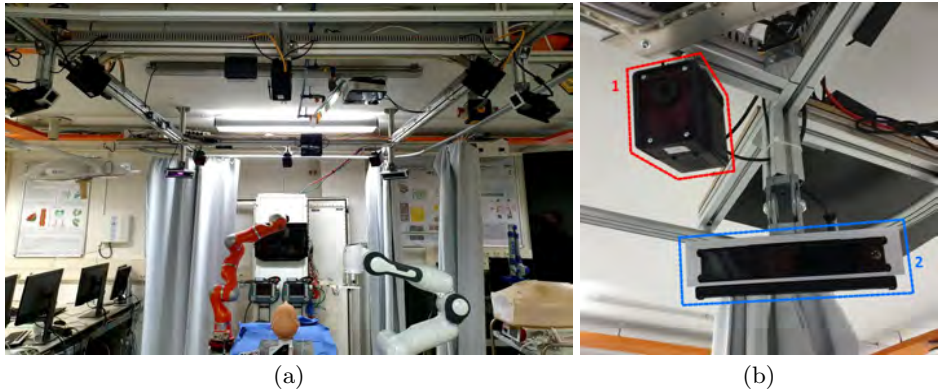

(a)                                              (b)

**Fig. 2.** a) OP:Sense system setup. b) Sensor setup with six ARTTRACK 2 IR cameras (red 1) and four Microsoft Kinect sensors (blue 2).

## 2.2 Patient Detection

To detect the patient's head, the coarse position is first determined with the YOLO v3 CNN [6], performed on the Kinect RGB image streams. The position in 3D is determined through the depth stream of the sensors. The OR table and the robots are tracked with retroreflective markers by the ARTTRACK system. This step reduces the spatial search area for fine adjustment. The Franka Panda has an attached Intel RealSense RGB-D camera as depicted in Fig. 3.

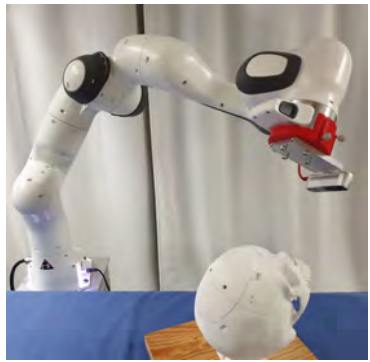

**Fig. 3.** Franka Panda with attached RGB-D sensor.

The precise determination of the position is performed on the depth data with surface matching. The robot scans the area of the coarsely determined position of the patient's head. A combined surface matching approach with feature-based and ICP matching was implemented. The process to perform the surface matching is depicted in Fig. 4. In clinical reality, a CT scan of the patient head is always performed prior to a ventricular puncture for diagnosis, such that we can safely assume the availability of CT data. A process to segment the patient models from CT data was proposed by Kunz et al. in [3]. The algorithm processes the CT data extremely fast in under two seconds. The data format is '.nrrd', a volume model format, which can easily be converted into surface models or point clouds. The point cloud of the patient's head CT scan is the reference model that needs to be found in OR space. The second point cloud is recorded from the RealSense depth stream mounted on the Panda robot by scanning the previously determined rough position of the patient head. All points are recorded in world coordinate space. The search space is further restricted with a segmentation step by filtering out points that are located on the OR table. Additionally, manual changes can be made by the surgeon. In a performance optimization, the resolution of the point clouds is reduced to decrease processing time without loosing too much accuracy. The normals of both point clouds generated from CT data and from the recorded RealSense depth stream are subsequently calculated and harmonised. During this step, the harmonisation is especially important as the normals are often misaligned. This misalignment occurs because the CT data is a combination of several individual scans. For alignment of all normals, a point inside the patient's head is chosen manually as a reference point, followed by orienting all normals in the direction of this point and subsequently inverting all normals to the outside of the head (see Fig. 5).

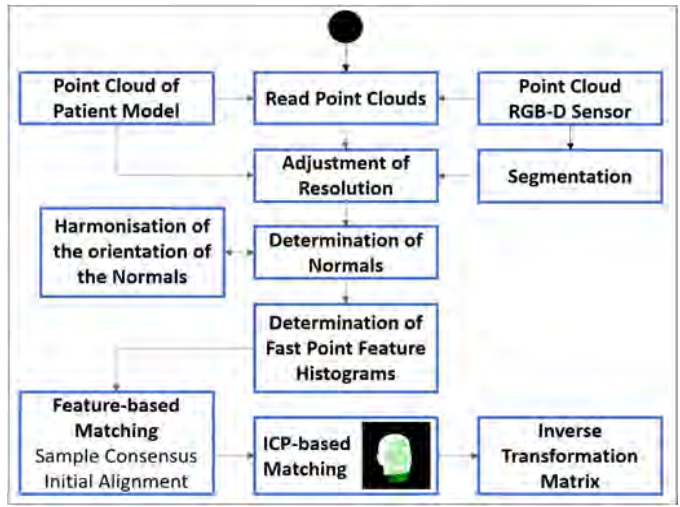

**Fig. 4.** Surface matching process.

After the preprocessing steps, the first surface fitting step is executed. It is based on the Initial alignment algorithm proposed by Rusu et al. [8]. An implementation within the point cloud library (PCL) is used. Therefore fast point feature histograms need to be calculated as a preprocessing step. In the last step an iterative closest point (ICP) algorithm is used to refine the surface matching result. After the two point clouds have been aligned to each other the inverse transformation matrix can be calculated to get the correct transformation from marker system to patient model coordinate space.

## 2.3 Catheter Tracking

As outlined in Fig. 6, catheter tracking was implemented based on semantic segmentation using a Full-Resolution Residual Network (FRRN) [7]. After the semantic segmentation of the RGB stream of the Kinect cameras, the image is fused with the depth stream to determine the voxels in the point cloud belonging to the catheter. As a further step a density based clustering approach [2] is performed on the chosen voxels. This is due to noise especially on the edges of the instrument voxels in the point cloud. Based on the found clusters an estimation of the three dimensional structure of the catheter is performed. For this purpose, a narrow cylinder with variable length is constructed. The length is changed accordingly to the semantic segmentation and the clustered voxels of the point cloud. The approach is applicable to identify a variety of instruments.

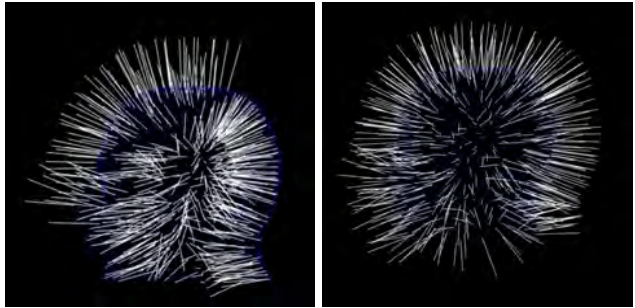

**Fig. 5.** Before (left) and after (right) harmonisation of wrongly oriented normals.

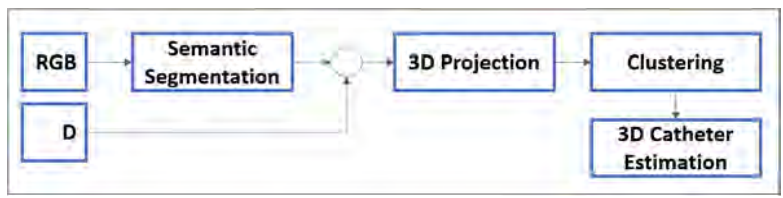

**Fig. 6.** Process to track the catheter using a RGB-D sensor.

## 2.4 Tracking of Medical Staff

The OpenPose [1] library is used to track key points on the bodies of the medical staff. Available ROS nodes have been modified to integrate OpenPose in the OP:Sense ROS environment. The architecture is outlined in Fig. 7.

## 3 Results

In this chapter the results of the patient, catheter and medical staff tracking are described. The approach to find the coarse position of a patient's head was performed on a phantom head placed on the OR table within OP:Sense. Multiple scenarios with changing illumination and occlusion conditions were recorded. The results are depicted in Fig. 8 and the evaluation results are depicted in Table 1.

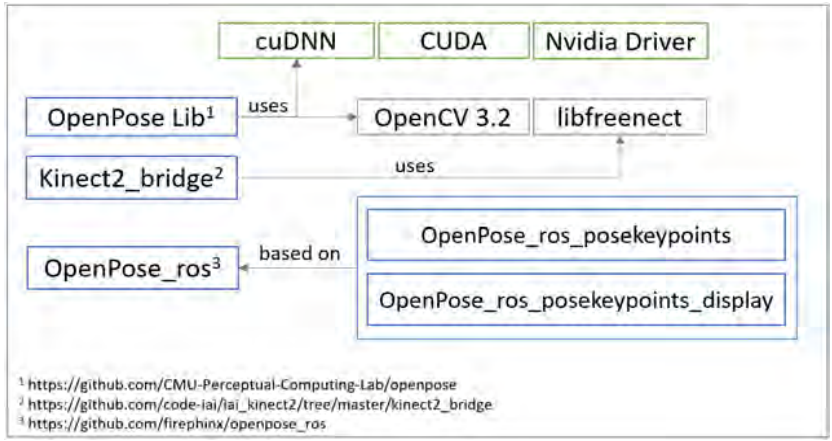

**Fig. 7.** Integration of the OpenPose library into the ROS environment of OP:Sense.

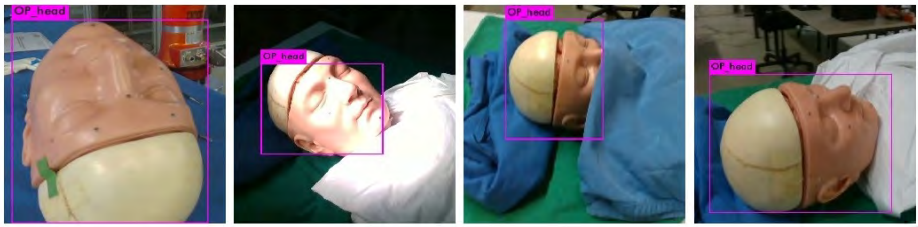

**Fig. 8.** Coarse determination of the patient's head on the OR table with YOLO v3.

**Table 1.** Evaluation results for the phantom search with YOLO v3.

|  | Precision | Recall | F1-Score | average IoU | mAP |
|---|---|---|---|---|---|
| **Normal OR conditions** | 92% | 99% | 95% | 67.59% | 90.35% |
| **Occlusion** | 99% | 93% | 96% | 75.77% | 90.86% |
| **Strong Illumination** | 62% | 66% | 64% | 41.01% | 62.23% |
| **Illumination and Occlusion** | 65% | 51% | 57% | 41.83% | 45.62% |

Precision detection of the patient was performed with a two-stage surface matching approach. Different point cloud resolutions were tested with regard to runtime behaviour. Voxel grid edge sizes of 6, 4 and 3 mm have been tested, with a higher edge size corresponding to a smaller point cloud. The matching results of the two point clouds were analyzed manually. An average accuracy of 4.7 mm was found with an accuracy range between 3.0 and 7.0 mm.

In the first stage of the surface matching, the two point clouds are coarsely aligned as depicted in Fig. 9. In the second step ICP is used for fine adjustment. A two-stage process was implemented as ICP requires a good initial alignment of the two point clouds.

**Table 2.** Evaluation results of the two-step surface matching.

| Voxel Grid Size (mm) | Points | Points after Adjustment of Resolution | Processing Time (in minutes) | Mean Accuracy (mm) |
|---|---|---|---|---|
| 6 | 17418 | 1293 | 1.98 | 6.45 |
| 4 | 17418 | 2706 | 7.57 | 3.52 |
| 3 | 17418 | 4405 | 17.87 | 4.12 |
| 4 | 47530 | 2860 | 7.53 | 3.94 |
| 4 | 49917 | 2774 | 7.71 | 3.0 |
| 4 | 30521 | 3035 | 8,09 | 7.0 |

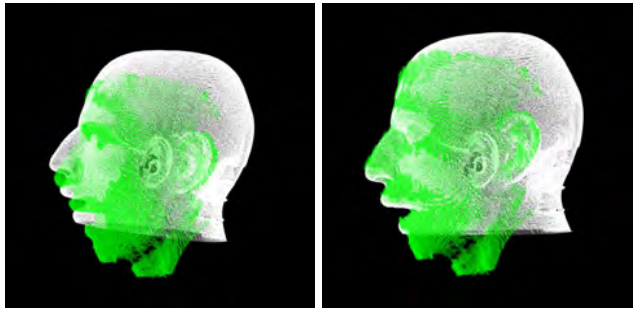

**Fig. 9.** Results of the two-stage surface matching. Left: After Initial Alignment, Right: After ICP.

For catheter tracking a precision of the semantic segmentation between 47% and 84% is reached (see Table 3). Tracking of instruments, especially neurosurgical catheters, are challenging due to their thin structure and non-rigid shape. Detailed results on catheter tracking have been presented in [7].

**Table 3.** Evaluation results for semantic segmentation [7].

| Dataset | Normal conditions 1 | Normal conditions 2 | Catheter horizontal | Catheter vertical | Catheter diagonal | Bright | Bright, patient covered |
|---|---|---|---|---|---|---|---|
| **Precision** | 84.1% | 64.0% | 77.1% | 47.7% | 81.0% | 59.0% | 72.9% |
| **Recall** | 58.7% | 61.9% | 51.2% | 19.2% | 31.0% | 15.4% | 43.1% |

The 3D estimation of the catheter is shown in Fig. 10. The catheter was moved in front of the camera and the 3D reconstruction was recorded simultaneously. Over a long period of the recording over 90% of the catheter are tracked correctly. In some situations this drops to under 50% or lower.

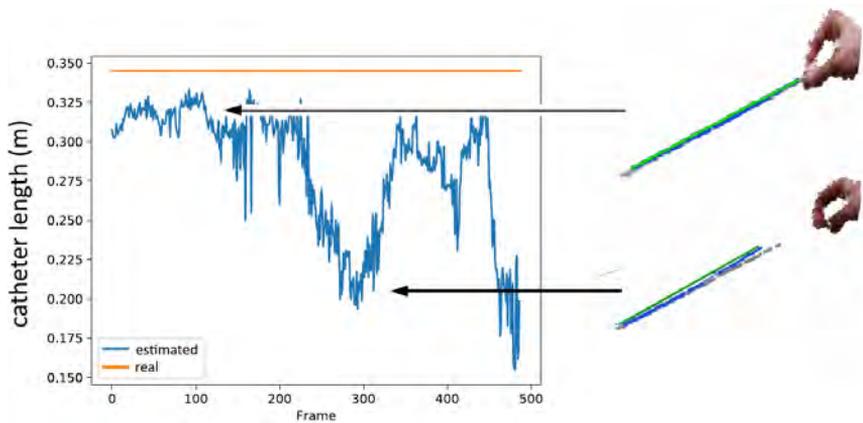

**Fig. 10.** Results of the 3D estimation of the catheter.

The tracking of medical personnel is shown in Fig. 11. The different body parts and joint positions are determined, e.g. the head, eyes, shoulders, elbows, etc. The library yielded very good results as described in [1]. We reached a performance of 21 frames per second on a workstation (Intel i7-9700k, GeForce 1080 Ti) processing 1 stream.

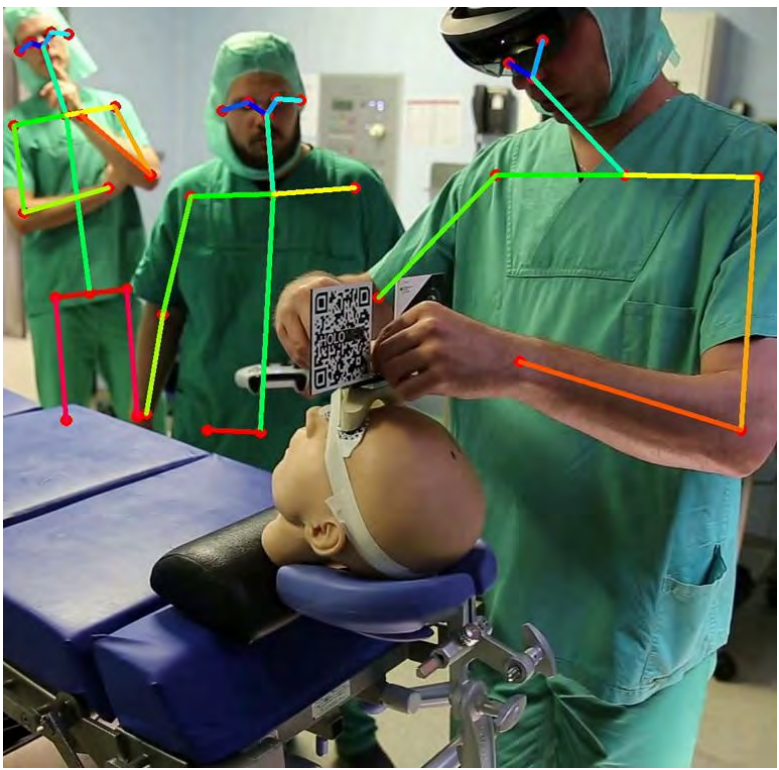

**Fig. 11.** Results of the medical staff tracking.

# 4    Discussion

As shown in the evaluation, our approach succeeds in detecting the patient in an automated two-stage process with an accuracy between 3 and 7 mm. The coarse position is determined by using a YOLO v3 net. The results under normal OR conditions are very satisfying. The solution performance drops strongly under bright illumination conditions. This is due to large flares that occur on the phantom as it is made of plastic or silicone. However, these effects do not occur on human skin. The advantage of our system is that the detection is performed on all four Kinect RGB streams enable different views on the operation area. Unfavourable illumination conditions normally don't occur on all of these streams. Therefore a robust detection is still possible. In the future the datasets will be expanded with samples with strong illumination conditions.

The following surface matching of the head yields good results and a robust and precise detection of the patient. Most important is a good preprocessing of the CT data and the recorded point cloud of the search area, as described in the methods. The algorithm does not manage to find a result if there are larger holes in the point clouds or if the normals are not calculated correctly. Additionally, challenges that have to be considered include skin deformities and noisy CT data. The silicone skin is not fixed to the skull (as human skin is), which leads to changes in position, some of which are greater than 1 cm. Also the processing time of 7 minutes is quite long and must be optimized in the future. The processing time may be shortened by reducing the size of the point clouds. However, in this case the matching results may also become worse.

Catheter tracking [7] yielded good results, despite the challenging task of segmenting a very thin ( 2.5 mm) and deformable object. Additionally, a 3D estimation of the catheter was implemented. The results showed that in many cases over 90% of the catheter can be estimated correctly. However, these results strongly depend on the orientation and the quality of the depth stream. Using higher quality sensors could improve the detection results.

For tracking of the medical staff OpenPose as a ready-to-use people detection algorithm was used and integrated into ROS. The library produces very good results, despite medical staff wearing surgical clothing.

# 5    Conclusion

In this work the integration of augmented reality into the digital operating room OP:Sense is demonstrated. This makes it possible to expand the capabilities of current AR glasses. The system can determine the precise patient's position by implementing a two-stage process. First a YOLO v3 net is used to coarsly detect the patient to reduce the search area. In a second subsequent step a two-stage surface matching process is implemented for refined detection. This approach allows for precise location of the patient's head for later tracking.

Further, a FRNN-based solution to track the surgical instruments in the OR was implemented and demonstrated on a thin neurosurgical catheter for ventricular punctures. Additionally, OpenPose was integrated into the digital OR to track the surgical personnel. The presented solution will enable the system to react to the current situation in the operating room and is the base for an integration into the surgical workflow.

# 6    Acknowledgement

This work was partially funded by the Federal Ministry of Education and Research within the project 'HoloMed - Context-sensitive support of a surgeon in the operating room by Augmented Reality'. The authors acknowledge the expertise provided by neurosurgeons from the Department of Neurosurgery at the University Hospital Ulm/Günzburg.

# Scene-Adaptive Disparity Estimation in Depth Sensor Networks Using Articulated Shape Models

Samuel Zeitvogel, Johannes Wetzel, and Astrid Laubenheimer

Intelligent Systems Research Group (ISRG)
Karlsruhe University of Applied Sciences
{samuel.zeitvogel,johannes.wetzel,astrid.laubenheimer}@hs-karlsruhe.de

**Abstract.** In this work a scene-adaptive approach for disparity estimation in depth sensor networks is presented. Our approach makes use of a priori scene knowledge to improve the low-level 3D data acquisition. We fit an articulated shape model to the given 3D data and leverage the resulting high-level scene information to make the estimation of disparities more robust to local ambiguities. We present early qualitative results to show the applicability of our method.

**Keywords:** scene-adaptive sensor networks, guided stereo matching, context-ware disparity estimation, articulated human shape models, stereo matching.

## 1    Introduction

Due to the emergence of commodity depth sensors many classical computer vision tasks are employed on networks of multiple depth sensors e.g. people detection [1] or full-body motion tracking [2]. Existing methods approach these applications using a sequential processing pipeline where the depth estimation and inference are performed on each sensor separately and the information is fused in a post-processing step. In previous work [3] we introduce a scene-adaptive optimization schema, which aims to leverage the accumulated scene context to improve perception as well as post-processing vision algorithms (see Fig. 1). In this work we present a proof-of-concept implementation of the scene-adaptive optimization methods proposed in [3] for the specific task of stereo-matching in a depth sensor network. We propose to improve the 3D data acquisition step with the help of an articulated shape model, which is fitted to the acquired depth data. In particular, we use the known camera calibration and the estimated 3D shape model to resolve disparity ambiguities that arise from repeating patterns in a stereo image pair. The applicability of our approach can be shown by preliminary qualitative results.

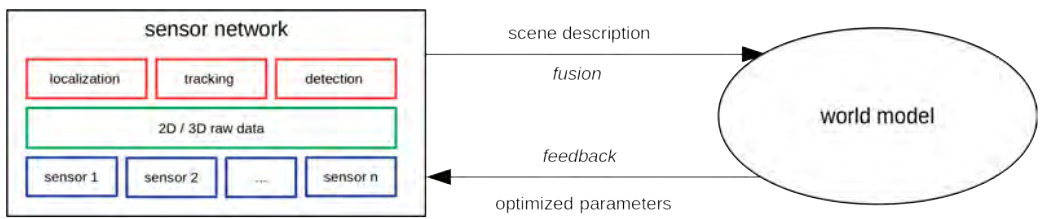

Fig. 1: Scene-Adaptive optimization introduced by [3].

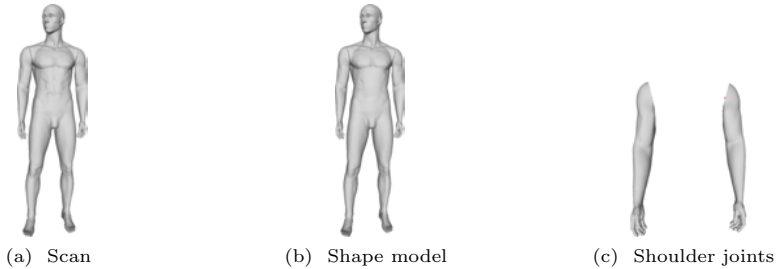

(a) Scan         (b) Shape model        (c) Shoulder joints

Fig. 2: Laser scan (a), articulated shape model (b) and corresponding shoulder joints (c).

## 2   Related Work

In previous work [3] we introduce a general framework for scene-adaptive optimization of depth sensor networks. It is suggested to exploit inferred scene context by the sensor network to improve the perception and post-processing algorithms themselves. In this work we apply the proposed ideas in [3] to the process of stereo disparity estimation, also referred to as stereo matching.

While stereo matching has been studied for decades in the computer vision literature [4, 5] it is still a challenging problem and an active area of research. Stereo matching approaches can be categorized into two main categories, *local* and *global* methods. While local methods, such as block matching [6], obtain a disparity estimation by finding the best matching point on the corresponding scan line by comparing local image regions, global methods formulate the problem of disparity estimation as a global energy minimization problem [7]. Local methods lead to highly efficient real-time capable algorithms, however, they suffer from local disparity ambiguities. In contrast, global approaches are able to resolve local ambiguities and therefore provide high-quality disparity estimations. But they are in general very time consuming and without further simplifications not suitable for real-time applications.

The semi-global matching (SGM) introduced by Hirschmuller [8] aggregates many feasible local 1D smoothness constraints to approximate global disparity smoothness regularization. SGM and its modifications are still offering a remarkable trade-off between the quality of the disparity estimation and the run-time performance.

More recent work from Poggi et al. [9] focuses on improving the stereo matching by taking additional high-quality sources (e.g. LiDAR) into account. They propose to leverage sparse reliable depth measurements to improve dense stereo matching. The sparse reliable depth measurements act as a prior to the dense disparity estimation. The proposed approach can be used to improve more recent end-to-end deep learning architectures [10, 11], as well as classical stereo approaches like SGM.

This work is inspired by [9], however, our approach does not rely on an additional LiDAR sensor but leverages a priori scene knowledge in terms of an articulated shape model instead to improve the stereo matching process.

## 3   Approach

### 3.1   Experimental Setup

We set up four stereo depth sensors with overlapping fields of view. The sensors are extrinsically calibrated in advance, thus their pose with respect to a world coordinates system is known. The stereo sensors are pointed at a mannequin and capture eight greyscale images (one image pair for each stereo sensor, the left image of each pair is depicted in Fig. 3a). For our experiments we use a high-quality laser scan of the mannequin as ground truth. We assume that the proposed algorithm has access to an existing shape model that can express the observed geometry of the scene in some capacity. In our experimental setup, we assume a shape model of a mannequin with two articulated shoulders and a slightly different shape in the belly area of the mannequin (see Fig. 2). In the remaining section we use the provided shape model to improve the depth data generation of the sensor network.

## 3.2 Semi-global Block Matching

First, we estimate the disparity values of each of the four stereo sensors with SGM without using the human shape model. Let $p$ denote a pixel and $q$ denote an adjacent pixel. Let $d$ denote a disparity map and $d_p, d_q$ denote the disparity at pixel location $p$ and $q$. Let $\mathcal{P}$ denote the set of all pixels and $\mathcal{N}$ the set of all adjacent pixels. Then the SGM cost function can be defined as

$$E(d) = \sum_{p \in \mathcal{P}} D(p, d_p) + \sum_{(p,q) \in \mathcal{N}} R(p, d_p, q, d_q), \qquad (1)$$

where $D(p, d_p)$ denotes the matching term (here the sum of absolute differences in a $7 \times 7$ neighborhood) which assigns a matching cost to the assignment of disparity $d_p$ to pixel $p$ and $R(p, d_p, q, d_q)$ penalizes disparity discontinuities between adjacent pixels $p$ and $q$. In SGM the objective given in (1) is minimized with dynamic programming, leading to the resulting disparity map

$$\bar{d} = \arg \min_d E(d). \qquad (2)$$

As input for the shape model fitting we apply SGM on all four stereo pairs leading to four disparity maps as depicted in Fig. 4a.

## 3.3 Shape Model Fitting

To be able to exploit the articulated shape model for stereo matching we initial need to fit the model to the 3D data obtained by classical SGM as described in 3.2. To be more robust to outliers we do only use disparity values from pixels with high contrast and transform them into 3D point clouds. Since we assume that the relative camera poses are known, it is straight forward to merge the resulting point clouds in one world coordinate system. Finally the shape model is fitted to the merged point cloud by optimizing over the shape model parameters, namely the pose of the model and the rotation of the shoulder joints. We use an articulated mannequin shape model in this work as a proxy for an articulated human shape model (e.g. [2]) as proof-of-concept and plan to transfer the proposed approach on real humans in future work.

## 3.4 Synthetic Disparity Maps

Once the model parameters of the shape model are obtained we can reproject the model fit to each sensor view by making use of the known projection matrices. Fig. 3b shows the rendered wireframe mesh of the fitted model as an overlay on the camera images. For our guided stereo matching approach we then need the synthetic disparity map which can be computed from the synthetic depth maps (a byproduct of 3D rendering). We denote the synthetic disparity image by $d^{\text{synth}}$. One synthetic disparity image is created for each stereo sensor, see Fig. 4b.

### 3.5 Model-based Guided Stereo Matching

In the final step we exploit the existing shape model fit, in particular the synthetic disparity image $d^{\text{synth}}$ of each stereo sensor and combine it with SGM (inspired by guided stereo matching [9]). Our augmented objective is defined as

$$E'(d) = \sum_{p \in \mathcal{P}} D'(p, d_p) + \sum_{(p,q) \in \mathcal{N}} R(p, d_p, q, d_q), \tag{3}$$

with

$$D'(p, d_p) = \begin{cases} D(p, d_p) & \text{if } |d_p^{\text{synth}} - d_p| \le 3 \\ \infty & \text{else.} \end{cases} \tag{4}$$

The introduced objective is very similar to SGM and can be minimized in a similar fashion leading to the final disparity estimation in our scene-adaptive depth sensor network

$$\hat{d} = \arg\min_d E'(d). \tag{5}$$

To summarize our approach, we exploit an articulated shape model fit to enhance SGM with minor adjustments.

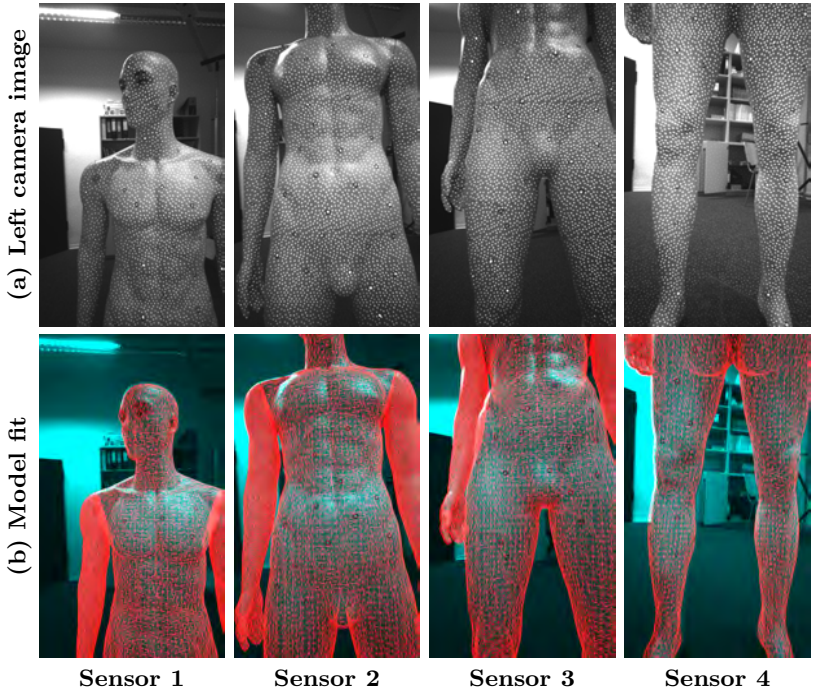

Fig. 3: Left camera images (a) and fitted shape model projected on camera images (b).

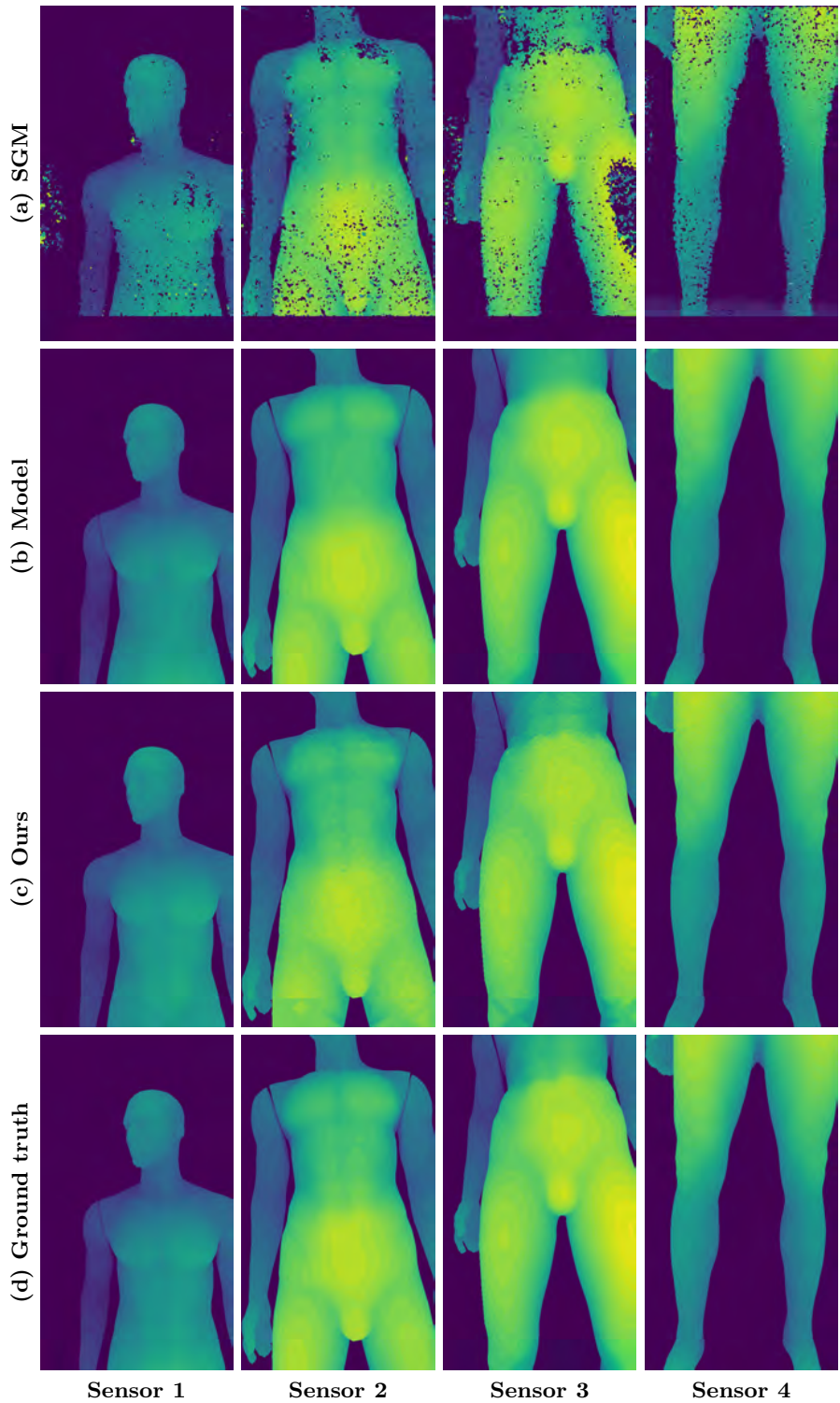

Fig. 4: Comparison of obtained disparity maps for all sensor views.

## 4    Qualitative Evaluation

To show the applicability of our approach we present preliminary qualitative results. The results are depicted in Fig. 4. Using SGM without exploiting the provided articulated shape model leads to reasonable results, but the disparity map is very noisy and no clean silhouette of the mannequin is extracted (see Fig. 4a). Fitting our articulated shape model to the data leads to clean synthetic disparity maps as shown in Fig. 4c, with a clean silhouette. In the belly area the synthetic model disparity map (Fig. 4b) does not agree with the ground truth (Fig. 4d). The articulated shape model is not general enough to explain the recorded scene faithfully. Using the guided stereo matching approach, we construct a much cleaner disparity map than SGM. In addition, the approach takes the current sensor data into account and exploits an existing articulated shape model.

## 5    Conclusion

In this work we have proposed a method for scene-adaptive disparity estimation in depth sensor networks. Our main contribution is the exploitation of a fitted human shape model to make the estimation of disparities more robust to local ambiguities. Our early results indicate that our method can lead to more robust and accurate results compared to classical SGM. Future work will focus on a quantitative evaluation as well as incorporating sophisticated statistical human shape models into our approach.

# Inverse Process-Structure-Property Mapping

Norbert Link[1], Tarek Iraki[1], and Johannes Dornheim[1]

Intelligent Systems Research Group,
Karlsruhe University of Applied Sciences
`[tarek.iraki, johannes.dornheim, norbert.link]`@hs-karlsruhe.de

**Abstract.** Workpieces for dedicated purposes must be composed of materials which have certain properties. The latter are determined by the compositional structure of the material. In this paper, we present the scientific approach of our current DFG funded project Tailored material properties through microstructural optimization: Machine Learning Methods for the Modeling and Inversion of Structure-property relationships and their application to sheet metals. The project proposes a methodology to automatically find an optimal sequence of processing steps which produce a material structure that bears the desired properties. The overall task is split in two steps: First find a mapping which delivers a set of structures with given properties and second, find an optimal process path to reach one of these structures with least effort. The first step is achieved by machine learning the generalized mapping of structures to properties in a supervised fashion, and then inverting this relation with methods delivering a set of goal structure solutions. The second step is performed via reinforcement learning of optimal paths by finding the processing sequence which leads to the best reachable goal structure. The paper considers steel processing as an example, where the microstructure is represented by Orientation Density Functions and elastic and plastic material target properties are considered. The paper shows the inversion of the learned structure-property mapping by means of Genetic Algorithms. The search for structures is thereby regularized by a loss term representing the deviation from process-feasible structures. It is shown how reinforcement learning is used to find deformation action sequences in order to reach the given goal structures, which finally lead to the required properties.

**Keywords:** Computational Materials Science, Property-Structure-Mapping, Texture Evolution Optimization, Machine Learning, Reinforcement Learning

## 1 Introduction

The derivation of processing control actions to produce materials with certain, desired properties is the "inverse problem" of the causal chain "process control" - "microstructure instantiation" - "material properties". The main goal of our current project is the creation of a new basis for the solution of this problem by using modern approaches from machine learning and optimization.

The inversion will be composed of two explicitly separated parts: "Inverse Structure-Property-Mapping" (SPM) and "microstructure evolution optimization". The focus of the project lies on the investigation and development of methods which allow an inversion of the structure-property-relations of materials relevant in the industry. This inversion is the basis for the design of microstructures and for the optimal control of the related production processes. Another goal is the development of optimal control methods yielding exactly those structures which have the desired properties. The developed methods will be applied to sheet metals within the frame of the project as a proof of concept. The goals include the development of methods for inverting technologically relevant "Structure-Property-Mappings" and methods for efficient microstructure representation by supervised and unsupervised machine learning. Adaptive processing path-optimization methods, based on reinforcement learning, will be developed for adaptive optimal control of manufacturing processes.

We expect that the results of the project will lead to an increasing insight into technologically relevant process-structure-property-relationships of materials. The instruments resulting from the project will also promote the economically efficient development of new materials and process controls.

## 2 Related Work

In general, approaches to microstructure design make high demands on the mathematical description of microstructures, on the selection and presentation of suitable features, and on the determination of structure-property relationships. For example, the increasingly advanced methods in these areas enable Microstructure Sensitive Design (MSD), which is introduced in [1] and [2] and described in detail in [3].

The relationship between structures and properties descriptors can be abstracted from the concrete data by regression in the form of a Structure-Property-Mapping. The idea of modeling a Structure-Property-Mapping by means of regression and in particular using artificial neural networks was intensively pursued in the 1990s [4] and is still used today. The approach and related methods presented in [5] always consist of a Structure-Property-Mapping and an optimizer (in [5] Genetic Algorithms) whose objective function represents the desired properties.

The inversion of the SPM can be alternatively reached via Generative Models. In contrast to discriminative models (e.g. SPM), which are used to map conditional dependencies between data (e.g. classification or regression), generative models map the composite probabilities of the variables and can thus be used to generate new data from the assumed population. Established, generative methods are for example Mixture Models [6], Hidden Markov Models [7] and in the field of artificial neural networks Restricted Boltzmann Machines [8]. In the field of deep learning, generative models, in particular generative adversarial networks [9], are currently being researched and successfully applied in the context of image processing. Conditional Generative Models can generalize the probability of occurrence of structural features under given material properties. In this way, if desired, any number of microstructures could be generated.

Based on the work on the SPM, the process path optimization in the context of the MSD is treated depending on the material properties. For this purpose, the process is regarded as a sequence of structure-changing process operations which correspond to elementary processing steps. Shaffer et al. [10] construct a so called *texture evolution network* based on process simulation samples, to represent the process. The texture evolution network can be considered as a graph with structures as vertices, connected by elementary processing steps as edges. The structure vertices are points in the structure-space and are mapped to the property-space by using the SPM for property driven process path optimization. In [11] one-step deformation processes are optimized to reach the most reachable element of a texture-set from the inverse SPM. Processes are represented by so called *process planes*, principal component analysis (PCA) projections of microstructures reachable by the process. The optimization then is conducted by searching for the process plane which best represents one of the texture-set elements. In [12], a generic ontology based semantic system for processing path hypothesis generation (MATCALO) is proposed and showcased.

# 3 Research Concept

## 3.1 Inverse Structure-Property-Mapping (SPM)

The required mapping of the structures to the properties is modeled based on data from simulations. The simulations are based on Taylor models. The structures are represented using textures in the form of orientation density functions (ODF), from which the properties are calculated. In the investigations, elastic and plastic properties are considered in particular. Structural features are extracted from the ODF for a more compact description. The project uses spectral methods such as generalized spherical harmonics (GSH) to approximate the ODF. As an alternative representation we investigate the discretization in the orientation-space, where the orientation density is represented by a histogram.

The solution of the inverse problem consists of a Structure-Property-Mapping and an optimizer: As [4] described, the SPM is modeled by regression using artificial neural networks. In this investigation, we use a multilayer perceptron.

Differential evolution (DE) is used for the optimization problem. DE is an evolutionary algorithm developed by Rainer Storn and Kenneth Price [13]. It is a optimization method, which repeatedly improves a candidate solution set under consideration of a given quality measure over a continuous domain. The DE algorithm optimizes a problem by taking a population of candidate solutions and generating new candidate solutions (structures) by mutation and recombination existing ones. The candidate solution with the best fitness is considered for further processing. So, for the generated structures the reached properties are determined using the SPM.

The fitness $\mathcal{F}$ is composed of two terms: The property loss $\mathcal{L}_\mathcal{P}$, which expresses, how close the property of a candidate is to the target property, and the structure loss $\mathcal{L}_\mathcal{S}$, which represents the degree of feasibility of the candidate structure in the process

$$\mathcal{F}(\vec{s}, \hat{\vec{s}}, \vec{p_r}, \vec{p_d}) = \mathcal{L}_\mathcal{P}(\vec{p_r}, \vec{p_d}) + \mathcal{L}_\mathcal{S}(\vec{s}, \hat{\vec{s}}), \qquad (1)$$

The property loss is the mean squared error (MSE) between the reached properties $p_r \in \vec{p_r}$ and the desired properties $p_d \in \vec{p_d}$:

$$\mathcal{L}_\mathcal{P}(\vec{p_r}, \vec{p_d}) = \frac{1}{N} \sum_{i=1}^{N} (p_{r_i} - p_{d_i})^2 \qquad (2)$$

Considering the goal that the genetic algorithm generates reachable structures, a neural network is formed which functions as an anomaly detector. The data basis of this neural network are structures that can be reached by a process. The goal of anomaly detection is to exclude unreachable structures. The anomaly detection is implemented using an autoencoder [14]. This is a neural network (see Fig. 1) which consists of the following two parts: the encoder and the decoder. The encoder converts the input data to an embedding space. The decoder converts the embedding space as close as possible to the original data. Due to the reduction to an embedding space, the autoencoder uses data compression and extracts relevant features. The cost function for the structures is a distance function in the ODF-space, which penalizes the network if it produces outputs that differ from the input. The cost function is also known as the *reconstruction loss*:

$$\mathcal{L}_{\mathcal{S}}(\vec{s}, \hat{\vec{s}}) = \sum_{i=1}^{N} \frac{(s_i - \hat{s}_i)^2}{(s_i + \hat{s}_i + \lambda)}, \tag{3}$$

with $s_i \in \vec{s}$ as the original structures, $\hat{s}_i \in \hat{\vec{s}}$ as the reconstructed structures and $\lambda = 0.001$ to avoid division by zero.

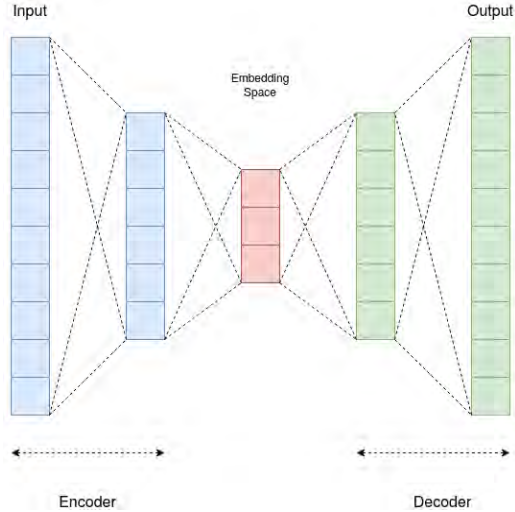

**Fig. 1.** Autoencoder for determining the structure loss

When using the anomaly detection, the autoencoder determines a high reconstruction loss if the input data are structures that are very different from the reachable structures. The overall approach is shown in Fig. 2 and consists of the following steps:

1. The genetic algorithm generates structures.
2. The SPM determines the reached properties of the generated structures.
3. The structure loss $\mathcal{L}_{\mathcal{S}}$ is determined by the reconstruction loss of the anomaly detector for the generated structures with respect to the reachable structures.
4. The property loss $\mathcal{L}_{\mathcal{P}}$ is determined by the MSE of the reached properties and the desired properties.
5. The fitness is calculated as the sum of the structure loss $\mathcal{L}_{\mathcal{S}}$ and the property loss $\mathcal{L}_{\mathcal{P}}$.

The structures, resulting from the described approach form the basis for optimal process control.

## 3.2  Texture Evolution Optimization

Due to the forward mapping, the process evolution optimization based on texture evolution networks ([10]) is restricted to a-priori sampled process paths. [11] relies on linearization assumptions and is applicable to short process sequences only. [12] relies on a-priori learned process models in the form of regression trees and is also applicable to relatively short process sequences only.

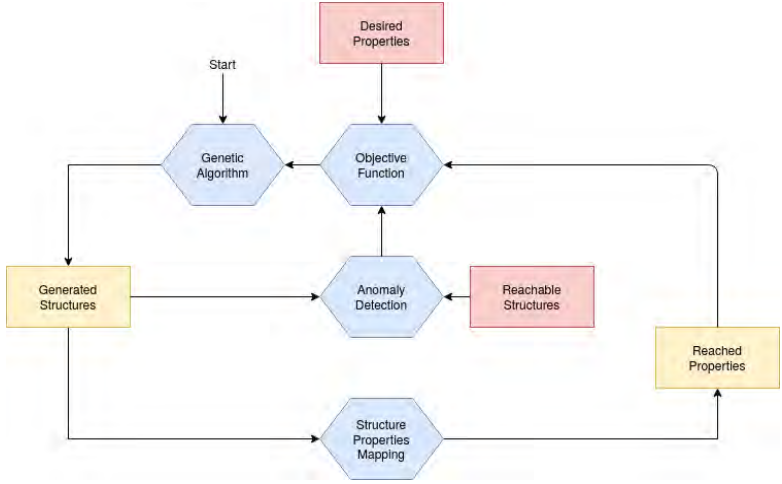

**Fig. 2.** Overview of our approach to find reachable structures with desired properties

As an adaptive alternative for texture evolution optimization, that can be trained to find process-paths of arbitrary length, we propose methods from reinforcement learning. For desired material properties $\vec{p_d}$. The inverted SPM determines a set of goal microstructures $\vec{s_d} \in G$, which are very likely reachable by the considered deformation process. The texture evolution optimization objective is then to find the shortest process path $\mathcal{P}^*$ starting from a given structure $\vec{s_0}$, and leading close to one of the structures from $G$.

$$\mathcal{P}^* = \arg\min |\mathcal{P}| : E(s_0, \mathcal{P}) \in G_\tau, \qquad (4)$$

where $\mathcal{P} = (a_k)_{k=0,\dots,K}; K \leqq T$ is a path of process actions $a$, $T$ is the maximum allowed process length. The mapping $E(s, \mathcal{P}) = s_k$ delivers the resulting structure, when applying $\mathcal{P}$ to the structure $s$. Here, for the sake of simplicity, we assume the process to be deterministic, although the reinforcement learning methods we use are not restricted to deterministic processes. $G_\tau$ is a neighbourhood of $G$, the union of all open balls with radius $\tau$ and center points from $G$.

To solve the optimization problem by reinforcement learning approaches, it must be reformulated as markov decision process (MDP), which is defined by the tuple $(S, A, P, R)$. In our case $S$ is the space of structures $\vec{s}$, $A$ is the parameter-space of the deformation process, containing process actions $\vec{a}$, $P : S \times A \mapsto S$ is the transition function of the deformation process, which we assume to be deterministic. $R_g : S \times A \mapsto \mathbb{R}$ is a goal-specific reward function. The objective of the reinforcement learning agent is then to find the optimal goal-specific policy $\pi_g^*(s_t) = a_t$ that maximizes the discounted future goal-specific reward

$$\mathcal{V}_g(s_t) = \sum_{k=t}^{K} \gamma^{k-t} R_g(s_k, a_k), \qquad (5)$$

where $\gamma \in [0, 1]$ discounts early attained rewards, the policy $\pi_g(s_k)$ determines $a_k$ and the transition function $P(s_k, a_k)$ determines $s_{k+1}$.

For a distance function $d$ in the structure space, the binary reward function

$$R_g(s,a) = \begin{cases} 1, & \text{if } d(P(s,a),g) < \tau \\ 0, & \text{otherwise} \end{cases} \tag{6}$$

if maximized, leads to an optimal policy $\pi_g^*$ that yields the shortest path to $g$ from every $s$ for $\gamma < 1$. Moreover, if $\mathcal{V}_g$ is given for every microstructure from $G$, $\mathcal{P}$ from eq. 4 is identical with the application of the policy $\pi_\zeta^*$, where $\zeta = \arg\max_g[\mathcal{V}_g]$.

$\pi_g^*$ can be approached by methods from reinforcement learning. Value-based reinforcement learning is doing so by learning expected discounted future reward functions [15]. One of these functions is the so called *value-function* $V$. In the case of a deterministic MDP and for a given $g$, this expectation value function reduces to $\mathcal{V}_g$ from eq. 4 and $\zeta$ can be extracted if $V$ is learned for every $g$ from $G$. For doing so, a generalized form of expectation value functions can be learned as it is done e.g. in [16].

This exemplary MDP formulation shows how reinforcement learning can be used for texture evolution optimization tasks. The optimization thereby is operating in the space of microstructures and does not rely on a-priori microstructure samples. When using off-policy reinforcement learning algorithms and due to the generalization over goal-microstructures, the functions learned while solving a specific optimization task can be easily transferred to new optimization tasks (i.e. different desired properties or even a different property space).

# A New Approach to Gesture Based Real-Time Robot Programming Using Mixed Reality

Luisa Hornung, Simon Lawo, and Christian Wurll

Karlsruhe University of Applied Sciences
`luisa.hornung@hs-karlsruhe.de`
`christian.wurll@hs-karlsruhe.de`

**Abstract.** While being increasingly used in larger industry companies, industrial robots have not yet prevailed in smaller enterprises. Not least, this is due to the time-consuming programming and the requirement for robotics experts. An intuitive control and programming concept can decisively reduce the need for expert knowledge. Using modern rendering software and innovative visualization frameworks, a gesture-based programming approach was developed at Karlsruhe University of Applied Sciences. Here, the user creates the robot program by virtually executing and chaining robot poses and gripper instructions. An evaluation shows the advantages of the developed concept and a comparison with competing methods.

**Keywords:** Robot Programming; Virtual Robot; Mixed Reality; 3D-Engine; Small and Medium-sized Enterprises

## 1   Introduction

Industrial robots are mainly deployed in large-scale production, especially in the automotive industry. Today, there are already 26.1 industrial robots deployed per 1,000 employees on average in these industry branches. In contrast, Small and Medium-sized Enterprises (SMEs) only use 0.6 robots per 1,000 employees [1]. Reasons for this low usage of industrial robots in SMEs include the lack of flexibility with great variance of products and the high investment expenses due to additional peripherals required, such as gripping or sensor technology. The robot as an incomplete machine accounts for a fourth of the total investment costs [2]. Due to the constantly growing demand of individualized products, robot systems have to be adapted to new production processes and flows [3]. This development requires the flexibilization of robot systems and the associated frequent programming of new processes and applications as well as the adaption of existing ones. Robot programming usually requires specialists who can adapt flexibly to different types of programming for the most diverse robots and can follow the latest innovations. In contrast to many large companies, SMEs often have no in-house expertise and a lack of prior knowledge with regard to robotics. This often has to be obtained externally via system integrators, which, due to high costs, is one of the reasons for the inhibited use of robot systems. During the initial generation or extensive adaption of process flows with industrial robots, there is a constant risk of injuring persons and damaging the expensive hardware components. Therefore, the programs have to be tested under strict safety precautions and usually in a very slow test mode. This makes the programming of new processes very complex and therefore time- and cost-intensive.

The concept presented in this paper combines intuitive, gesture-based programming with simulation of robot movements. Using a mixed reality solution, it is possible to create a simulation-based visualization of the robot and project, to program and to test it in the working environment without disturbing the workflow. A virtual control panel enables the user to adjust, save and generate a sequence of specific robot poses and gripper actions and to simulate the developed program. An interface to transfer the developed program to the robot controller and execute it by the real robot is provided.

The paper is structured as follows. First, a research on related work is conducted in Section 2, followed by a description of the system of the gesture-based control concept in Section 3. The function of robot positioning and program creation is described in Section 4. Last follow the evaluation in Section 5 and conclusion in Section 6.

## 2   Related Work

Various interfaces exist to program robots, such as Lead-Trough, Offline or Walk-Trough programming, Programming by demonstration, vision based programming or vocal commanding. In the survey of Villani et al. [4] a clear overview on existing interfaces for robot programming and current research is provided. Besides the named interfaces, the programming of robots using a virtual or mixed reality solution aims to provide intuitiveness, simplicity and accessibility of robot programming for non-experts. Designed for this purpose, Guhl et al. [5] developed a generic architecture for human-robot interaction based on virtual and mixed reality. In the marker tracking based approach presented by [6] and [7], the user defines a collision-free-volume and generates and selects control points while the system creates and visualizes a path through the defined points. Others [8], [9], [10] and [11] use handheld devices in combination with gesture control and motion tracking. Herein, the robot can be controlled through gestures, pointing or via the device, while the path, workpieces or the robot itself are visualized on several displays. Other gesture and virtual or mixed reality based concepts are developed by Cousins et al. [12] or Tran et al. [13]. Here, the robots perspective or the robot in the working environment is presented to the user on a display (head-mounted or stationary) and the user controls the robot via gestures. Further concepts using a mixed reality method enable an image of the workpiece to be imported into CAD and the system automatically generates a path for robot movements [14] or visualizing the intended motion of the robot on the Microsoft HoloLens, that the user knows where the robot will move to next [15]. Other methods combine pointing at objects on an screen with speech instructions to control the robot [16]. Sha et al. [17] also use a virtual control panel in their programming method, but for adjusting parameters and not for controlling robots. Another approach pursues programming based on cognition, spatial augmented reality and multimodal input and output [18], where the user interacts with a touchable table.

Krupke et al. [19] developed a concept in which humans can control the robot by head orientation or by pointing, both combined with speech. The user is equipped with a head-mounted display presenting a virtual robot superimposed over the real robot. The user can determine pick and place position by specifying objects to be picked by head orientation or by pointing. The virtual robot then executes the potential pick movement and after the user confirms by voice command, the real robot performs the same movement. A similar concept based on gesture and speech is persued by Quintero et al. [20], whose method offers two different types of programming. On the one hand, the user can determine a pick and place position by head orientation and speech commands. The system automatically generates a path which is displayed to the user, can be manipulated by the user and is simulated by a virtual robot. On the other hand, it is possible to create a path on a surface by the user generating waypoints. Ostanin and Klimchik [21] introduced a concept to generate collision-free paths. The user is provided with virtual goal points that can be placed in the mixed reality environment and between which a path is automatically generated. By means of a virtual menu, the user can set process parameters such as speed, velocity etc.. Additionally, it is possible to draw paths with a virtual device and the movement along the path is simulated by a virtual robot.

Differently to the concept described in this paper, only a pick and place task can be realized with the concepts of [19] and [20]. A differentiation between movements to positions and gripper commands as well as the movement to several positions in succession and the generation of a program structure are not supported by these concepts. Another distinction is that the user only has the possibility to show certain objects to the robot, but not to move the robot to specific positions. In [19] a preview of the movement to be executed is provided, but the entire program (pick and place movements) is not simulated. In contrast to [21], with the concept presented in this paper it is possible to integrate certain gripper commands into the program. With [21] programming method, the user can determine positions but exact axis angles or robot poses cannot be set.

Overall, the approach presented in this paper offers an intuitive, virtual user interface without the use of handheld devices (cf. [6], [7], [8], [9], [10] and [11]) which allows the exact positions of the robot to be specified. Compared to other methods, such as [12], [13], [14], [15] or [16], it is possible to create more complex program structures, which include the specification of robot poses and gripper positions, and to simulate the program in a mixed reality environment with a virtual robot.

# 3 Mixed Reality Robot Programming System

In this section the components of the Mixed Reality Robot Programming System are introduced and described. The system consists of multiple real and virtual interactive elements, whereby the virtual components are projected directly into the field of view using a Mixed Reality (MR) approach. Compared to the real environment, which consists entirely of real objects and virtual reality (VR), which consists entirely of virtual objects and which overlays the real reality, in MR the real scene here is preserved and only supplemented by the virtual representations [22]. In order to interact in the different realities, head-mounted devices similar to glasses, screens or mobile devices are often used. Figure 1 provides an overview of the systems components and their interaction.

The system presented in this paper includes KUKAs collaborative, lightweight robot LBR iiwa 14 R820 combined with an equally collaborative gripper from Zimmer as real components and a virtual robot model and a user interface as virtual components. The virtual components are presented on the Microsoft HoloLens. For calculation and rendering the robot model and visualization of the user interface, the 3D- and physics-engine of the Unity3D development framework is used. Furthermore, for supplementary functions, components and for building additional MR interactable elements, the Microsoft Mixed Reality Toolkit (MRTK) is utilized.

For spatial positioning of the virtual robot, marker tracking is used, a technique supported by the Vuforia framework. In this use case, the image target is attached to the real robot's base, such that in MR the virtual robot superimposes the real robot. The program code is written in C♯.

The robot is controlled and programmed via an intuitive and virtual user interface that can be manipulated using the so-called Airtap gesture, a gesture provided by Microsoft HoloLens.

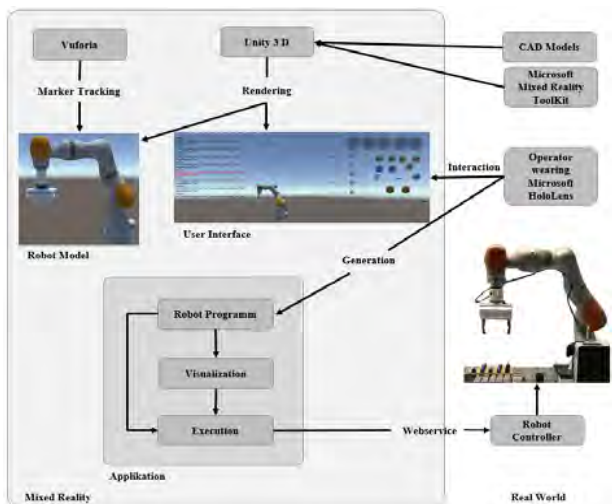

**Fig. 1.** Diagram of the systems components and their interaction.

### 3.1 Virtual Robot Model

To ensure that the virtual robot mirrors the motion sequences and poses of the real robot, the most exact representation of the real robot is employed. The virtual robot consists of a total of eight links, matching the base and the seven joints of iiwa 14 R820:

- the base frame,
- five joint modules,
- the central hand and
- the media flange.

The eight links are connected together as a kinematic chain. The model is provided as open source files from [23] and [24] and is integrated into the Unity3D project.

The individual links are created as GameObjects in a hierarchy, with the base frame defining the top level and are limited similar to those of the real robot. The CAD data of the deployed gripping system is also imported into Unity3D and linked to the robot model.

### 3.2 User Interface

The canvas of the head-up displayer of the Microsoft HoloLens is divided into two parts and rendered at a fixed distance in front of the user and on top of the scene. At the top left side of the screen the current joint angles (A1 to A7) are displayed and on the left side the current program is shown. This setting simplifies the interaction with the robot as the informations do not behave like other objects in the MR scene, but are attached to the Head Up Display (HUD) and move with the user's field of view. The user interface, which consists of multiple interactable components, is placed into the scene and is shown at the right side of the head-up display.

At the beginning of the application the user interface is in "Clear Screen" mode, i.e. only the buttons "Drag", "Cartesian", "Joints", "Play" and "Clear Screen" and the joint angles at the top left of the screen are visible. For interaction with the robot, the user has to switch into a particular control mode by tapping the corresponding button.
The user interface provides three different control modes for positioning the virtual robot:

 − Drag Mode, for rough positioning,
 − Cartesian Mode, for Cartesian positioning and
 − Joint Mode, for the exact adjustment of each joint angle.

Figure 2 shows the interactable components that are visible and therefore controllable in the respective control modes.

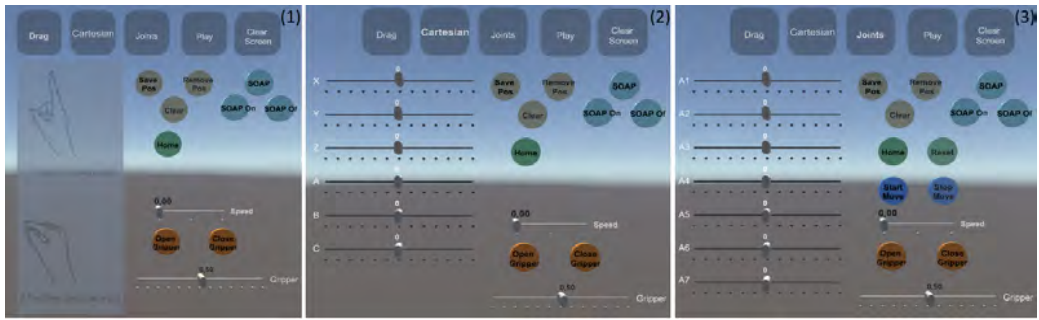

**Fig. 2.** User interface control modes and visible interactable elements: (1): Drag Mode; (2): Cartesian Mode; (3): Joints Mode.

Depending on the selected mode, different interactable components become visible in the user interface, with whom the virtual robot can be controlled. In addition to the control modes, the user interface offers further groups of interactable elements:

 − Motion Buttons, with which e.g. the speed of the robot movement can be adjusted or the robot movement can be started or stopped,
 − Application Buttons, to save or delete specific robot poses, for example,
 − Gripper Buttons, to adjust the gripper and
 − Interface Buttons, that enable communication with the real robot.

## 4  Usage

This section focuses on the description of the usage of the presented approach. In addition to the description of the individual control modes, the procedure for creating a program is also described. As outlined in Section 3.2, the user interface consists of three different control modes and four groups of further interactable components. Through this concept, the virtual robot can be moved efficiently to certain positions with different movement modes, the gripper can be adjusted, the motion can be controlled and a sequence of positions can be chained.

### 4.1  Control Modes

**Drag**  By gripping the tool of the virtual robot with the Airtap gesture, the user can "drag" the robot to the required position. Additionally, it is possible to rotate the position of the robot using both hands. This mode is particularly suitable for moving the robot very quickly to a certain position.

**Cartesian** This mode is used for the subsequent positioning of the robot tool with millimeter precision. The tool can be translated to the required positions using the Cartesian coordinates X, Y, Z and the Euler angles A, B, C. The user interface provides a separate slider for each of the six translation options.The tool of the robot moves analogously to the respective slider button, which the user can set to the required value.

**Joints** This mode is an alternative to the Cartesian method for exact positioning. The joints of the virtual robot can be adjusted precisely to the required angle, which is particularly suitable for e.g. bypassing an obstacle. There is a separate slider for each joint of the virtual robot. In order to set the individual joint angles, the respective slider button is dragged to the required value, which is also displayed above the slider button for better orientation.

## 4.2 Programming the Robot

To program the robot, the user interface provides various application buttons, such as saving and removing robot poses from the chain and a display of the poses in the chain. The user directs the virtual robot to the desired position and confirms using the corresponding button. The pose of the robot is then saved as joint angles from A1 to A7 and one gripper position in a list and is displayed on the left side of the screen. When running the programmed application, the robot moves to the saved robot poses and gripper positions according to the defined sequence. For a better orientation, the robots current target position changes its color from white to red. After testing the application, the list of robot poses can be sent to the controller of the real robot via a webservice. The real robot then moves analogously to the virtual robot to the corresponding robot poses and gripper positions.

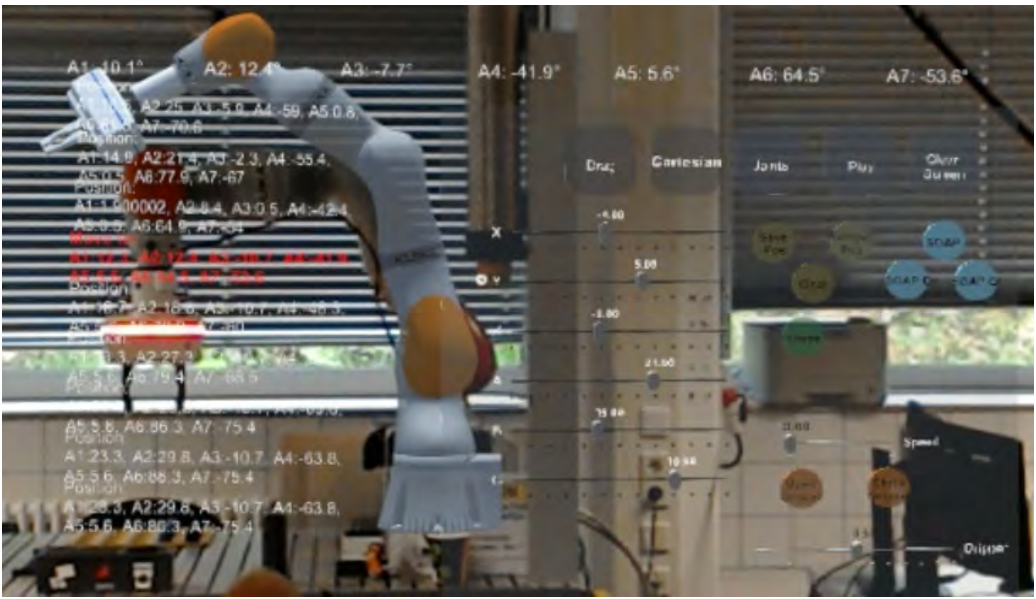

**Fig. 3.** The user's view in the mixed reality environment.

# 5 Evaluation of the Programming Concept

The purpose of the evaluation is how the gesture-based control concept compares to other concepts regarding intuitiveness, comfort and complexity. For the evaluation, a study was conducted with seven test persons, who had to solve a pick and place task with five different operating concepts and subsequently evaluate them. The developed concept based on gestures and MR was evaluated against a lead through procedure, programming with Java, programming with a simplified programming concept and approaching and saving points with KUKA SmartPad. The test persons had no experience with Microsoft HoloLens and MR, no to moderate experience with robots and no to moderate programming skills. The Questionnaire for the Evaluation of Physical Assistive Devices (QUEAD) developed by Schmidtler et al [25] was used to evaluate and compare the five control concepts. The questionnaire is classified into five categories (perceived usefulness, perceived ease of use, emotions, attitude and comfort) and contains a total of 26 questions, rated on an ordinal scale from 1 (entirely disagree) to 7 (entirely agree).

Firstly, each test person received a short introduction to the respective control concept, conducted the pick and place task and immediately afterwards evaluated the respective control concept using QUEAD.

## 5.1 Results of the QUEAD

Figure 4 provides an extract of the results of the study.

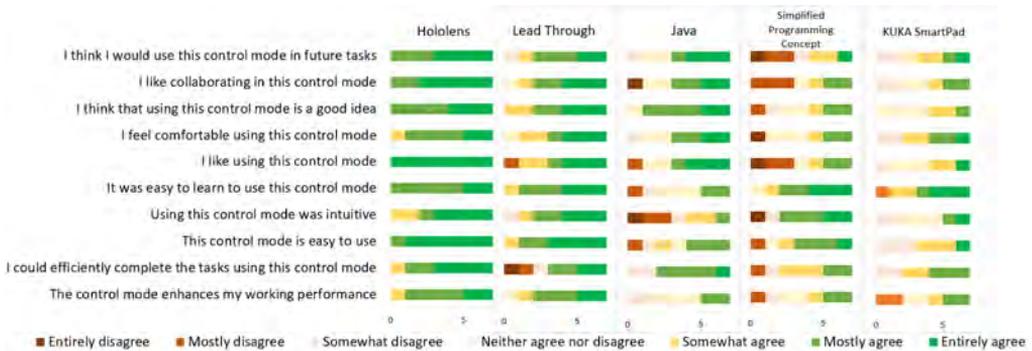

**Fig. 4.** Extract from the results of the study to compare the control concepts (from left to right): gesture-based control concept, Lead Through, Java, Simplified Programming Concept, KUKA SmartPad.

All test persons agreed that they would reuse the concept in future tasks (3 mostly agree, 4 entirely agree). In addition, the test persons considered the gesture-based concept to be intuitive (1 mostly agree, 4 entirely agree), easy to use (5 mostly agree, 2 entirely agree) and easy to learn (1 mostly agree, 6 entirely agree). Two test persons mostly agree and four entirely agree that the gesture-based concept enabled them to solve the task efficiently and four entirely agree and two entirely agree that the concept enhances their work performance. All seven subjects were comfortable using the gesture-based concept (4 mostly agree, 2 entirely agree).

Overall, the concept presented in this paper was evaluated as more comfortable, more intuitive and easier to learn than the other control concepts evaluated. In comparison to them, the new operating concept was perceived as the most useful and easiest to use. The test persons felt physically and psychologically most comfortable when using the concept and were most positive in total.

# 6   Conclusion

In this paper, a new concept for programming robots based on gestures and MR and for simulating the created applications was presented. This concept forms the basis for a new, gesture-based programming method, with which it is possible to project a virtual robot model of the real robot into the real working environment by means of a MR solution, to program it and to simulate the workflow. Using an intuitive virtual user interface, the robot can be controlled by three control modes and further groups of interactable elements and via certain functions, several robot positions can be chained as a program. By using this concept, test and simulation times can be reduced, since on the one hand the program can be tested directly in the MR environment without disturbing the workflow. On the other hand, the robot model is rendered into the real working environment via the MR approach, thus eliminating the need for time-consuming and costly modeling of the environment.

The results of the user study indicate that the control concept is easy to learn, intuitive and easy to use. This facilitates the introduction of robots and especially in SMEs, since no expert knowledge is required for programming, programs can be created rapidly and intuitively and processes can be adapted flexibly. In addition, the user study showed that tasks can be solved efficiently and the concept is perceived as performance-enhancing.

Potential directions of improvement are: Implement various movement types, such as point-to-point, linear and circular movements in the concept. This makes the robot motion more flexible and efficient, since positions can be approached in different ways depending on the situation. Another improvement is to extend the concept with collaborative functions of the robot, such as force sensitivity or the ability to conduct search movements. In this way, the functions that make collaborative robots special can be integrated into the program structure. A further approach for improvement is to engage in a larger scale study.

# 7   Acknowledgement

This research and development project is funded by the German Federal Ministry of Education and Research (BMBF) and the European Social Fund (ESF) within the program "Future of work" (02L17C550) and implemented by the Project Management Agency Karlsruhe (PTKA). The author is responsible for the content of this publication.

# Classification of Maritime Vessels using Convolutional Neural Networks

Mathias Anneken[1]*, Moritz Strenger[1]*, Sebastian Robert[2], and Jürgen Beyerer[1][2]

[1] Vision and Fusion Laboratory (IES), Karlsruhe Institute of Technology (KIT)
*equal contribution, alphabetical order
mathias.anneken@kit.edu, moritz.strenger@alumni.kit.edu
[2] Fraunhofer IOSB, Karlsruhe; Fraunhofer Center for Machine Learning
sebastian.robert@iosb.fraunhofer.de, juergen.beyerer@iosb.fraunhofer.de

**Abstract** Due to a steady increase in traffic at sea, the need for support in sur­veillance task is growing for coast guards and other law enforcement units all over the world. An important cornerstone is a reliable vessel classification, which can be used for detecting criminal activities like illegal, unreported and unregulated fishing or smuggling operations. As many ships are required to transmit their po­sition by using the automatic identification system (AIS), it is possible to generate a large dataset containing information on the world wide traffic. This dataset is used for implementing deep neural networks based on residual neural networks for classifying the most common shiptypes based on their movement patterns and geographical features. This method is able to reach a competitive result. Further, the results show the effectiveness of residual networks in time-series classification.

**Keywords:** residual neural network, time series classificaiton, convolutional neural networks, maritime domain, ship classification

## 1 Introduction

In 2019 the world's commercial fleet consists of 95,402 ships with a total capacity of 1,976,491 thousand dwt. (a plus of 2.6 % in carrying capacity compared to last year) [1]. According to the International Chamber of Shipping, the shipping industry is responsible for about 90 % of all trade [2]. In order to ensure the safe voyage of all participant in the international travel at sea, the need for monitoring is steadily increasing.

While more and more data regarding the sea traffic is collected by using cheaper and more powerful sensors, the data still needs to be processed and understood by human operators. In order to support the operators, reliable anomaly detection and situation recognition systems are needed. One cornerstone for this development is a reliable auto­matic classification of vessels at sea.

For example by classifying the behaviour of non cooperative vessels in ecological protected areas, the identification of illegal, unreported and unregulated (IUU) fishing activities is possible. IUU fishing is in some areas of the world a major problem, e. g., »in the wider-Caribbean, Western Central Atlantic region, IUU fishing compares to 20-30 percent of the legitimate landings of fish« [3] resulting in an estimated value between USD 700 and 930 million per year.

One approach for gathering information on the sea traffic is based on the automatic identification system (AIS)[3]. It was introduced as a collision avoidance system. As each vessel is broadcasting its information on an open channel, the data is often used for other purposes, like training and validating of machine learning models.

---

[3] https://gpsd.gitlab.io/gpsd/AIVDM.html

AIS provides dynamic data like position, speed and course over ground, static data like MMSI[4], shiptype and length, and voyage related data like draught, type of cargo, and destination about a vessel.

The system is self-reporting, it has no strong verification of transmission, and many of the fields in each message are set by hand. Therefore, the data can not be fully trusted. As Harati-Mokhtari et al. [4] stated, half of all AIS messages contain some erroneous data. As for this work, the dataset is collected by using the AIS stream provided by AISHub[5], the dataset is likely to have some amount of false data. While most of the errors will have no further consequences, minor coordinate inaccuracies or wrong vessel dimensions are irrelevant, some false information in vessel information can have an impact on the model performance.

## 2   Related Work

Classification of maritime trajectories and the detection of anomalies is a challenging problem, e.g., since classifications should be based on short observation periods, only limited information is available for vessel identification. Riveiro et al. [5] give a survey on anomaly detection at sea, where shiptype classification is a subtype.

Jiang et al. [6] present a novel TrajectoryNet capable of point-based classification. Their approach is based on the usage of embedding GPS coordinates into a new feature space. The classification itself is accomplished using an long short-term memory (LSTM) network.

Further, Jiang et al. [7] propose a partition-wise LSTM (PLSTM) for point-based binary classification of AIS trajectories into fishing or non-fishing activity. They evaluated their model against other recurrent neural networks and achieve a significantly better result than common recurrent network architectures based on LSTM or gated recurrent units.

A recurrent neural network is used by Nguyen et al. in [8] to reconstruct incomplete trajectories, detect anomalies in the traffic data and identify the real type of a vessel. They are embedding the position data to generate a new representation as input for the neural network.

Besides these neural network based approaches, other methods are also used for situation recognition tasks in the maritime domain. Especially expert-knowledge based systems are used frequently, as illegal or at least suspicious behaviour is not recorded as often as desirable for deep learning approaches.

Conditional Random Fields are used by Hu et al. [9] for the identification of fishing activities from AIS data. The data has been labelled by an expert and contains only longliner fisher boats.

Saini et al. [10] propose an hidden Markov model (HMM) based approach to the classification of trajectories. They combine Global-HMM and Segmental-HMM using a genetic algorithm. In addition, they tested the robustness of the framework by adding Gaussian noise.

In [11] Fischer et al. introduce a holistic approach for situation analysis based on Situation-Specific Dynamic Bayesian Networks (SSDBN). This includes the modelling of the SSDBN as well as the presentation to end-users. For a Bayesian Network, the parametrisation of the conditional probability tables is crucial. Fischer introduces an algorithm for choosing these parameters in a more transparent way. Important for the functionality is the ability of the network to model the domain knowledge and the handling of noisy input data. For the evaluation, simulated and real data is used to assess the detection quality of the SSDBN.

---

[4] "Maritime Mobile Service Identity" as unique identifier for the association of data
[5] https://www.aishub.net

Based on DBNs, Anneken et al. [12] implemented an algorithm for detecting illegal diving activities in the North Sea. As explained by de Rosa et al. [13] an additional layer for modelling the reliability of different sensor sources is added to the DBN.

## 3  Preprocessing

In order to use the AIS data, preprocessing is necessary. This includes cleaning wrong data, filtering data, segmentation, and calculation of additional features. The whole workflow is depicted in Figure 1. The input in form of AIS data and different maps is shown as blue boxes. All relevant MMSIs are extracted from the AIS data. For each MMSI, the position data is used for further processing. Segmentation into separate trajectories is the next step (yellow). The resulting trajectories are filtered (orange). Based on the remaining trajectories, geographic (green) and trajectory (purple) based features are derived. For each of the resulting sequences, the data is normalized (red), which results in the final dataset. Only the 6 major shiptypes in the dataset are used for the evaluation. These are "Cargo", "Tanker", "Fishing", "Passenger", "Pleasure Craft" and "Tug". Due to their similar behaviour, "Cargo" and "Tanker" will combined to a single class "Cargo-Tanker".

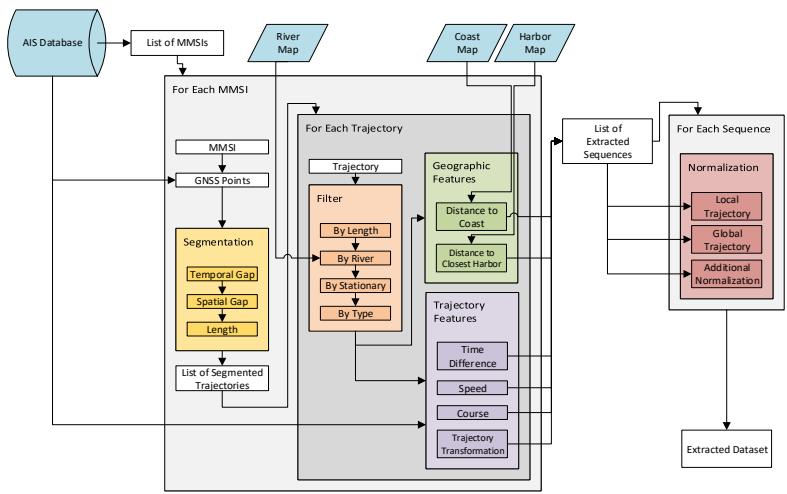

Figure 1: Visualization of all preprocessing steps. Input in blue, segmentation in yellow, filtering in orange, geographic features in green, trajectory feature in purple and normalization in red.

### 3.1  Trajectory features

Four different trajectory features are used:

  – Time difference
  – Speed over ground
  – Course over ground
  – Trajectory transformation

As the incoming data from AIS is not necessarily uniformly distributed in time, there is a need to create a feature representing the time dimension. Therefore, the time difference between two samples is introduced.

As the speed and course over ground is directly accessible through the AIS data, the network will be directly fed with these features. The vessel's speed is a numeric value in 0.1-knot resolution in the interval $[0; 1022]$ and the course is the negative angle in degrees relative to true north and therefore in the interval $[0; 359]$.

The position will be transformed in two ways. The first transformation, further called "relative-to-first", will shift the trajectory to start at the origin. The second transformation, henceforth called "rotate-to-zero", will rotate the trajectory, in such a way, that the end point is on the x-axis.

## 3.2 Geographic features

Additional to the trajectory based features, two geographic features are derived by using coastline maps[6] and a map of large harbours. The coastline map consists of a list of line strips. In order to reduce complexity, the edge points are used to calculate the "Distance-To-Coast". Further, only a lower resolution of the shapefile itself is used. In Figure 2, the resolution "high" and "low" for some fjords in Norway are shown. Due to the geoindex' cell size set to 40 km, a radius of 20 km can be queried.

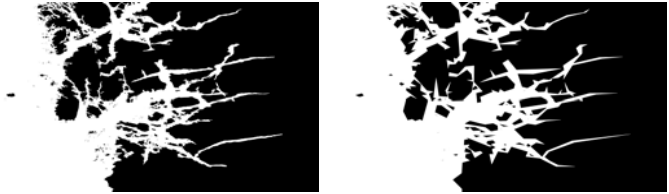

Figure 2: Comparison between the "high" (left) and "low" (right) resolution of the coastline.

The world's 140 major harbours based on the world port index[7] are used to calculate the "Distance-to-Closest-Harbor". As fishing vessels are expected to stay near to a certain harbour, this feature should support the network to identify some shiptypes. The geoindex' cell size is set for this feature to 5,000 km, resulting in a maximum radius of 2,500 km.

## 3.3 Segmentation

The data is split into separate trajectories by using gaps in either time or space, or the sequence length. As real AIS data is used, package loss during the transmission is common. This problem is tackled by splitting the data

---

[6] http://www.soest.hawaii.edu/wessel/gshhg/
[7] https://msi.nga.mil/Publications/WPI

- if the time between two successive samples is larger than 2 hours, or
- if the distance between two successive samples is large.

Regarding the distance, even though the great circle distance is more accurate, the euclidean distance is used. For simplification the distance value is squared and as a threshold $10^{-4}$ is used. Depending on latitude this corresponds to a value of about 1 km at the equator and only about 600 m at 60° N. Since the calculation includes approximation a relatively high threshold is chosen.

As the neural network depends on a fixed input size, the data is split into fitting chunks by cutting and padding with these rules:

- Longer sequences are split into chunks according to the desired sequence length.
- Any left over sequence shorter than 80 % of the desired length is discarded.
- The others will be padded with zeroes.

This results in segmented trajectories of similar but not necessarily same duration.

## 3.4 Filter

As this work is about the vessel behaviour at sea, stationary vessels (anchored and moored vessels) and vessels traversing rivers are removed from the segmented trajectories. The stationary vessels are identified by using a measure of movement in a trajectory:

$$\alpha_{\text{stationary}} = \frac{\sum_{i=1}^{n} |P_i - P_{i-1}|}{n}, \tag{1}$$

where $n$ as the sequence length and $P_i$ its data points. A trajectory will be removed if $\alpha_{\text{stationary}}$ is below a certain threshold.

A shapefile[8] containing the major and most minor rivers (compare ??) is used in order to remove the vessels not on the high seas. A sequence with more than 50 % of its points on a river is removed from the dataset.

## 3.5 Normalization

In order to speed up the training process, the data is normalized in the interval $[0; 1]$ by applying

$$X' = \frac{X - X_{\min}}{X_{\max} - X_{\min}}. \tag{2}$$

Here, for the positional features a differentiation between "global normalization" and "local normalization" is taken into account. The "global normalization" will scale the input data for the maximum $X_{\max}$ and minimum $X_{\min}$ calculated over the entire data set, while "local normalization" will estimate the maximum $X_{\max}$ and minimum $X_{\min}$ only over the trajectory itself. As the data is processed parallel, the parameters for the "global normalization" will be calculated only for each chunk of data. This will result in slight deviations in the minimum and maximum, but for large batches this should be neglectable.

All other additional features are normalized as well. For the geographic features "Distance-to-Coast" and "Distance-to-Closest-Harbor" the maximum distance, that can be queried depending on grid size, is used as $X_{max}$ and 0 is used as the lower bound $X_{min}$.

The time difference feature is scaled using a minimum $X_{min}$ of 0 and the threshold for the temporal gap since this is the maximum value for this feature. Speed and course are normalized using 0 and their respective maximum values.

---

[8] http://www.soest.hawaii.edu/wessel/gshhg/

## 3.6   Resulting dataset

For the dataset, a period between 2018-07-24 and 2018-11-15 is used. Altogether 209,536 unique vessels with 2,144,317,101 raw data points are included. Using this foundation and the previously described methods, six datasets are derived. All datasets use the same spatial and temporal thresholds. In addition, filter thresholds are identical as well. The datasets differentiate in their sequence length and by applying only the "relative-to-first" transformation or additionally the "rotate-to-zero" transformation. Either 360, 1,080, or 1,800 points per sequence are used resulting in approximate 1 h, 3 h, or 5 h long sequences. In Figure 3, the distribution of shiptypes in the datasets after applying the different filters is shown.

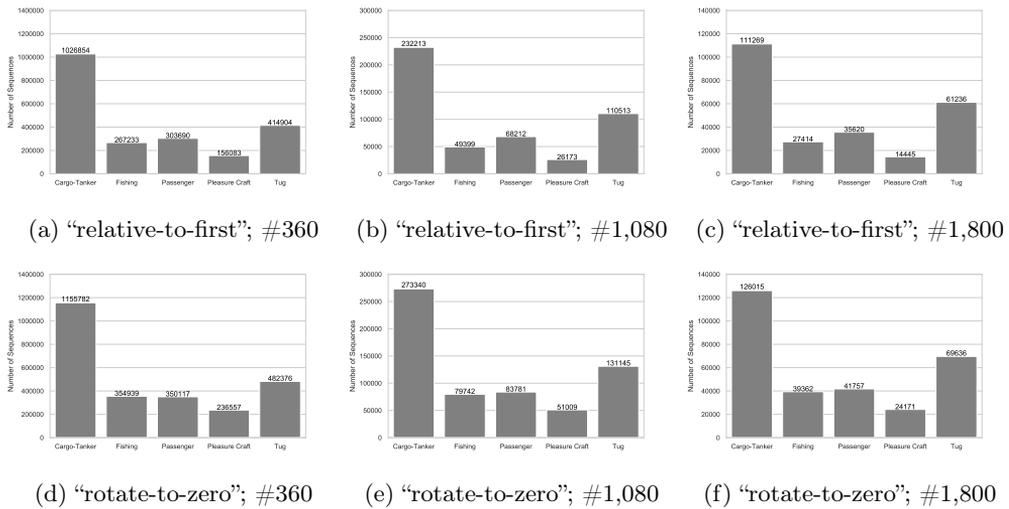

(a) "relative-to-first"; #360    (b) "relative-to-first"; #1,080    (c) "relative-to-first"; #1,800

(d) "rotate-to-zero"; #360    (e) "rotate-to-zero"; #1,080    (f) "rotate-to-zero"; #1,800

Figure 3: Number of sequences per class.

## 4   Neural Network Design

For the shiptype classification, neural networks are chosen. The different networks are implemented using Keras [14] with TensorFlow as backend [15].

Fawaz et al. [16] have shown, that, despite their initial design for image data, a residual neural network (ResNet) can perform quite well on time-series classification. Thus, as foundation for the evaluated architectures the ResNet is used. The main difference to other neural network architectures is the inclusion of "skip connections". This allows for deeper networks by circumventing the vanishing gradient problem during the training phase.

Based on the main idea of a ResNet, several architectures are designed and evaluated for this work. Some information regarding the structure are given in Table 1. Further, the single architectures are depicted in Figures 4a to 4f.

The main idea behind these architectures is to analyse the impact of the depth of the networks. Furthermore, as the features itself are not necessarily logically linked with each other, the hope is to be able to capture the behaviour better by splitting up the network path for each feature.

To verify the necessity of CNNs two multilayer perceptron (MLP) based networks are tested: One with two hidden layers and one with four hidden layers, all with 64 neurons and fully connected with their adjacent layers. The majority of the parameters for the two networks are bound in the first layer. They are necessary to map the large number of input neurons, e. g., for the 360 samples dataset $360 * 9 = 3,240$ input neurons, to the first hidden layer.

Table 1: Parameter of neural network architectures.

| Name | Depth | # Parameters |
|---|---|---|
| Tiny ResNet | 11 | 29,125 |
| Shallow ResNet | 21 | 440,837 |
| Deep ResNet | 66 | 1,327,877 |
| Stretched Deep ResNet | 66 | 3,280,645 |
| Split ResNet | 26 | 390,701 |
| Total Split ResNet | 26 | 364,461 |
| MLP for 360 samples | 4 | 211,909 |
| deeper MLP for 360 samples | 6 | 220,229 |

## 5 Training

Each of the datasets is split into three parts: 64 % for the training set, 16 % for the validation set, and 20 % for the test set. For solving or at least mitigating the problem of overfitting, regularization techniques (input noise, batch normalization, and early stopping) are used.

Small noise on the input in the training phase is used to support the generalization of the network. For each feature a normal distribution with a standard deviation of 0.01 and a mean of 0 is used as noise.

Furthermore, batch normalization is implemented. This means, before each ReLU-layer a batch normalization layer is added, allowing higher learning rates. Therefore, the initial learning rate is doubled. Additionally, the learning rate is halved if the validation error does not improve after ten training epochs, improving the training behaviour during oscillation on a plateau.

In order to prevent overfitting, an early stopping criteria is introduced. The Training will be interrupted if the validation error is not decreasing after 15 training epochs.

To counter the dataset imbalance, class weights were considered but ultimately did not lead to better classification results and were discarded.

## 6 Evaluation

The different neural network architectures are evaluated on a AMD Ryzen Threadripper 1920X 12-Core Processor with 64 GB of memory and 4x Nvidia Geforce GTX 1080 Ti. Each network is evaluated on the six datasets. For the ResNet based networks, the

(a) Tiny ResNet

(b) Shallow ResNet

(c) Deep ResNet

(d) Stretched Deep ResNet

(e) Split ResNet

(f) Split ResNet

Figure 4: Schematic architectures.

batch normalization and the input noise is tested. The initial learning rate is set to 0.001 without batch normalization and 0.002 with batch normalization activated. The maximum number of epochs is set to 600. The batch sizes are set to 64, 128, and 256 for 360, 1,080, and 1,800 samples per sequence respectively.

In total 144 different setups are evaluated. Furthermore, 4 additional networks are trained on the 360 samples dataset with "relative-to-first" transformation. Two MLPs to verify the need of deep neural networks, and the Shallow and Deep ResNet trained without geographic features to measure the impact of these features.

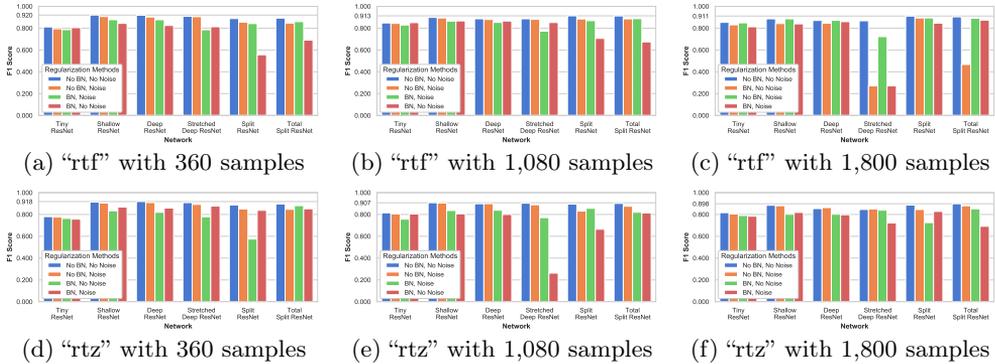

(a) "rtf" with 360 samples    (b) "rtf" with 1,080 samples    (c) "rtf" with 1,800 samples

(d) "rtz" with 360 samples    (e) "rtz" with 1,080 samples    (f) "rtz" with 1,800 samples

Figure 5: $F_1$-Scores of all networks. In addition, the regularization methods used are shown. The first row shows the results for the "relative-to-first" (rtf) transformation, the second for the "rotate-to-zero" (rtz) transformation.

The results for the six different architectures are depicted in Figure 5. For 360 samples the Shallow ResNet and the Deep ResNet outperformed the other networks. In case of the "relative-to-first" transformation (see Figure 5a), the Shallow ResNet achieved an $F_1$-Score of 0.920, while the Deep ResNet achieved 0.919. For the "rotate-to-zero" transformation (see Figure 5d), the Deep ResNet achieved 0.918 and the Shallow ResNet 0.913. In all these cases the regularization methods lead to no improvements.

The "relative-to-first" transformation performs slightly better overall. For the datasets with 360 samples per sequence, the standard ResNet variants achieve higher $F_1$-Scores compared to the Split ResNet versions. But this difference is relatively small. As expected, the Tiny ResNet is not large and deep enough to classify the data on a similar level.

For the "relative-first" transformation and trajectories based on 1080 samples (see Figure 5b), the Split ResNet and the Total Split ResNet achieve the best results. The first performed well with an $F_1$-Score of 0.913, while the latter is slightly worse with 0.912. In both cases again the regularization did not improve the result. For the "rotate-to-zero" transformation (see Figure 5e), the Shallow ResNet achieved an $F_1$-Score of 0.907 without any regularization and 0.905 with only the the noise added to the input.

For the largest sequence length of 1,800 samples, the split based networks slightly outperform the standard ResNets. For the "relative-to-first" transformation (see Figure 5c), the Split ResNet achieved an $F_1$-Score of 0.911, while for the "rotate-to-zero" transformation (see Figure 5f) the Total Split ResNet reached an $F_1$-Score of 0.898. Again without noise and batch normalization.

| | Cargo-Tanker | Fishing | Passenger | Pleasure Craft | Tug |
|---|---|---|---|---|---|
| **Cargo-Tanker** | 95.1%<br>97636 | 1.2%<br>1250 | 0.8%<br>831 | 0.8%<br>832 | 2.1%<br>2135 |
| **Fishing** | 6.2%<br>1669 | 89.1%<br>24074 | 0.8%<br>225 | 1.3%<br>343 | 2.6%<br>699 |
| **Passenger** | 5.5%<br>1700 | 0.9%<br>289 | 91.0%<br>28284 | 1.4%<br>426 | 1.2%<br>374 |
| **Pleasure Craft** | 7.9%<br>1148 | 2.7%<br>397 | 2.4%<br>356 | 84.6%<br>12296 | 2.4%<br>344 |
| **Tug** | 7.3%<br>3032 | 1.5%<br>637 | 0.7%<br>270 | 0.7%<br>284 | 89.8%<br>37301 |

_Actual_ (row axis) / **Prediction** (column axis)

Figure 6: Confusion matrix of the Shallow ResNet on 360 samples with "relative-to-first" transformation, without added input noise and batch normalization.

To verify, that the implementation of CNNs is actually necessary, additional tests with MLPs were carried out. Two different MLPs are trained on the 360 samples dataset with "relative-to-first" transformation since this dataset leads to best results for the ResNet architectures. Both networks lead to no results as their output always is the "Cargo-Tanker" class regardless of the actual input. The only thing the models are able to learn is, that the "Cargo-Tanker" class is the most probable class based on the uneven distribution of classes.

An MLP is not the right model for this kind of data and performs badly. The large dimensionality of even the small sequence length makes the use of the fully connected networks impracticable. Probably, further hand-crafted feature extraction is needed to achieve better results.

To measure the impact the feature "Distance to Coast" and "Distance to Closest Harbor" have on the overall performance, a Shallow ResNet and a Deep ResNet are trained on the 360 sample length data set with the "relative-to-first" transformation excluding these features. The trained networks have $F_1$-Scores of 0.888 and 0.871 respectively. This means, by including this features, we are able to increase the performance by 3.5 %.

## 7   Discussion

The "relative-to-first" transformation compared to the "rotate-to-zero" transformation yields the better results. Especially, this is easily visible for the longest sequence length. A possible explanation can be seen in the "stationary" filter. This filter removes more trajectories for the "relative-to-first" transformation than for the additional "rotate-to-zero" transformation. A problem might be, that the end point is used for rotating the trajectory. This adds a certain randomness to the data, especially for round trip sequences.

In some cases, the Stretched Deep ResNet is not able to learn the classes. It is possible, that there is a problem with the structure of the network or the large number of parameters. Further, there seems to be a problem with the batch normalization, as seen in Figures 5c and 5e.

The overall worse performance of the "rotate-to-zero" transformation could be because of the difference in the "stationary" filter. In the "rotate-to-zero" dataset, fewer sequences are filtered out. The filter leads to more "Fishing" and "Pleasure Craft" sequences in relation to each other as described in section 3.6. This could also explain the difference in class prediction distribution since the network is punished more for mistakes in these classes because more classes are overall from this type.

For the evaluation, the expectation based on previous work by other authors was, that the shorter sequence length should perform worse compared to the longer ones. Instead the shorter sequences outperform the longer ones. The main advantages of the shorter sequences are essentially the larger number of sequences in the dataset. For example the 360 samples dataset with "relative-to-first" transformation contains about 2.2 million sequences, while the corresponding 1,800 sample dataset contains only approximately 250,000 sequences.

In addition, the more frequent segmentation can yield more easily classifiable sequences: The behaviour of a fishing vessel in general contains different characteristics, like travelling from the harbour to the fishing ground, the fishing itself, and the way back. The travelling parts are similar to other vessels and only the fishing part is unique. A more aggressive segmentation will yield more fishing sequences, that will be easier to classify regardless of observation length.

The Shallow ResNet has the overall best results by using the 360 samples dataset and the "relative-to-first" transformation. The results for this setup are shown in the confusion matrix in Figure 6. As expected, the Tiny ResNet is not able to compete with the others. The other standard ResNet architectures performed well, especially on shorter sequences.

The Split architectures are able to perform better on datasets with longer sequences, with the Shallow ResNet achieving similar performance. Comparing the number of parameters, all three architectures have about 400,000 the Shallow ResNet about 50,000 more, the Total Split ResNet about 40,000 less.

Only on the dataset with more sequences, the Deep ResNet performs well. This correlates with the need of more information due to the larger parameter count. Due to the reduced flexibility, the Split architecture can be interpreted as a "head start". This means, that the network has already information regarding the structure of the data, which in turn does not need to be extracted from the data. This can result in a better performance for smaller datasets.

All in all, the best results are always achieved by omitting the suggested regularization methods. Nevertheless, the batch normalization had an effect on the learning rate and needed training epochs: The learning rate is higher and less epochs are needed before convergence.

## 8 Conclusion

Based on the ResNet, several architectures are evaluated for the task of shiptype classification. From the initial dataset based on AIS data with over 2.2 billion datapoints six datasets with different trajectory length and preprocessing steps are derived. Further to the kinematic information included in the dataset, geographical features are generated.

Each network architecture is evaluated with each of the datasets with and without batch normalization and input noise. Overall the best result is an $F_1$-Score of 0.920 with the Shallow ResNet on the 360 samples per sequence dataset and a shift of the trajectories to the origin. Additionally, we are able to show, that the inclusion of geographic features yield an improvement in classification quality.

The achieved results are quite promising, but there is still some room for improvement. First of all, the the sequence length used for this work might still be too long for real world use cases. Therefore, shorter sequences should be tried. Additionally, interpolation for creating data with the same time delta between two samples or some kind of embedding or alignment layer might yield better results. As there are many sources for additional domain related information, further research in the integration of these sources is necessary.

## Acknowledgments

Underlying projects to this article are funded by the WTD 81 of the German Federal Ministry of Defense. The authors are responsible for the content of this article.

This work was developed in the Fraunhofer Cluster of Excellence "Cognitive Internet Technologies".

# Comparison of CNN for the detection of small ojects based on the example of components on an assembly table

Jonas Hansert[1], Madlon Pécaut[2], and Thomas Schlegel

[1] Institute of Ubiquitous Mobility Systems, Karlsruhe University of Applied Sciences
`jonas.hansert@hs-karlsruhe.de`

[2] Institute of Ubiquitous Mobility Systems, Karlsruhe University of Applied Sciences
`madlon.pecaut@googlemail.com`

**Abstract.** This paper presents the results of a comparison of deep neural networks for detection of small objects typical for manual manufacturing tasks. We created a set of training, validation and evaluation data and selected four state of the art deep neural networks for object detection. We trained them with the same number of epochs, 200 epochs per network architecture and compared the training time, accuracy and prediction time on evaluation data. Additional we compared the neural networks on thirty images of three very small and similar components.

**Keywords:** We would like to encourage you to list your keywords within the abstract section

## 1 Use Case and Requirements

Many tasks which only a few years ago had to be performed by humans can now be performed by robots or will be performed by robots in the near future. Nevertheless, there are some tasks in assembly processes which cannot be automated in the next few years. This applies especially to workpieces that are only produced in very small series or tasks that require a lot of tact and sensitivity, such as inserting small screws into a thread or assembling small components.

In conversations with companies we have found out that a big problem for the workers is learning new production processes. This is currently done with instructions and by supervisors. But this requires a lot of time. This effort can be significantly reduced by modern systems, which accompany workers in the learning process. Such intelligent systems require not only instructions that describe the target status and the individual work steps that lead to it, but also information on the current status at the assembly workstation. One way to obtain this information is to install cameras above the assembly workstation and use image recognition to calculate where an object is located at any given moment.

The individual parts, often very small compared to the work surface, must be reliably detected. We have trained and tested several deep neural networks for this purpose.

We have developed an assembly workstation where work instructions can be projected directly onto the work surface using a projector. At a distance, 21 containers for components are arranged in three rows, slightly offset to the rear, one above the other. These containers can also be illuminated by the projector. Thus a very flexible Pick-by-Light system can be implemented. In order for the system behind it to automatically switch to the next work step and, in the event of errors, to point them out and provide support in correcting them, it is helpful to be able to identify the individual components on the work surface.

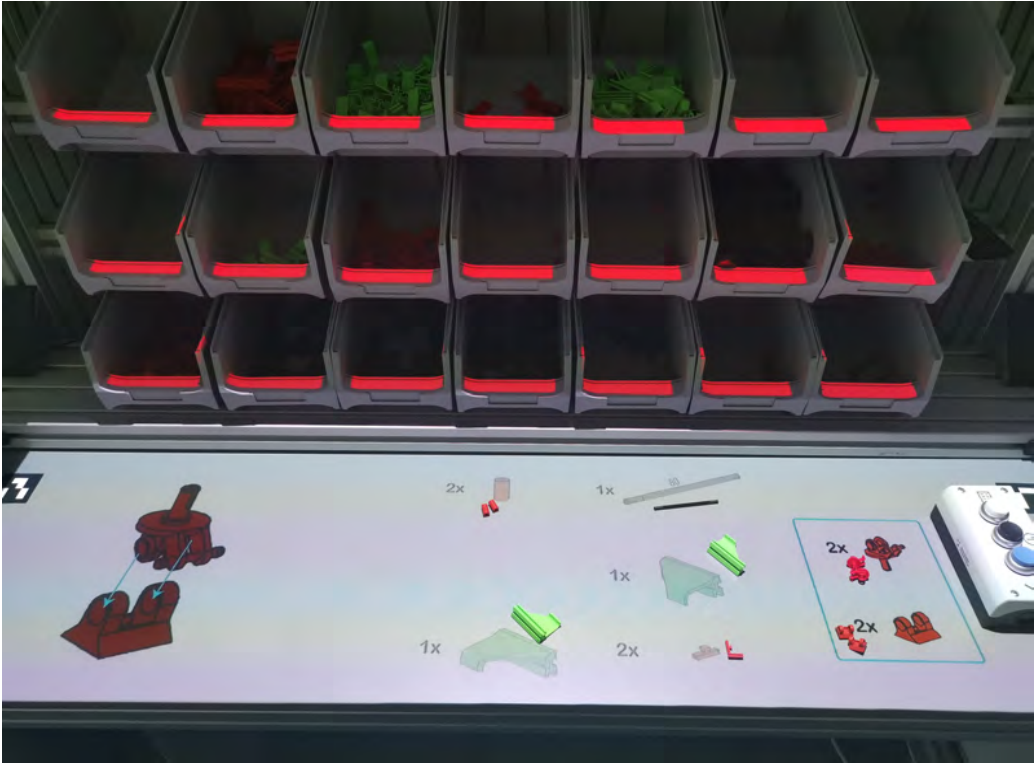

**Fig. 1.** Our assembly Workstation with a projector, a camera and 21 containers for the components.

We use a RealSense depth camera for this purpose, from which, however, we are currently only using the colour image. The camera is mounted in a central position at a height of about two meters above the work surface. Thus the camera image includes the complete working surface as well as the 21 containers and a small area next to the working surface.

The objects to be detected are components of a kit for the construction of various toy cars. The kit contains 25 components in total. Some of the components vary considerably from each other, but some others are very similar to each other. Since it is the same with real components of a production, the choice of the kit seemed appropriate for the purposes of this project.

## 2   Related Work

Object detection, one of the most fundamental and challenging problems in computer vision, seeks to local object instances from a large number of predefined categories in natural images.

Until the beginning of 2000, a similar approach was mostly used in object detection. Keypoints in one or more images of a category were searched for automatically. At these points a feature vector was generated. During the recognition process, keypoints in the image were again searched, the corresponding feature vectors were generated and compared with the stored feature vectors. After a certain threshold an object was assigned to the category. One of the first approaches based on machine learning was published by Viola and Jones in 2001 [1]. They still selected features, in their case they were selected by using a Haar basis function [2] and then using a variant of AdaBoost [3].

Starting in 2012 with the publication of AlexNet by Krizhevsky et al. [4], deep neural networks became more and more the standard in object detection tasks. They used a convolutional neural network which has 60 million parameters in five convolutional layers, some of them are followed by max-pooling layers, three fully-connected layers and a final softmax layer. They won the ImageNet LSVRC-2012 competition with a error rate almost half as high as the second best.

Inception-v2 is mostly identical to Inception-v3 by Szegedy et al. [5]. It is based on Inception-v1 [6]. All Inception architectures are composed of dense modules. Instead of stacking convolutional layers, they stack modules or blocks, within which are convolutional layers. For Inception-v2 they redesigned the architecture of Inception-v1 to avoid representational bottlenecks and have more efficient computations by using factorisation methods. They are the first using batch normalisation in object detection tasks.

In previous architectures the most significant difference has been the increasing number of layers. But with the network depth increasing, accuracy gets saturated and then degrades rapidly. Kaiming et al. [7] addressed this problem with ResNet using skip connections, while building deeper models.

In 2017 Howard et al. presented MobileNet architecture [8]. MobileNet was developed for efficient work on mobile devices with less computational power and is very fast. They used depthwise convolutional layers for a extremely efficient network architecture.

One year later Sandler et al. [9] published a second version of MobileNet. Besides some minor adjustments, a bottleneck was added in the convolutional layers, which further reduced the dimensions of the convolutional layers. Thus a further increase in speed could be achieved.

In addition to the neural network architectures presented so far, there are also different methods to detect in which area of the image the object is located. The two most frequently used are described briefly below. To bypass the problem of selecting a huge number of regions, Girshick et al. [10] proposed a method where they use selective search by the features of the base CNN to extract just 2000 regions proposals from the image. Liu et al. [11] introduced the Single Shot Multibox Detector (SSD). They added some extra feature layers behind the base model for detection of default boxes in different scales and aspect ratios. At prediction time, the network generates scores for the presence of each object in each default box. Then it produces adjustments to the box to better match the object shape.

There is just one publication over the past few years which gives an survey of generic object detection methods. Liu et al. [12] compared 18 common object detection architectures for generic object detection. There are many other comparisons of specific object detection tasks. For example pedestrian detection [13], face detection [14] and text detection [15].

## 3 Training Dataset

The project is based on the methodology of supervised learning. Thereby the models are trained using a training dataset consisting of many samples. Each sample within the training dataset is tagged with a so called label (also called annotation). The label provides the model with information about the desired output for this sample. During training, the output generated by the model is then compared to the desired output (labels) and the error is determined. This error on the one hand gives information about the current performance of the model and, on the other hand it is used for further mathematical computations to adjust the model's parameters, so that the model's performance improves.

For the training of neural networks in the field of computer vision the following rule of thumb applies: The larger and more diverse the training dataset, the higher the accuracy that can be achieved by the trained model. If you have too little data and/or run it through the model too often, this can lead to so-called overfitting. Overfitting means that instead of learning an abstract concept that can be applied to a variety of data, the model basically memorizes the individual samples [16, 17]. If you train neural networks for the purpose of this project from scratch, it is quite possible that you will need more than 100,000 different images - depending on the accuracy that the model should finally be able to achieve. However, the methodology of the so-called Transfer Learning offers the possibility to transfer results of neural networks, which have already been trained for a specific task, completely or partially to a new task and thus to save time and resources [18]. For this reason, we also applied transfer learning methods within the project.

The training dataset was created manually: A tripod, a mobile phone camera (10 megapixel format 3104 x 3104) and an Apeman Action Cam (20 megapixel format 5120x3840) were used to take 97 images for each of the 25 classes. This corresponds to 2,425 images in total (actually 100 images were taken per class, but only 97 were suitable for use as training data). All images were documented and sorted into close-ups (distance between camera and object less than or equal to 30 cm) and standards (distance between camera and object more than 30 cm). This procedure should ensure the traceability and controllability of the data set. In total, the training data set contains approx. 25% close-ups and approx. 75% standards, each taken on different backgrounds and under different lighting conditions (see Fig. 2). The LabelImg tool was used for the labelling of the data. With the help of this tool, bounding boxes, whose coordinates are stored in either YOLO or Pascval VOC format, can be marked in the images [19].

For the training of the neural networks the created dataset was finally divided into:

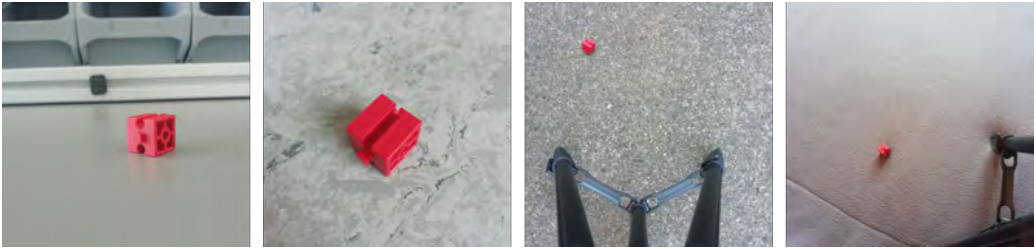

**Fig. 2.** An excerpt from the training data set - images of the component 32_064

– **Training Data (90% of all labelled images):** Images that are used for the training of the models and that pass through the models multiple times during the training.
– **Test Data (10% of all labelled images):** Images that are used for later testing or validation of the training results. In contrast to the images used as training data, the model is presented these images for the first time after training. The goal of this approach, which is common in Deep Learning, is to see how well the neural network recognizes objects in images, that it has never seen before, after the training. Thus it is possible to make a statement about the accuracy and to be able to meet any further training needs that may arise.

## 4   Implementation

The training of deep neural networks is very demanding on resources due to the large number of computations. Therefore, it is essential to use hardware with adequate performance. Since the computations that run for each node in the graph can be highly parallelized, the use of a powerful Graphical Processing Unit (GPU) is particularly suitable. A GPU with its several hundred computing cores has a clear advantage over a current CPU with four to eight cores when processing parallel computing tasks [20]. These are the outline parameters of the project computer in use:

– **Operating System (OS):** Ubuntu 18.04.2 LTS
– **GPU:** GeForce ®GTX 1080 Ti (11 GB GDDR5X-Memory, data transfer speed 11 Gbit/s)

### 4.1   Selected Models

For the intended comparison the Tensorflow Object Detection API was used. Tensorflow Object Detection API is an open source framework based on TensorFlow, which among other things provides implementations of pre-trained object detection models for transfer learning [21, 22]. The API was chosen because of its good and easy to understand documentation and its variety of pre-trained object detection models. For the comparison the following models were selected:

– **ssd_mobilenet_v1_coco:**[11, 23, 24]
– **ssd_mobilenet_v2_coco:**[11, 25, 26]
– **faster_rcnn_inception_v2_coco:**[27–29]
– **rfcn_resnet101_coco:**[30–32]

To ensure comparability of the networks, all of the selected pre-trained models were trained on the COCO dataset [33]. Fundamentally, the algorithms based on CNN models can be grouped into two main categories: region-based algorithms and one-stage algorithms [34].

While both SSD models can be categorized as one-stage algorithms, Faster R-CNN and R-FCN fall into the category of region-based algorithms. One-stage algorithms predict both - the fields (or the bounding boxes) and the class of the contained objects - simultaneously. They are generally considered extremely fast, but are known for their trade-off between accuracy and real-time processing speed. Region-based algorithms consist of two parts: A special region proposal method and a classifier. Instead of splitting the image into many small areas and then working with a large number of areas like conventional CNN would proceed, the region-based algorithm first proposes a set of regions of interest (ROI) in the image and checks whether one of these fields contains an object. If an object is contained, the classifier classifies it [34]. Region-based algorithms are generally considered as accurate, but also as slow. Since, according to our requirements, both accuracy and speed are important, it seemed reasonable to compare models of both categories.

## 4.2   Training Configuration

Besides the collection of pre-trained models for object detection, the Tensorflow Object Detection API also offers corresponding configuration files for the training of each model. Since these configurations have already shown to be successful, these files were used as a basis for own configurations. The configuration files contain information about the training parameters, such as the number of steps to be performed during training, the image resizer to be used, the number of samples processed as a batch before the model parameters are updated (batch size) and the number of classes which can be detected.

To make the study of the different networks as comparable as possible, the training of all networks was configured in such a way that the number of images fed into the network simultaneously (batch size) was kept as small as possible. Since the configurations of some models did not allow batch sizes larger than one, but other models did not allow batch sizes smaller than two, no general value for all models could be defined for this parameter. During training, each of the training images should be passed through the net 200 times (corresponds to 200 epochs). The number of steps was therefore adjusted accordingly, depending on the batch size. If a fixed shape resizer was used in the base configurations, two different dimensions of resizing (default: 300x300 pixels and custom: 512x512 pixels) were selected for the training. Table 1 gives an overview of the training configurations used for the training of the different models.

**Table 1.** Overview of different training runs, configurations and durations

| Model | Batch Size | Steps | Epochs | Image Resizer | | Total Loss | Training Duration |
|---|---|---|---|---|---|---|---|
| | | | | keep_aspect_ratio_resizer min_dimension / max_dimension | fixed_shape_resizer height / width | | |
| ssd_mobilenet_v1_coco | 2 | 217 500 | 200 | - | 512 / 512 | 5.759 | 11h 51m 45s |
| ssd_mobilenet_v1_coco | 2 | 217 500 | 200 | - | 300 / 300 | 5.889 | 09h 45m 45s |
| ssd_mobilenet_v2_coco | 2 | 217 500 | 200 | - | 512 / 512 | 3.289 | 12h 23m 49s |
| ssd_mobilenet_v2_coco | 2 | 217 500 | 200 | - | 300 / 300 | 3.516 | 09h 41m 47s |
| faster_rcnn_inception_v2_coco | 1 | 435 000 | 200 | 600 / 1024 | - | 0.066 | 14h 35m 46s |
| rfcn_resnet101_coco | 1 | 435 000 | 200 | 600 / 1024 | - | 0.031 | 26h 39m 27s |

# 5  Evaluation

In this section we will first look at the training, before we then focus on evaluating the quality of the results and the speed of the selected convolutional neural networks.

## 5.1  Training

When evaluating the training results, we first considered the duration that the neural networks require for 200 epochs (see Fig. 3). It becomes clear that especially the two Region Based Object Detectors (Faster R-CNN Inception V2 and RFCN Resnet101) took significantly longer than the Single Shot Object Detectors (SSD Mobilenet V1 and SSD Mobilenet V2). In addition, the Single Shot Object Detectors clearly show that the size of the input data also has a decisive effect on the training duration: While SSD Mobilenet V2 with an input data size of 300x300 pixels took the shortest time for the training with 9 hours 41 minutes and 47 seconds, the same neural network with an input data size of 512x512 pixels took almost three hours more for the training, but is still far below the time required by RFCN Resnet101 for 200 epochs of training.

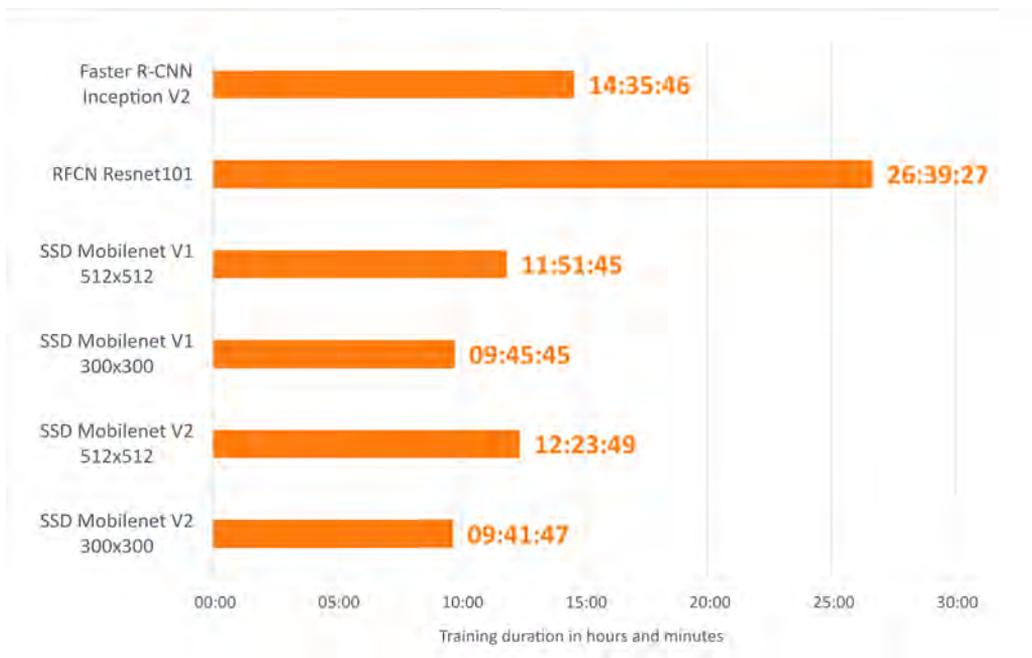

**Fig. 3.** The training duration for 200 epochs and two different input image sizes.

## 5.2  Quality of detection

The next point in which we compared the different networks was accuracy (see Fig. 4). We focused on seeing which of the nets were correct in their detections and how often (absolute values), and we also wanted to see what proportion of the total detections were correct (relative values). The latter seemed to us to make sense especially because some of the nets showed more than three detections for a single object. The probability that the correct classification will be found for the same object with more than one detection is of course higher in this case than if only one detection per object is made. With regard to the later use at the assembly table, however, it does not help us if the neural net provides several possible interpretations for the classification of a component.

Figure 4 shows that, in this comparison, the two Region Based Object Detectors generally perform significantly better than the Single Shot Object Detectors - both in terms of the correct detections and their share of the total detections. It is also noticeable that for the Single Shot Object Detectors, the size of the input data also appears to have an effect on the comparison point on the result. However, there is a clear difference to the previous comparison of the required training durations: While the training duration increased uniformly with increasing size of the images with the Single Shot Object Detectors, such a uniform observation cannot be made with the accuracy, concerning the relation to the input data sizes. While SSD Mobilenet V2 achieves good results with an input data size of 512x512 pixels, SSD Mobilenet V1 delivers the worst result of this comparison for the same input data size (regarding the number of correct detections as well as their share of the total detections). With an input data size of 300x300 pixels, however, the result improves with SSD Mobilenet V1, while the change to a smaller input data size has a deteriorating effect on the result with SSD Mobilenet V2. The best result of this comparison - judging by the absolute values - was achieved by Faster R-CNN Inception V2. However, in terms of the proportion of correct detections in the total detections, the Region Based Object Detector is two percentage points behind RFCN Resnet 101, also a Region Based Object Detector.

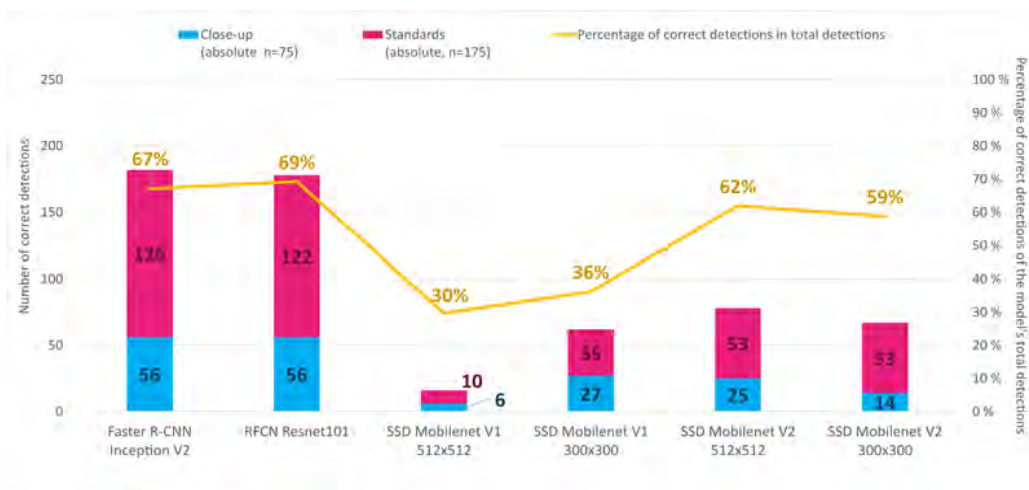

**Fig. 4.** The accuracy of the CNNs we are looking at.

We were particularly interested in how neural networks would react to particularly similar, small objects. Therefore, we decided to investigate the behavior of neural networks within the comparison using an example to illustrate the behavior of the three very similar objects. Figure 5 shows the selected components for the experiment.

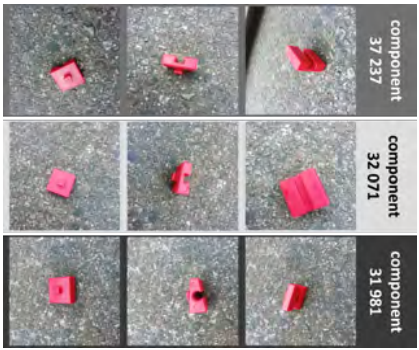

**Fig. 5.** Very similar objects, which we particularly looked at.

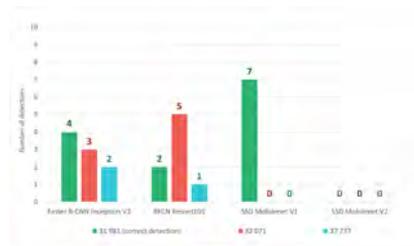

**Fig. 6.** Overview of the behaviour of different neural networks when detecting parts with strong similarity using the example of detections in images of part 31 981.

For each of these three components we examined how often it was correctly detected and classified by the compared neural networks and how often the network misclassified it with which of the similar components. The first and the second component was detected in nearly all cases by both region based approaches. The classification by Inception-v2 and Resnet-101 failed in about one third of images. The SSD networks detected the object in just one of twenty cases but Mobilenet classified this correct.

It has been surprising, that the results for the third component looks very different to the others (see Fig. 6). SSD Mobilenet V1 correctly identified the component in seven of ten images and did not produce any detections that could be interpreted as misclassifications with one of the similar components. SSD Mobilenet V2 did not detect any of the three components, as in the two previous investigations. The results of the two region based object detectors are rather moderate. Faster R-CNN Inception V2 has detected the correct component in four of ten images, but still five misclassifications with the other two components. RFCN Resnet101 has caused many misclassifications with the other two components. Only two of ten images were correctly detected but it had six misclassifications with the similar components.

An other important aspect of the study is the speed, or rather the speed at which the neural networks can detect objects, especially with regard to later use at the assembly table. For the comparison of the speeds on the one hand the data of the GitHub repository of the TensorFlow Object Detection API for the individual neural nets were used, on the other hand the actual speeds of the neural nets within this project were measured. It becomes clear that the speeds measured in the project are clearly below the achievable speeds that are mentioned in the GitHub repository of the TensorFlow Object-Detection API. On the other hand, the differences between the speeds of the Region Based Object Detectors and the Single Shot Object Detectors in the project are far less drastic than expected.

## 6 Conclusions

We have created a training dataset with small, partly very similar components. With this we have trained four common deep neural networks. In addition to the training times, we examined the accuracy and the recognition time with general evaluation data.

In addition, we examined the results for ten images each of three very similar and small components.

None of the networks we trained produced suitable results for our scenario. Nevertheless, we were able to gain some important insights from the results. At the moment, the runtime is not yet suitable for our scenario, but it is also not far from the minimum requirements, so that these can easily be achieved with smaller optimizations and better hardware. It was also important to realize that there are no serious runtime differences between the different network architectures.

The two region based approaches delivered significantly better results than the SSD approaches. However, the results of the detection of the third small component suggest that Mobilenet in combination with a faster R-CNN could possibly deliver even better results. Longer training and training data better adapted to the intended use could also significantly improve the results of the object detectors.

# Optical 3D Object Recognition for Automated Driving

Raphael Schwarz[1], Marin Marinov[2] and Stefan Hensel[1]

[1]Offenburg University of Applied Sciences, [2]Technical University of Sofia
`rschwar1@stud.hs-offenburg.de`, `mbm@tu-sofia.bg`, `stefan.hensel@hs-offenburg.de`

**Abstract.** In this contribution, we propose an system setup for the detection and classification of objects in autonomous driving applications. The recognition algorithm is based upon deep neural networks, operating in the 2D image domain. The results are combined with data of a stereo camera system to finally incorporate the 3D object information into our mapping framework. The detection system is locally running upon the onboard CPU of the vehicle. Several network architectures are implemented and evaluated with respect to accuracy and run-time demands for the given camera and hardware setup.

**Keywords:** Deep neural networks, autonomous driving, optical recognition, system setup

## 1    Introduction

Team Schluckspecht from Offenburg University of Applied Sciences is a very successful participant of the *Shell Eco Marathon* [1]. In this contest, student groups are to design and build their own vehicles with the aim of low energy consumption. Since 2018 the event features the additional *autonomous driving* contest.

In this area, the vehicle has to fulfill several tasks, like driving a parcour, stopping within a defined parking space or circumvent obstacles, autonomously.

For the upcoming season, the *Schluckspecht V* car of the so called *urban concept class* has to be augmented with the according hardware and software to reliably recognize (i. e. detect and classify) possible obstacles and incorporate them into the software framework for further planning.

In this contribution we describe the additional components in hard- and software that are necessary to allow an opitcal 3D object detection. Main criteria are accuracy, cost effectiveness, computational complexity for relative real time performance and ease of use with regard to incorporation in the existing software framework and possible extensibility.

This paper consists of the following sections. At first, the Schluckspecht V system is described in terms of hard- and software components for autonomous driving and the additional parts for the visual object recognition. The second part scrutinizes the object recognition pipeline. Therefore, software frameworks, neural network architecture and final data fusion in a global map is depicted in detail. The contribution closes with an evaluation of the object recognition results and conclusions.

## 2    System Setup

### 2.1    Schluckspecht Car

The Schluckspecht V is a self designed and self build vehicle according to the requirements of the Eco Marathon rules. The vehicle is depicted in Figure 1.

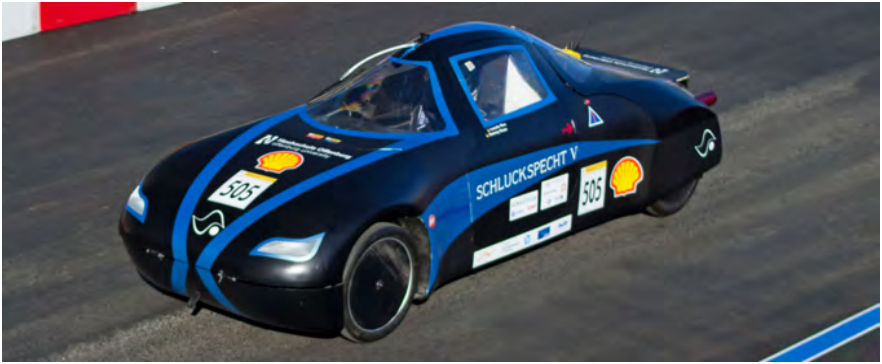

**Fig. 1.** *Schluckspecht V* energy efficient vehicle in urban concept class.

The main features are the relatively large size, including driver cabin, motor area and a large trunk, a fully equipped lighting system and two doors that can be opened separately.

For the autonomous driving challenges, the vehicle is additionally equipped with several essential parts, that are divided into hardware, consisting of actuators, sensors, computational hardware and communication controllers. The software is based on a middle ware, CAN-Open communication layers, localization, mapping and path planning algorithms that are embedded into a high level state machine.

## 2.2 Autonomous Driving Hardware

***Actuators*** The car is equipped with two actors, one for steering and one for braking. Each actor is paired with sensors for measuring steering angle and braking pressure.

***Environmental Sensors*** Several sensors are needed for localization and mapping. Backbone is a multilayer 3D laser scanning system (LiDAR), which is combined with an inertial navigation system that consists of accelerometers, gyroscopes and magnetic field sensors all realized as triads. Odometry information is provided from a global navigation satellite system (GNSS) and two wheel encoders.

***Communication Controller*** The communication is based on two separate CAN-Bus-Systems, one for basic operations and an additional one for the autonomous functions. The hardware CAN nodes are designed and build from the team coupling USB-, I2C-, SPI- and CAN-Open-Interfaces. Messages are send from the central processing unit or the driver depending on drive mode.

***Central Computing Unit*** The trunk of the car is equipped with an industrial grade high performance CPU and an additional graphics processing unit (GPU). CAN communication is ensured with an internal card, remote access is possible via generic wireless components.

## 2.3 Software

***Software Structure*** The Schluckspecht uses a modular software system consisting of several basic modules that are activated and combined within a high level state ma-chine as needed. An overview of the main modules and possible sensors and actuators is depicted in Figure 2

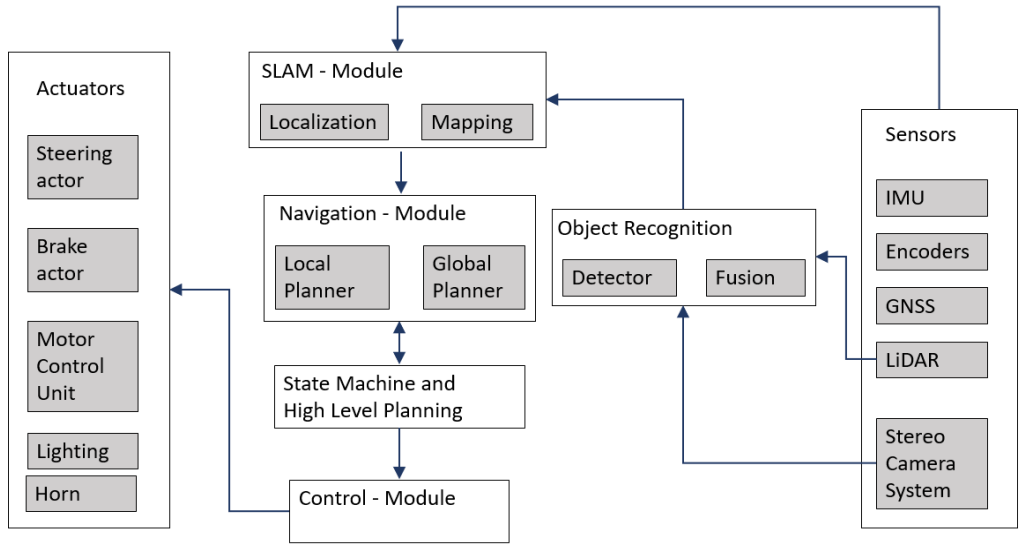

**Fig. 2.** Software modules and component structure overview for autonomous driving.

***Localization and Mapping*** The Schluckspecht V is running a simultaneous localization and mapping (SLAM) framework for navigation, mission planning and environment representation. In its current version we use a graph based SLAM approach based upon the *cartographer* framework developed by Google [2]. We calculate a dynamic occupancy grid map that can be used for further planning. Sensor data is provided by the LiDAR, inertial navigation and odometry systems. An example of a drivable map is shown in Figure 3. This kind of map is also used as base for the localization and placement of the later detected obstacles.

The maps are accurate to roughly 20 centimeters, providing relative localization towards obstacles or homing regions.

***Path Planning*** To make use of the SLAM created maps, an additional module calculates the motion commands from start to target pose of the car. The Schluckspecht is a classical car like mobile system which means that the path planning must take into account the non holonomic kind of permitted movement. Parking maneuvers, close by driving on obstacles or planning a trajectory between given points is realized as a combination of local control commands based upon modeled vehicle dynamics, the so called *local planner*, and optimization algorithms that find the globally most cost efficient path given a cost function, the so called *global planner*. We employ a kinodynamic strategy, the elastic band method presented in [3], for the local planning. Global planning is realized with a variant of the A* algorithm as described in [4].

***Middleware and Communication*** All submodules, namely, localization, mapping, path planning and high-level state machines for each competition are implemented within the *robot operating system (ROS)* middleware [5]. ROS provides a messaging system based upon the subscriber/publisher principle. The single modules are capsuled in a process, called *node*, capable to asynchronously exchange messages as needed. Due to its open source character and an abundance on drivers and helper functions, ROS provides additional features like hardware abstraction, device drivers, visualization and data storage. Data structures for mobile robotic systems, e.g. static and dynamic maps or velocity control messages, allow for rapid development.

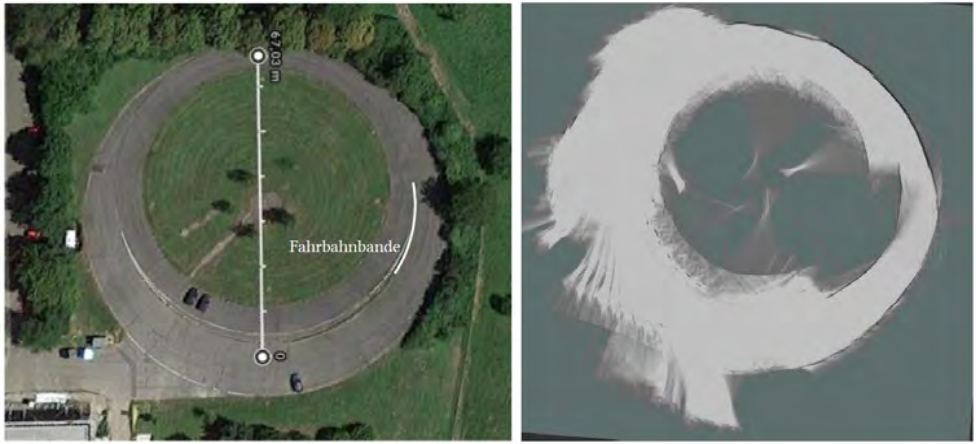

**Fig. 3.** Exemplary occupancy grid map of the Offenburg test track, based on LiDAR, inertial and odometry sensors.

### 2.4 3D Object Recognition

The LiDAR sensor system has four rays, enabling only the incorporation of walls and track delimiters within a map. Therefore, a stereo camera system is additionally implemented to allow for object detection of persons, other cars, traffic signs or visual parking space delimiters and simultaneously measure the distance of any environmental objects.

***Camera Hardware*** A ZED-stereo-camera system is installed upon the car and incorporated into the ROS framework. The system provides a color image streams for each camera and a depth map from stereo vision. The camera images are calibrated to each other and towards the depth information. The algorithms for disparity estimation are running around 50 frames per second making use of the provided GPU.

***Software Framework*** The object recognition relies on deep neural networks. To seamlessly work with the other software parts and for easy integration, the networks are evaluated with tensorflow [6] and pyTorch [7] frameworks. Both are connected to ROS via the openCV image formats providing ROS-nodes and -topics for visualization and further processing.

## 3 Optical Object Recognition

The object recognition pipeline relies on a combination of mono camera images and calibrated depth information to determine object and position. Core algorithm is a deep learning approach with convolutional neural networks.

### 3.1 Deep Convolutional Networks

Main contribution of this paper is the incorporation of a deep neural network object detector into our framework. Object detection with deep neural networks can be subdivided into two approaches, one being a two step approach, where regions of interest are identified in a first step and classified in a second one. The second are so called single shot detectors (like[8]), that extract and classify the objects in one network run. Therefore, two network architectures are evaluated, namely YOLOv3 [9] as a single shot approach and Faster R-CNN [10] as two step model. All are trained on public data sets and fine tuned to our setting by incorporating training images from the Schluckspecht V in the ZED image format.

The models are pre-selected due to their real time capability in combination with the expected classification performance. This excludes the current best instance segmentation network Mask R-CNN [11] due to computational burdens and fast but inaccurate networks based on the mobileNet backbone [12]. The class count is adapted for the contest, in the given case eight classes, including the relevant pedestrian, car, van, tram and cyclist.

### 3.2 Architectures and Training

For this paper, the two chosen network architectures were trained in their respective framework, i. e. *darknet* for the YOLOv3 detector and *tensorflow* for the Faster R-CNN detector. YOLOv3 is used in its standard form with the *Darknet 53* backbone, Faster R-CNN is designed with the *ResNet 101* [13] backbone.

The models were trained on local hardware with the KITTI [14] data set. Alternatively, an open source data set from the teaching company *udacity*, with only three classes (truck, car, pedestrian) was tested.

To deal with the problem of domain adaptation, the training images for YOLOv3 were pre-processed to fit the aspect ratio of the ZED camera. The Faster R-CNN net can cope with ratio variations as it uses a two stage approach for detection based on regions of interest pooling.

Both networks were trained and stored. Afterward, their are incorporated into the system via a ROS node making use of standard python libraries.

### 3.3 Information Fusion within Dynamic Map

The detector output is represented by several labeled bounding boxes within the 2D image. Three dimensional information is extracted from the associated depth map by calculating the center of gravity of each box to get a $x$ and $y$ coordinate within the image. Interpolating the depth map pixels accordingly one gets the distance coordinate $z$ from the depth map to determine the object position $p(x, y, z)$ in the stereo camera coordinate system.

The ease of projection between dieeferent coordinate systems is one reason to use the ROS middleware. The complete vehicle is modeled in a so calle *tranform tree (tf-tree)*, that allows the direct interpolation between different coordinate systems in all six spatial degrees of freedom.

The dynamic map, created in the SLAM subsystem, is now augmented with the current obstacles in the car coordinate system. The local path planner can take these into account and plan a trajectory including kinodynamic constraints to prevent collision or initiate a breaking maneuver.

## 4  Evaluation and Results

### 4.1  Training set evaluation

Both newly trained networks were first evaluated upon the training data. Exemplary results for the KITTI data set are shown in Figure 4.

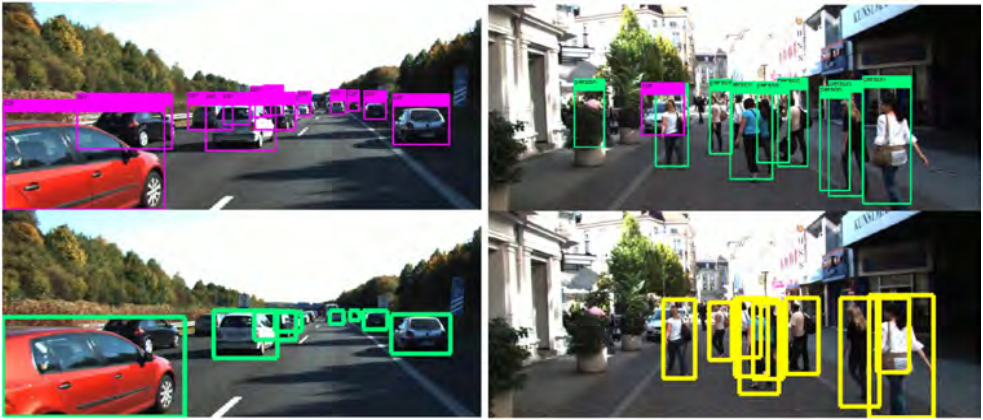

**Fig. 4.** Exemplary results for the object detectors on two KITTI images. YOLOv3 is in the upper half, Faster R-CNN in the lower half (The image was cut to enlarge results).

The results clearly indicate an advantage for the YOLOv3 system, both in speed and accuracy. The Figure depicts good results for occlusions (e. g. the car on the upper right) or high object count (see the black car on the lower left as example). The evaluation on a desktop system showed 50 fps for YOLOv3 and approximately 10 fps for Faster R-CNN.

### 4.2  ZED camera system evaluation

After validating the performance upon the training data, both networks were started as a ROS node and tested upon real data of the Schluckspecht vehicle.

**Table 1.** Qualitative comparison of detection architectures.

| Architecture | Frame rate [fps] | Detection quality | | Quality w.r.t. size | |
| | | Vehicles | Pedestrians | large objects | small objects |
| --- | --- | --- | --- | --- | --- |
| YOLOv3 | 9 − 10 | ++ | − | + | + |
| Faster R-CNN | 4 − 5 | ++ | + | ++ | − |

As the training data differs from the ZED-camera images in format and resolution, several adaptions were necessary for the YOLOv3 detector. The images are cropped in real time before presented to the neural net to emulate the format of the training images. The R-CNN like two stage networks are directly connected to the ZED node.

The test data is not labeled as ground truth. It is therefore not possible to give quantitative results for the recognition task. Table 1 gives a quantitative overview of the object detection and classification, the subsequent Figures give some expression of exemplary results.

The evaluation on the Schluckspecht videos showed an advantage for the YOLOv3 network. Main reason is the faster computation, which results in a frame rate nearly twice as high compared to two stage detectors. In addition, the recognition of objects in the distance, i. e. smaller objects is a strong point of YOLO. The closer the camera gets, the bigger is the balance shift towards Faster R-CNN, that outperforms YOLO on all categories for larger objects.

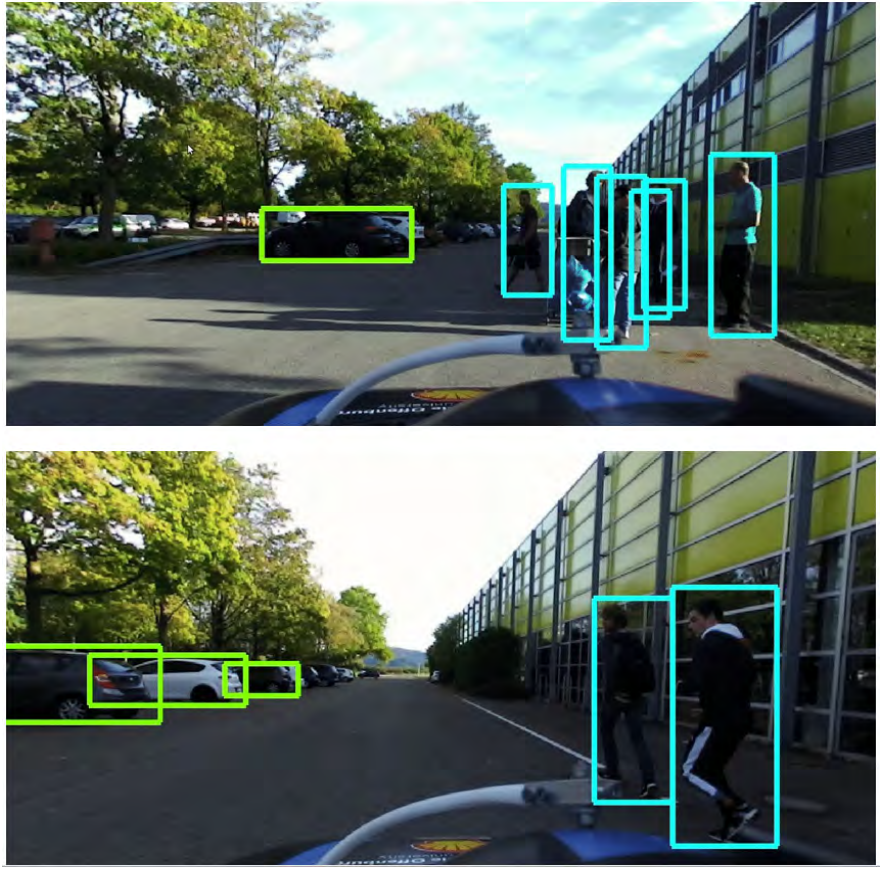

**Fig. 5.** Exemplary results for the object detectors on real Schluckspecht images (Persons colored cyan, cars green).

Figure 5 shows results from the YOLO-detector for a mixture of persons and cars. What becomes apparent is a maximum detection distance of approximately 30 meters, from which on cars become to small in size. Figure 6 shows an additional result demonstrating the detection power for partially obstructed objects.

Another interesting finding was the capability of the networks to generalize. Faster R-CNN copes much better with new object instances than YOLOv3. Persons with so far unknown cloth color or darker areas with vehicles remain a problem for YOLO, but commonly not for the R-CNN. The domain transfer from training data in Berkeley and KITTI to real ZED vehicle images proved problematic.

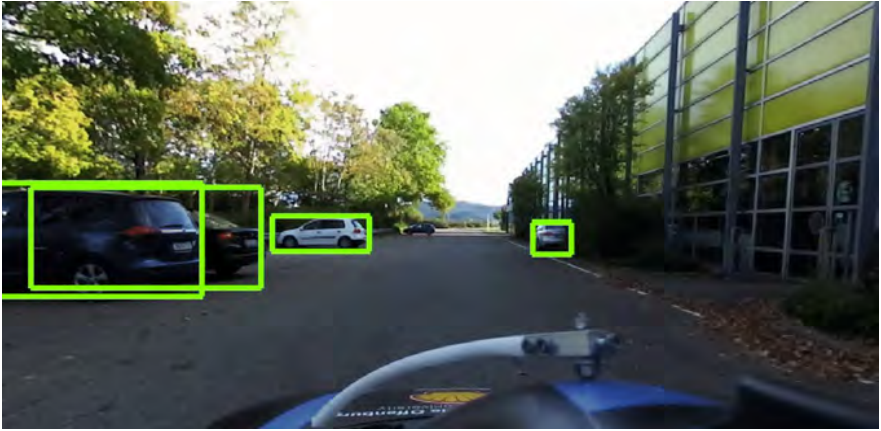

**Fig. 6.** Exemplary results for the object detectors on real Schluckspecht images.

## 5    Conclusions

This contribution describes an optical object recognition system in hard- and software for the application in autonomous driving under restricted conditions, within the Shell Eco Marathon competition. An overall overview of the system and the incorporation of the detector within the framework is given.

Main focus was the evaluation and implementation of several neural network detectors, namely YOLOv3 as one shot detector and Faster R-CNN as a two step detector, and their combination with distance information to gain a three dimensional information for detected objects. For the given application, the advantage clearly lies with YOLOv3. Especially the achievable frame rate of minimum 10 Hz allows a seamless integration into the localization and mapping framework. Given the velocities and map update rate, the object recognition and integration via sensor fusion for path planning and navigation works in quasi real-time.

For future applications we plan to further increase the detection quality by incorporating new classes and modern object detector frameworks like M2Det [15]. This will additionally increase frame rate and bounding box quality. For more complex tasks, the data of the 3D-LiDAR system shall be directly incorporated into the fusion framework to enhance the perception of object boundaries and object velocities.

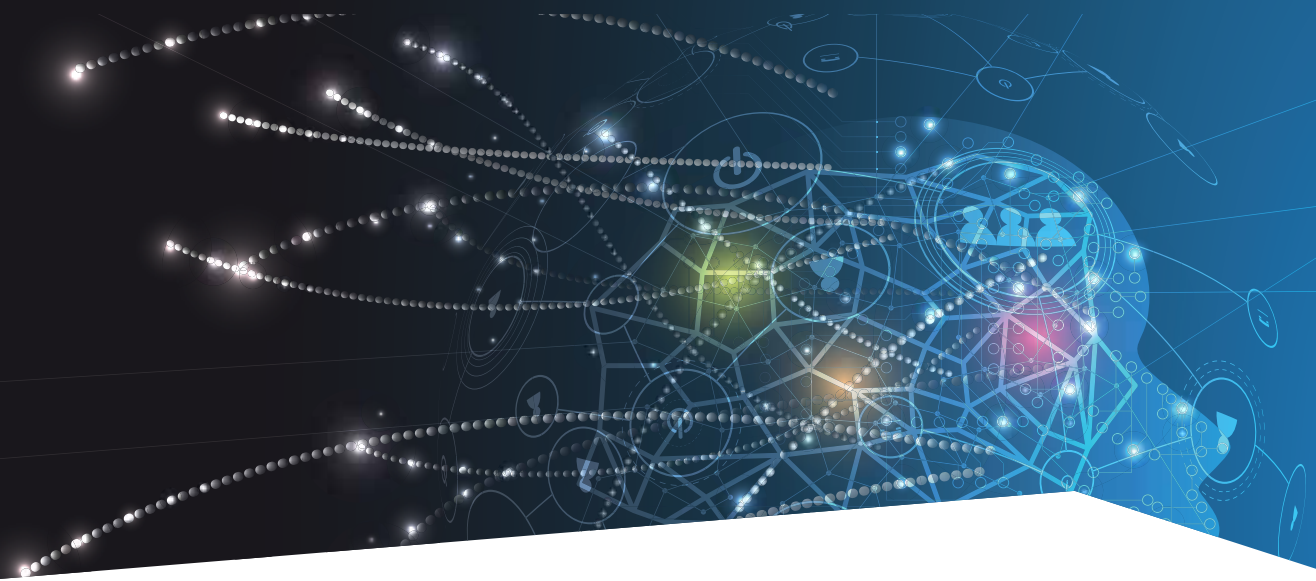

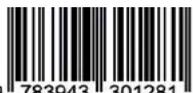